\documentclass{article}

\PassOptionsToPackage{numbers, compress}{natbib}

\usepackage[main, final]{neurips_2026}

\usepackage[utf8]{inputenc} %
\usepackage[T1]{fontenc}    %
\usepackage{hyperref}       %
\usepackage{url}            %
\usepackage{booktabs}       %
\usepackage{amsfonts}       %
\usepackage{nicefrac}       %
\usepackage{microtype}      %
\usepackage{xcolor}         %
\usepackage{wrapfig}        %

\usepackage{graphicx}
\usepackage{subcaption}
\usepackage{amsmath}
\usepackage{amssymb}
\usepackage{mathtools}
\usepackage{amsthm}
\usepackage{nicematrix}
\usepackage{tikz}
\usetikzlibrary{calc}
\usepackage[capitalize,noabbrev]{cleveref}
\usepackage{titlesec}

\titlespacing*{\paragraph}{0pt}{0.0em}{0.5em}

\titlespacing*{\section}{0pt}{8pt plus 2pt minus 2pt}{4pt plus 2pt minus 2pt}
\titlespacing*{\subsection}{0pt}{6pt plus 2pt minus 2pt}{3pt plus 2pt minus 2pt}
\titlespacing*{\subsubsection}{0pt}{5pt plus 2pt minus 2pt}{2pt plus 2pt minus 2pt}

\AtBeginDocument{%
  \setlength{\abovedisplayskip}{6pt plus 2pt minus 2pt}
  \setlength{\belowdisplayskip}{6pt plus 2pt minus 2pt}
  \setlength{\abovedisplayshortskip}{3pt plus 1pt minus 1pt}
  \setlength{\belowdisplayshortskip}{3pt plus 1pt minus 1pt}
}

\theoremstyle{plain}

\theoremstyle{definition}

\theoremstyle{remark}

\ExplSyntaxOn
\makeatletter

\cs_set:Npn \dotrule
  {
    \noalign \bgroup
    \peek_meaning:NTF [
      { \__dose_dotrule: }
      { \__dose_dotrule: [ \lightrulewidth ] }
  }

\cs_set:Npn \__dose_dotrule: [ #1 ]
  {
    \skip_vertical:n { \aboverulesep + \belowrulesep + #1 }
    \egroup
    \tl_gput_right:Nn \g_nicematrix_code_after_tl
      { \__dose_dotrule:nn { \c@iRow } { #1 } }
  }

\cs_new_protected:Nn \__dose_dotrule:nn
  {
    {
      \dim_set:Nn \l_tmpa_dim { \aboverulesep + ( #2 ) / 2 }
      \CT@arc@
      \tikz \draw [ dotted , line~width = #2 ]
        ([yshift=-\l_tmpa_dim]\the#1-|1)
        --
        ([yshift=-\l_tmpa_dim]\the#1-| \int_eval:n { \the\c@jCol + 1 }) ;
    }
  }

\makeatother
\ExplSyntaxOff

\newcommand{\bzero}{\boldsymbol{0}}
\newcommand{\bdelta}{\boldsymbol{\delta}}
\newcommand{\btheta}{\boldsymbol{\theta}}
\newcommand{\btau}{\boldsymbol{\tau}}
\newcommand{\bphi}{\boldsymbol{\phi}}

\newcommand{\bPhi}{\boldsymbol{\Phi}}
\newcommand{\bg}{\boldsymbol{g}}
\newcommand{\bh}{\boldsymbol{h}}
\newcommand{\bu}{\boldsymbol{u}}
\newcommand{\bv}{\boldsymbol{v}}
\newcommand{\bx}{\boldsymbol{x}}
\newcommand{\by}{\boldsymbol{y}}
\newcommand{\bl}{\boldsymbol{l}}
\newcommand{\bH}{\boldsymbol{H}}
\newcommand{\bU}{\boldsymbol{U}}
\newcommand{\bW}{\boldsymbol{W}}
\newcommand{\bX}{\boldsymbol{X}}
\newcommand{\bbeta}{\boldsymbol{\beta}}
\newcommand{\set}[1]{\mathcal{#1}}
\newcommand{\bref}[1]{(\ref{#1})}
\newcommand{\argmin}{\mathop{\rm arg~min}\limits}

\usepackage[disable,textsize=tiny]{todonotes}

\title{Transferring Visual Explainability from Self-Explaining to Prediction-Only Vision Transformers via Task Arithmetic}

\author{%
  Yuya Yoshikawa \\
  STAIR Lab, Chiba Institute of Technology \\
  \texttt{yoshikawa@stair.center} \\
  \And
  Ryotaro Shimizu \\
  ZOZO Research \\
  \texttt{ryotaro.shimizu@zozo.com} \\
  \And
  Takahiro Kawashima \\
  ZOZO Research \\
  \texttt{takahiro.kawashima@zozo.com} \\
  \And
  Yuki Saito \\
  ZOZO Research \\
  \texttt{yuki.saito@zozo.com} \\
}

\begin{document}

\maketitle

\begin{abstract}
In image classification scenarios where both prediction and explanation efficiency are required, self-explaining models that perform both tasks in a single inference are effective.
However, for users who already have prediction-only models, training a new self-explaining model from scratch is costly in both labeling and computation.
This study proposes a method to transfer the visual explanation capability of self-explaining Vision Transformer (ViT) models learned in a source domain to prediction-only, zero-shot-capable ViT models based on vision-language models (VLMs) in a target domain via task arithmetic, without target-side explanation supervision.
The proposed method endows target models with explanation capability by adding an \textbf{explainability vector} induced by explanation supervision in the source domain based on the task arithmetic framework.
Experiments on 10 diverse target datasets show that a single explainability vector learned on ImageNet-1k augmented with patch-level explanation supervision transfers consistently, improving explanation quality while largely preserving classification accuracy.
Within-model faithfulness evaluation and comparisons with control vectors (e.g., random vectors of the same norm) show that the transferred explanations are faithful to the model's own predictions and that the effect is specific to the explainability vector.
Beyond the transfer itself, we further investigate whether transfer success can be anticipated \emph{a priori} from model-internal features alone, and find that, for a fixed source, a small set of such features carries non-trivial signal about the transfer outcome on held-out targets.
Code is available at \url{https://github.com/yuyay/explainability_vector}.
\end{abstract}

\section{Introduction}\label{sec:intro}
Visual explainability, which provides human-understandable explanations for a prediction model's outputs, is important for various applications, such as medical image diagnosis~\cite{nazir2023survey}, autonomous driving~\cite{atakishiyev2024explainable}, manufacturing~\cite{alexander2024interrogative}, cybersecurity~\cite{capuano2022explainable}, and model debugging~\cite{apicella2023strategies}.

Although explanations provide insights to users, generating them often costs far more computation than prediction alone.
When a trained model is already available, visual explanations can be generated using post-hoc explanation methods~\cite{yoshikawa2025explaining}, such as SHAP~\cite{Lundberg2017-ii} and Grad-CAM~\cite{Selvaraju2020-fx}.
However, SHAP typically requires hundreds of model inferences per image, and Grad-CAM requires a backward computation of gradients with respect to intermediate feature maps.
Such computational challenges can be addressed by using {\it self-explaining models}, which are designed to perform both prediction and explanation in a single inference~\cite{alvarez2018towards,ji2025comprehensive,yoshikawa2022neural,yoshikawa2023gaussian}.
Although self-explaining models are very efficient in the inference phase, their training can be costly, because it requires ground-truth (GT) explanation data for user-desired explanations~\cite{ross2017right,yang2023efficient} or numerous model forward computations to learn these explanations~\cite{wang2021shapley}.

The goal of this study is to efficiently equip prediction-only, zero-shot-capable Vision Transformer (ViT) classifiers based on vision-language models (VLMs), such as CLIP image classifiers, with visual explainability without paying the high training cost of building a self-explaining model in the target domain.
To this end, we focus on {\it task arithmetic}~\cite{ilharco2023editingmodelstaskarithmetic}, which enables model editing using {\it task vectors} that represent the model's capabilities learned for different tasks across different domains.
Task vectors can also express {\it task analogies}, \emph{``\textsf{A} is to \textsf{B} as \textsf{C} is to \textsf{D},''} with \textsf{A} and \textsf{B} in a source domain and \textsf{C} and \textsf{D} in a target domain. If the analogy holds, the task vector for \textsf{D} follows from those for \textsf{A}, \textsf{B}, and \textsf{C} without training the model on \textsf{D}.

Inspired by the idea, we propose a method to transfer the visual explainability of self-explaining ViTs learned in a source domain to such prediction-only models in a target domain.
Our self-explaining model extends VLM-based image classifiers such as CLIP~\cite{radford2021learning} by producing patch-level attributions from patch and class-label text embeddings alongside the class logits (Figure~\ref{fig:model}).
Because their classification head is built from frozen text embeddings of class names, all models finetuned from the same VLM checkpoint share one parameter space across label sets.
This is the standard finetuning setting for such models~\cite{wortsman2022robust,ilharco2023editingmodelstaskarithmetic}.
We train two source-domain models, with and without explanation supervision, and define the \textbf{explainability vector} as their parameter difference.
Adding this vector to a target-domain prediction-only model transfers the explanation capability, instantiating the task analogy \emph{``prediction-only is to prediction-with-explanation in the source domain as prediction-only is to prediction-with-explanation in the target domain.''} 

Our main contributions are as follows:
{
  \setlength{\leftmargini}{20pt}         %
  \vspace{-0.5\baselineskip}            %
  \begin{itemize}
    \setlength{\topsep}{0pt}
    \setlength{\partopsep}{0pt}
    \setlength{\itemsep}{0pt}
    \setlength{\parsep}{0pt}
    \item We propose the first task-arithmetic approach to transfer visual explainability from self-explaining to prediction-only VLM-based ViTs across domains without target-side explanation supervision, by defining an explainability vector as a task vector induced by explanation supervision in the source domain (Section~\ref{sec:proposed:transfer}).
    \item We empirically show that a single explainability vector learned on ImageNet-1k dataset augmented with patch-level explanation supervision transfers consistently to 10 diverse target datasets, improving explanation quality while largely preserving classification accuracy, and that the transferred explanations are faithful to the model itself and specific to the explainability vector (Section~\ref{sec:experiment:imagenet}).
    \item We additionally find that the success of such transfer is predictable to a meaningful degree. When the source is fixed and targets are held out, a small set of model-internal features computed without target-side explanation supervision yields held-out Pearson $r > 0.7$ on multiple explanation-quality metrics. This provides initial evidence that transferability is a measurable property of the source--target pair (Sections~\ref{sec:proposed:predict}, \ref{sec:experiment:predict}).
  \end{itemize}
}

\section{Related Work}\label{sec:RW}
\paragraph{Self-Explaining Vision Models}

To improve the faithfulness and computational efficiency of explanations, various self-explaining models that are capable of performing both prediction and explanation have been proposed~\cite{ibrahim2023explainable,ji2025comprehensive}.
In ViT-based models, the attention map between the class token and image patches can be visualized as the importance of each image patch~\cite{yu2023ex,qiang2023interpretability}.
Since some studies have raised concerns about the validity of attention as an explanation~\cite{serrano2019attention,jain2019attention,chefer2021transformer}, several studies have proposed methods for training models with constraints that explicitly encourage attention to be more aligned with human-interpretable explanations~\cite{yu2023ex,qiang2023interpretability}.
As with our study, several studies have also addressed training self-explaining models to directly produce Shapley value-based explanations~\cite{wang2021shapley,alkhatib2025prediction}.
These approaches encourage the explanations to approximate Shapley values through the design of the objective function or network architecture, without requiring precomputed GT Shapley value explanations.
Similar approaches have been proposed in the research on amortized explainers~\cite{chuang2023efficient}, which learn the behavior of post-hoc explainers for a trained prediction model in advance and generate only explanations efficiently in the inference phase.
Similar to ours, \citet{covert2023learning} proposed a method that trains a ViT to output patch-level attributions based on Shapley values from patch tokens.
While this approach can efficiently train models to produce Shapley value explanations, it is difficult to generalize it to explanations other than Shapley values due to its specialization.

\paragraph{Post-hoc Explainers for ViTs and CLIP}

Post-hoc explainers designed for ViTs trace relevance through the attention and token transformations of the network, e.g., relevancy propagation~\cite{chefer2021generic,chefer2021transformer} and TokenTM~\cite{wu2024tokentm}.
Several methods further improve CLIP-specific interpretability by analyzing or modifying its internal representations: attention surgery~\cite{li2023clip_surgery}, gradient-based explanations~\cite{zhao2024grad_eclip}, layer/head decomposition~\cite{gandelsman2024interpreting}, and adversarial finetuning~\cite{gong2025boosting}.
These works analyze a trained model or modify it to improve its interpretability, whereas our method transfers an attribution-producing capability via task arithmetic so that the model itself produces attributions in its forward pass.
Their attribution maps can also serve as the explanation supervision for the GT attributions (Section~\ref{sec:proposed:learning}).

\paragraph{Visual Explainability Transfer}

Some studies have proposed methods for transferring visual explainability to other domains.
Transferable Vision Explainer (TVE)~\cite{wang2024tve} pretrains an encoder on a large-scale dataset like ImageNet to output meta-attribute vectors from images.
In the inference phase, it uses a combination of this encoder and a pretrained classification head for the target task to output patch contributions.
TVE leaves the target classifier untouched and produces attributions with a separate explainer network, whereas our method edits the finetuned target so that the target itself produces explanations.
\citet{li2024surprising} showed that transferring attention patterns, which are often read as explanations, from a teacher to a student ViT recovers most of the benefit of weight finetuning, although the gap widens when the pretraining and downstream data distributions differ.

\section{Task Arithmetic}
\label{sec:arithmetic}
Our approach builds on the task arithmetic framework~\cite{ilharco2023editingmodelstaskarithmetic}.
Task arithmetic edits neural network models by manipulating differences in their parameter spaces.
The central idea is that the effect of learning a specific capability (or task) can be represented as a vector in the parameter space, and these vectors for different tasks can be algebraically combined to transfer or compose capabilities across models.
The validity of task arithmetic is supported by the observations that aggregating models by adding multiple model parameters can improve model performance~\cite{izmailov2018averaging,wortsman2022model,wortsman2022robust,yoshida2025mastering,yoshida2025distacconditioningtaskvectors}.
Beyond this validity, task arithmetic has been used to compose or remove skills such as mathematical reasoning~\cite{chen2025bring}, fairness~\cite{gonzalez2025on}, unlearning~\cite{ni2024forgetting}, and bias mitigation~\cite{shirafuji2025bias}, but its capability for transferring visual explainability remains unexplored.

Formally, let $\btheta_{\rm base} \in \mathbb{R}^D$ denote the parameters of a base model where $D$ is the dimensionality of the model parameters, and let $\btheta_{\rm ft} \in \mathbb{R}^D$ denote the parameters of the same model after finetuning on a specific task from the base parameters $\btheta_{\rm base}$.
The difference $\btau = \btheta_{\rm ft} - \btheta_{\rm base}$ is called the \emph{task vector} for that task, which is expected to contain the knowledge and capabilities necessary to perform the task.
The addition operation of task vectors allows us to construct a model that is capable of performing multiple tasks simultaneously, similar to multi-task learning.
Specifically, given $J$ task vectors $\{\btau_j\}_{j=1}^{J}$, where $j$ is the task index, a new model composed of capabilities from these tasks can be constructed as
\begin{equation}
\btheta = \btheta_{\rm base} + \lambda \sum_{j=1}^{J} \btau_j,
\label{eq:preliminary:addition}
\end{equation}
where $\lambda \in \mathbb{R}$ is a scaling factor controlling the extent to which the capabilities are added (if $\lambda > 0$) or removed (if $\lambda < 0$).
A single $\lambda$ is used in~\bref{eq:preliminary:addition} for simplicity. A separate coefficient $\lambda_j$ per task vector is a natural extension within the same framework (cf.~\cite{yoshida2025mastering}), and our transfer rule in Section~\ref{sec:proposed:transfer} uses separate coefficients for the classification and explainability vectors.

For a more complicated arithmetic operation that we are interested in, analogical relationships between tasks can be considered.
Suppose we have models finetuned for tasks \textsf{A} and \textsf{B} in the same domain (referred to as the \emph{source domain} for convenience), and for task \textsf{C} in another domain (referred to as the \emph{target domain} for convenience).
Our aim is to obtain model parameters $\btheta$ that achieve a new task \textsf{D} in the target domain.
If the relationship between \textsf{A} and \textsf{B} is analogous to that between \textsf{C} and \textsf{D}, we can estimate the parameters for task \textsf{D} using the following analogy:
\begin{equation}
\btau_{\textsf{A}} - \btau_{\textsf{B}} \approxeq \btau_{\textsf{C}} - \btau_{\textsf{D}}
\quad\rightarrow\quad
\btau_{\textsf{D}} \approxeq \btau_{\textsf{C}} - \btau_{\textsf{A}} + \btau_{\textsf{B}}.
\label{eq:preliminary:analogy}
\end{equation}
Then, we can endow the model with the capability for task \textsf{D} by adding the task vector $\btau_{\rm D}$ to the base parameters $\btheta_{\rm base}$ as follows:
\begin{equation}
\btheta = \btheta_{\rm base} + \lambda \btau_{\textsf{D}}.
\label{eq:preliminary:theta_analogy}
\end{equation}

This property transfers learned capabilities from the source to the target domain even when supervision for task \textsf{D} is unavailable there.
The next section applies it to visual explainability.

\section{Proposed Method}\label{sec:proposed}
\space%
\subsection{Self-Explaining Vision Transformer Model}\label{sec:proposed:model}

\begin{figure}[t]
\centering
\includegraphics[width=0.9\columnwidth]{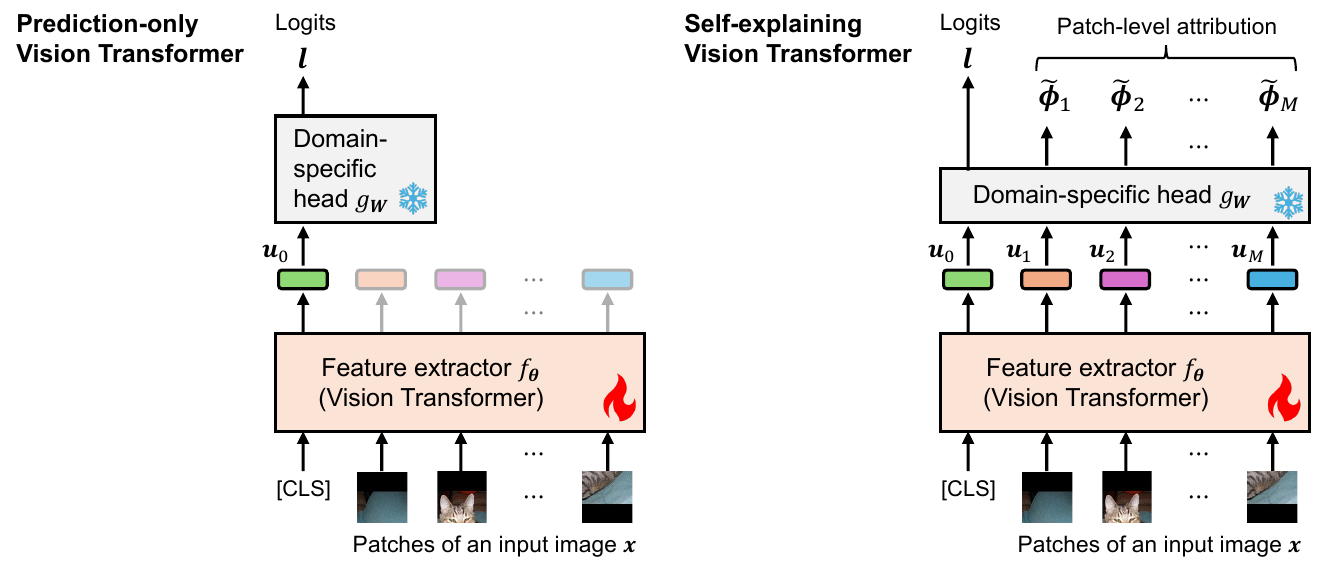}
\vspace{-2mm}
\caption{
Model architectures of prediction-only ViT (left) and our self-explaining ViT (right).
Gray boxes represent functions with fixed parameters.
All the learnable parameters $\btheta$ are only in the feature extractor (filled in orange).
}
\label{fig:model}
\end{figure}

The proposed method requires a self-explaining image classifier that predicts the logits of class labels and infers patch-level attributions corresponding to the contributions of the patches for each class.
As a requirement for the model to utilize task arithmetic, it is necessary that the model is represented by parameters $\btheta \in \mathbb{R}^D$ even if the source and target label sets are different.
In other words, the model should be capable of zero-shot classification.
To satisfy the requirement, our self-explaining model is designed as shown in Figure~\ref{fig:model}, which is an extension of ``prediction-only'' ViT image classifiers~\cite{radford2021learning}.
It consists of a feature extractor and a domain-specific head.
We denote the input space as $\set{X}$, and the number of patches in an input image as $M$.
The feature extractor $f_{\btheta}$ is a neural network with the ViT architecture.
It takes $M$ patches of an input image $\bx \in \set{X}$ and a special token $\mathtt{[CLS]}$ as input and outputs $M+1$ $K$-dimensional feature vectors.
The domain-specific head is a linear function $g_{\bW}: \mathbb{R}^{K} \rightarrow \mathbb{R}^{C}$ with weights $\bW \in \mathbb{R}^{K \times C}$ where $C$ is the number of classes.
Let $\bu_0 \in \mathbb{R}^{K}$ denote the output feature vector for $\mathtt{[CLS]}$ and $\bu_m \in \mathbb{R}^{K}$ that for the $m$th patch, where $m \in \{1,2,\ldots,M\}$ is a patch index.
The same head $g_{\bW}$ is applied to every token.
Applied to $\bu_0$, it yields the class logits $\bl = g_{\bW}(\bu_0) = [ l_1,l_2,\ldots,l_{C} ]^{\top} \in \mathbb{R}^{C}$, exactly as in the prediction-only model.
Applied to $\bu_m$, it yields the unnormalized patch-level attributions $\tilde{\bphi}_{m} = g_{\bW}(\bu_m) = [ \tilde{\phi}_{m1},\tilde{\phi}_{m2},\ldots,\tilde{\phi}_{mC} ]^\top \in \mathbb{R}^{C}$, whose $c$th entry is the contribution of the $m$th patch to the $c$th class.
Stacking the $M$ patch vectors as columns gives the attribution matrix $\tilde{\bPhi} = [ \tilde{\bphi}_{1},\tilde{\bphi}_{2},\ldots,\tilde{\bphi}_{M} ] \in \mathbb{R}^{C \times M}$, whose rows index classes and whose columns index patches. Its $c$th row $\tilde{\bPhi}_{c} \in \mathbb{R}^{M}$ is the patch-level explanation of the decision for the $c$th class.
$\bW$ is constructed from the class-name text embeddings produced by the text encoder of the VLM\footnote{For example, if the class name is ``cat,'' we use the average of text embeddings generated by the text encoder for multiple input texts, such as ``a photo of a cat,'' ``art of the cat,'' and so on.}, i.e., $\bW_{\cdot,c} \in \mathbb{R}^{K}$ is the text embedding associated with the $c$th class.
In our model, $\bW$ is fixed during training and is therefore not included in the model parameters.
Our model reuses the VLM's pretrained parameters as the base parameters $\btheta_{\rm base}$.
Note that they do not have the capability to explain the classification results because the VLM is not optimized for explanation.

\paragraph{Remarks}
The key difference between our self-explaining and prediction-only ViTs is that the domain-specific head $g_{\bW}$ also produces the patch-level attributions. This shared-head constraint does not degrade classification or explanation quality compared with a separate attribution head (Appendix~\ref{sec:appendix:shared_vs_separated}).
Although the two outputs are produced in parallel, head-level circuit analysis shows that classification and explanation share most of their backbone computation, with specialization emerging only in the final layers (Appendix~\ref{sec:appendix:circuit_overlap}).

\subsection{Learning Classification Capability and Explainability}\label{sec:proposed:learning}
We acquire explainability on a source domain by finetuning our self-explaining model in a supervised manner.
First, we construct a source dataset, a labeled image dataset with explanation supervision $\set{D}^{\rm S} = \{(\bx^{(n)},y^{(n)},\bphi^{(n)})\}_{n=1}^{N^{\rm S}}$ where $\bx^{(n)} \in \set{X}$ is the $n$th input image, $y^{(n)} \in \{1,2,\ldots,C^{\rm S}\}$ is the GT class label of the input $\bx^{(n)}$, and $\bphi^{(n)} \in \mathbb{R}^M$ is the GT patch-level attributions given $\bx^{(n)}$ and $y^{(n)}$.
We can design the GT patch-level attributions flexibly according to users' purposes and applications. They are typically given by human annotators or by computing a post-hoc explanation method, such as SHAP~\cite{Lundberg2017-ii,noauthor_undated-ca} and Grad-CAM~\cite{Selvaraju2020-fx}.
Since obtaining such attributions may be highly expensive, we assume that they are only available in the source domain.

Initialized at the base parameters $\btheta_{\rm base}$, we learn the model parameters by solving the following optimization problem:
\begin{equation}
\btheta^{\rm S}_{\rm ft\star} = \argmin_{\btheta}~ \alpha \set{L}_{\rm cls}(\btheta; \set{D}^{\rm S}) + (1-\alpha) \set{L}_{\rm exp}(\btheta; \set{D}^{\rm S}),
\label{eq:proposed:objective}
\end{equation}
where $\set{L}_{\rm cls}$ is the cross-entropy loss between the logits and the GT class label.
The explanation loss $\set{L}_{\rm exp}$ is defined as
\begin{equation}
\set{L}_{\rm exp}(\btheta; \set{D}^{\rm S}) = \frac{1}{M |\set{D}^{\rm S}|} \sum_{n=1}^{|\set{D}^{\rm S}|} \| \mathsf{Norm}(\tilde{\bPhi}^{(n)}_{y^{(n)}}) - \mathsf{Norm}(\bphi^{(n)})\|^2_2,
\label{eq:proposed:exp_loss}
\end{equation}
where $\tilde{\bPhi}^{(n)}_{y^{(n)}} \in \mathbb{R}^{M}$ is the $y^{(n)}$th row of $\tilde{\bPhi}$, which is the predicted patch-level attributions for the input $\bx^{(n)}$ when it is classified as the $y^{(n)}$th class, and $\mathsf{Norm}(\bv)$ is a normalization function such that $\mathsf{Norm}(\bv) = (\bv - \mu_{\bv}) / \sigma_{\bv}$ where $\mu_{\bv}$ and $\sigma_{\bv}$ are the mean and standard deviation of $\bv$, respectively.
The hyperparameter $\alpha \in [0,1]$ balances the two losses, $\set{L}_{\rm cls}$ and $\set{L}_{\rm exp}$.
When $\alpha = 1$, the learned parameters are identical to the finetuned parameters of the prediction-only VLM-based image classifier, $\btheta^{\rm S}_{\rm ft}$, where the subscript `{\rm ft}' stands for finetuned only for classification.
The case of $\alpha = 0$ results in amortized explainers~\cite{covert2023learning,yang2023efficient} where the learned parameters have only the explanatory capability.
What we want are the model parameters $\btheta^{\rm S}_{\rm ft\star}$ learned with $0 < \alpha < 1$ that have both the classification capability (denoted by `ft') and the explainability (denoted by `$\star$') in the source domain.
\space%
\subsection{Transferring Visual Explainability}\label{sec:proposed:transfer}
For the target domain, we assume that a prediction-only model $\btheta^{\rm T}_{\rm ft}$, obtained by minimizing $\set{L}_{\rm cls}$ on labeled target images, is available, but GT patch-level attributions are not.
Therefore, the self-explaining model in the target domain, $\btheta^{\rm T}_{\rm ft\star}$, cannot be trained directly.
Our goal is to obtain $\tilde{\btheta}^{\rm T}_{\rm ft\star}$, a good approximation of $\btheta^{\rm T}_{\rm ft\star}$, by transferring the explainability encoded in the source-domain model $\btheta^{\rm S}_{\rm ft\star}$ to $\btheta^{\rm T}_{\rm ft}$ without any additional training.

To this end, we utilize the task arithmetic framework introduced in Section \ref{sec:arithmetic}.
Based on the task analogy~\bref{eq:preliminary:analogy}, the task vectors obtained from the parameters trained on the source and target domains are expected to have the following relationship:
\begin{equation}
\btau^{\rm S}_{\rm ft\star} - \btau^{\rm S}_{\rm ft} \approxeq \btau^{\rm T}_{\rm ft\star} - \btau^{\rm T}_{\rm ft},
\label{eq:proposed:analogy}
\end{equation}
where $\btau^{\rm S}_{\rm ft\star} = \btheta^{\rm S}_{\rm ft\star} - \btheta_{\rm base}$ and $\btau^{\rm T}_{\rm ft\star} = \btheta^{\rm T}_{\rm ft\star} - \btheta_{\rm base}$ are the task vectors representing both the classification capability and explainability for the source and target domains, respectively. $\btau^{\rm S}_{\rm ft} = \btheta^{\rm S}_{\rm ft} - \btheta_{\rm base}$ and $\btau^{\rm T}_{\rm ft} = \btheta^{\rm T}_{\rm ft} - \btheta_{\rm base}$ are the task vectors representing only the classification capability for the source and target domains, respectively.
The equality in \bref{eq:proposed:analogy} holds if and only if $\btau^{\rm S}_{\star} = \btau^{\rm T}_{\star}$, where $\btau^{\rm S}_{\star} = \btheta^{\rm S}_{\rm ft\star} - \btheta^{\rm S}_{\rm ft}$ and $\btau^{\rm T}_{\star} = \btheta^{\rm T}_{\rm ft\star} - \btheta^{\rm T}_{\rm ft}$ are the task vectors representing the explainability for the source and target domains, respectively.
We refer to $\btau^{\rm S}_{\star}$ and $\btau^{\rm T}_{\star}$ as \textbf{explainability vectors} for the source and target domains, respectively.

From \bref{eq:proposed:analogy}, if $\btau^{\rm S}_{\star} = \btau^{\rm T}_{\star}$ holds, $\btau^{\rm T}_{\rm ft\star}$ can be expressed as
\begin{equation}
\btau^{\rm T}_{\rm ft\star}
= \underbrace{\btheta^{\rm T}_{\rm ft} - \btheta_{\rm base}}_{\btau^{\rm T}_{\rm ft}}
+ \underbrace{\btheta^{\rm S}_{\rm ft\star} - \btheta_{\rm base}}_{\btau^{\rm S}_{\rm ft\star}}
- \underbrace{(\btheta^{\rm S}_{\rm ft} - \btheta_{\rm base})}_{\btau^{\rm S}_{\rm ft}}
= \underbrace{\btheta^{\rm T}_{\rm ft} - \btheta_{\rm base}}_{\btau^{\rm T}_{\rm ft}}
+ \underbrace{\btheta^{\rm S}_{\rm ft\star} - \btheta^{\rm S}_{\rm ft}}_{\btau^{\rm S}_{\star}}.
\label{eq:proposed:tau_ftstar}
\end{equation}
Then, we can obtain the model parameters $\tilde{\btheta}^{\rm T}_{\rm ft\star}$ by adding the task vector $\btau^{\rm T}_{\rm ft\star}$ to the base parameters $\btheta_{\rm base}$.
Introducing two scaling coefficients, $\lambda_1 \in \mathbb{R}$ for classification capability and $\lambda_2 \in \mathbb{R}$ for explainability, we have
\begin{equation}
\tilde{\btheta}^{\rm T}_{\rm ft\star}
= \btheta_{\rm base} + \lambda_1 \btau^{\rm T}_{\rm ft} + \lambda_2 \btau^{\rm S}_{\star}.
\label{eq:proposed:theta_init}
\end{equation}
Since $\lambda_1$ and $\lambda_2$ are hyperparameters that do not affect model training, they can be tuned efficiently.

\paragraph{Remark on theoretical interpretation}
Appendix~\ref{sec:appendix:theory} interprets $\btau^{\rm S}_{\star}$ as the optimal perturbation under a local quadratic approximation of \bref{eq:proposed:objective} and shows that adding it to $\btheta^{\rm T}_{\rm ft}$ increases the target classification loss in proportion to $(\btau^{\rm S}_{\star})^{\top} \bH^{\rm T} \btau^{\rm S}_{\star}$, where the Hessian $\bH^{\rm T}$ needs no explanation supervision.

\subsection{Predicting Transfer Success}\label{sec:proposed:predict}

Task arithmetic provides a simple and efficient method for transferring explainability, but it does not guarantee that the transfer will always be successful because the analogy in \bref{eq:proposed:analogy} may not hold for all source--target pairs, and its benefit varies considerably across source--target pairs as exemplified later in our experiments.
This raises a natural question: can the resulting explanation-quality improvement be anticipated from quantities computable \emph{before} obtaining any target-side explanation supervision?
We formalize this question and examine whether a regression on model-internal features can recover the expected improvement of the transferred model.

\paragraph{Feature design}
We use six features grouped into two classes.
The regression must not require any target-side explanation supervision, because with such supervision one could train $\btheta^{\rm T}_{\rm ft\star}$ directly.
All features are therefore derived from $\btheta^{\rm S}_{\rm ft\star}, \btheta^{\rm S}_{\rm ft}, \btheta^{\rm T}_{\rm ft}, \btau^{\rm S}_{\star}$, and target images with their class labels.
\textit{(i) Classification-circuit features} (3 features) compare the classification circuit of $\btheta^{\rm T}_{\rm ft}$ with the transferred model and the source explanation circuit, motivated by the classification/explanation head-level overlap established in Appendix~\ref{sec:appendix:circuit_overlap}.
They compare the per-head classification effect vectors of $\tilde{\btheta}^{\rm T}_{\rm ft\star}$ and $\btheta^{\rm T}_{\rm ft}$ via cosine similarity and top-$k$ Jaccard overlap, plus the cosine between the source explanation effects of $\btheta^{\rm S}_{\rm ft\star}$ and the target classification effects of $\btheta^{\rm T}_{\rm ft}$.
\textit{(ii) $\btau$-output-shift features} (3 features) probe how strongly $\btau^{\rm S}_{\star}$ acts on target image content, on the hypothesis that a $\btau$ that barely perturbs target image features cannot transfer explainability there.
They summarize how the $\mathtt{[CLS]}$ feature of a frozen zero-shot CLIP backbone shifts on unlabeled target images under the perturbation $\btheta_{\rm base} \mapsto \btheta_{\rm base} + \btau^{\rm S}_{\star}$, via the mean and standard deviation of the per-image $\ell_2$ shift across images and the cosine of the image-mean (centroid) feature pair.
Formal definitions for all six features are in Appendix~\ref{sec:appendix:predict}.

\paragraph{Regression model}
Given $N$ source--target pairs available at training time, we form a feature matrix $\bX \in \mathbb{R}^{N \times 6}$ (with each column standardized to zero mean and unit variance) and a target vector $\by \in \mathbb{R}^{N}$ encoding the explanation-quality improvement of $\tilde{\btheta}^{\rm T}_{\rm ft\star}$ over $\btheta^{\rm T}_{\rm ft}$ on each pair, then fit a linear regression $\hat{\by} = \bX \hat{\bbeta}$ with regularization strength chosen by inner leave-one-target-out cross-validation.
At test time, the regression produces $\hat{y}$ for a held-out target without ever touching any target-side explanation supervision.
\space%

\section{Experiments}\label{sec:experiment}

We evaluate the proposed explainability transfer in two parts: Section~\ref{sec:experiment:imagenet} measures whether a single explainability vector learned on a large-scale source (ImageNet+X) transfers to 10 diverse target datasets, and Section~\ref{sec:experiment:predict} empirically tests the predictability examined in Section~\ref{sec:proposed:predict} with ImageNet+X as the source and 10 targets.
Complementary results are deferred to the appendix.
They include a backbone-scaling study on CLIP with ViT-L/14 (Appendix~\ref{sec:appendix:vitl}) and an all-pairs transferability evaluation across eight smaller-source datasets (Appendix~\ref{sec:appendix:pair}).

\paragraph{Common setup}
We use CLIP with ViT-B/32~\cite{radford2021learning,dosovitskiy2021an} pretrained by OpenAI as the self-explaining model, whose parameters serve as $\btheta_{\rm base}$.
The model input is a $224 \times 224$ RGB image, each patch is $32 \times 32$, and the number of patches is $M = 49$.
We fix the hyperparameter $\lambda_1 = 1.0$ to focus on the effect of explainability transfer via $\btau^{\rm S}_{\star}$ while preserving classification performance on the target datasets, and use $\alpha = 0.8$ and $\lambda_2 = 0.7$ unless otherwise specified. These values are a representative operating point on the accuracy--explanation trade-off rather than optimal values (Appendix~\ref{sec:appendix:lambda2} discusses selecting $\lambda_2$ without target-side explanation supervision).
We sweep $\alpha \in \{0.1, 0.2, \ldots, 0.9\}$ and $\lambda_2 \in \{0.4, 0.5, \ldots, 1.2\}$ in the sensitivity analysis (Figure~\ref{fig:experiment:imagenet:scaling_coef}).
The computing environments are described in Appendix~\ref{sec:appendix:environment}.

\paragraph{Evaluation metrics}
For classification quality, we report top-1 accuracy.
For explanation quality, we report E-RMSE between predicted and GT patch-level attributions, Kendall rank correlation~\cite{kendall1990rank,Jin2023-dk} (\textit{Rank corr.} for short) measuring the similarity between the rankings of predicted and GT attributions, and IoU@$k$ ($k \in \{1,5,10\}$), a top-$k$ variant of the IoU in~\cite{zhang2025saliency}, measuring the overlap between the indices of the top-$k$ predicted and GT patch-level attributions, which evaluates the correctness of highlighted patches.
These GT-agreement metrics directly evaluate our objective of reproducing a user-specified explanation on the target.
Whether the highlighted patches also matter for the transferred model's own predictions is evaluated separately with within-model faithfulness metrics in Section~\ref{sec:experiment:imagenet:faithfulness}.
\space%
\subsection{Transferring Explainability Learned on ImageNet}\label{sec:experiment:imagenet}

\paragraph{Setup}
We use ImageNet-1k~\cite{deng2009imagenet} as the source dataset, since its scale and class diversity make it a natural candidate for capturing transferable patch-level explanations.
We construct the ImageNet+X dataset, in which each sample consists of a labeled image from the original ImageNet-1k and a GT patch-level attribution.
The GT attributions are generated using Kernel SHAP~\cite{Lundberg2017-ii,noauthor_undated-ca} with $500$ perturbations per image, applied to the CLIP image classifier finetuned on ImageNet-1k. Similar approaches for generating explanation supervision have been used in previous studies~\cite{yang2023efficient,monteiro2024selective}.
As the target datasets, we use 10 image classification datasets: Cars~\cite{krause20133d}, DTD~\cite{cimpoi2014describing}, EuroSAT~\cite{helber2019eurosat}, GTSRB~\cite{stallkamp2011german}, RESISC45~\cite{cheng2017remote}, SUN397~\cite{xiao2016sun}, SVHN~\cite{netzer2011reading}, MNIST~\cite{lecun1998mnist}, CIFAR100~\cite{Krizhevsky2009LearningML}, and STL10~\cite{coates2011analysis}.
As with ImageNet+X, we add the GT patch-level attributions to the 10 target datasets for evaluation, following the same Kernel SHAP procedure with the per-target finetuned classifier.

\begingroup
\renewcommand{\arraystretch}{1.1} %
\begin{table}[t]
\caption{
Average performances and their standard deviations of ViT-B/32-based self-explaining models over the 10 target datasets when transferring explainability learned on ImageNet+X, compared with those of baselines.
The last row shows the oracle performances achieved by finetuning on the target datasets with explanation supervision (target+X).
\textbf{\underline{Bold and underlined values}} indicate the best among the proposed method and the baselines for each metric, while \textbf{bold values} indicate that there is no statistically significant difference from the best, as determined by a paired $t$-test at a significance level of 0.05.
}
\label{tab:experiment:imagenet}
\centering
\resizebox{1.0\textwidth}{!}{%
\begin{NiceTabular}{@{}rlcccccc@{}}
\toprule
 & Param. & Accuracy ($\uparrow$) & E-RMSE ($\downarrow$) & Rank corr. ($\uparrow$) & IoU@1 ($\uparrow$) & IoU@5 ($\uparrow$) & IoU@10 ($\uparrow$) \\ \midrule
Zero-shot & $\btheta_{\rm base}$ & $0.475 \pm 0.211$ & $1.508 \pm 0.069$ & $-0.095 \pm 0.059$ & $0.012 \pm 0.006$ & $0.040 \pm 0.013$ & $0.090 \pm 0.021$ \\
Finetuned on ImageNet & $\btheta^{\rm S}_{\rm ft}$ & $0.379 \pm 0.239$ & $1.528 \pm 0.081$ & $-0.128 \pm 0.076$ & $0.009 \pm 0.006$ & $0.031 \pm 0.016$ & $0.074 \pm 0.026$ \\
Finetuned on ImageNet+X & $\btheta^{\rm S}_{\rm ft\star}$ & $0.353 \pm 0.243$ & $\bf 1.149 \pm 0.092$ & $\bf 0.197 \pm 0.063$ & $\bf 0.106 \pm 0.059$ & $\bf 0.187 \pm 0.082$ & $\bf 0.261 \pm 0.079$ \\
Finetuned on target & $\btheta^{\rm T}_{\rm ft}$ & \underline{$\bf 0.890 \pm 0.106$} & $1.562 \pm 0.084$ & $-0.145 \pm 0.068$ & $0.009 \pm 0.005$ & $0.033 \pm 0.013$ & $0.076 \pm 0.021$ \\
Proposed & $\tilde{\btheta}^{\rm T}_{\rm ft\star}$ & $\bf 0.886 \pm 0.113$ & \underline{$\bf 1.125 \pm 0.116$} & \underline{$\bf 0.200 \pm 0.061$} & \underline{$\bf 0.133 \pm 0.092$} & \underline{$\bf 0.216 \pm 0.114$} & \underline{$\bf 0.279 \pm 0.093$} \\ \dotrule[1.5pt]
Finetuned on target+X & $\btheta^{\rm T}_{\rm ft\star}$ & $0.909 \pm 0.100$ & $1.028 \pm 0.224$ & $0.242 \pm 0.104$ & $0.248 \pm 0.217$ & $0.291 \pm 0.189$ & $0.328 \pm 0.140$ \\ \bottomrule
\end{NiceTabular}%
}
\end{table}
\endgroup

\subsubsection{Quantitative Results}
Table~\ref{tab:experiment:imagenet} shows the average performances of the explainability transferred by the proposed method, in comparison to the baseline models with the parameters $\btheta_{\rm base}$, $\btheta^{\rm S}_{\rm ft}$, $\btheta^{\rm S}_{\rm ft\star}$, and $\btheta^{\rm T}_{\rm ft}$.
For reference, we also show the performances of the model finetuned on the target dataset with explanation supervision, i.e., the oracle $\btheta^{\rm T}_{\rm ft\star}$.
The table shows that compared to the baselines, the proposed method achieves the second-best accuracy while consistently achieving the best explainability performance.
The accuracy gap to the oracle $\btheta^{\rm T}_{\rm ft\star}$ is only $0.023$ ($\approx 2.5\%$ relative), and the E-RMSE gap is $\approx 9\%$ relative. The proposed method thus transfers explainability acquired on ImageNet+X while largely preserving classification accuracy.
The most accurate baseline, $\btheta^{\rm T}_{\rm ft}$, has the worst or second-worst explainability, while the second-best explainability, that of $\btheta^{\rm S}_{\rm ft\star}$, comes with the lowest accuracy.
The proposed method combines the high target-domain accuracy of $\btheta^{\rm T}_{\rm ft}$ with the explainability vector extracted from $\btheta^{\rm S}_{\rm ft\star}$, attaining classification accuracy on par with $\btheta^{\rm T}_{\rm ft}$ and explanation quality close to the oracle $\btheta^{\rm T}_{\rm ft\star}$.

\begin{figure}[t]
\centering
\includegraphics[width=\textwidth]{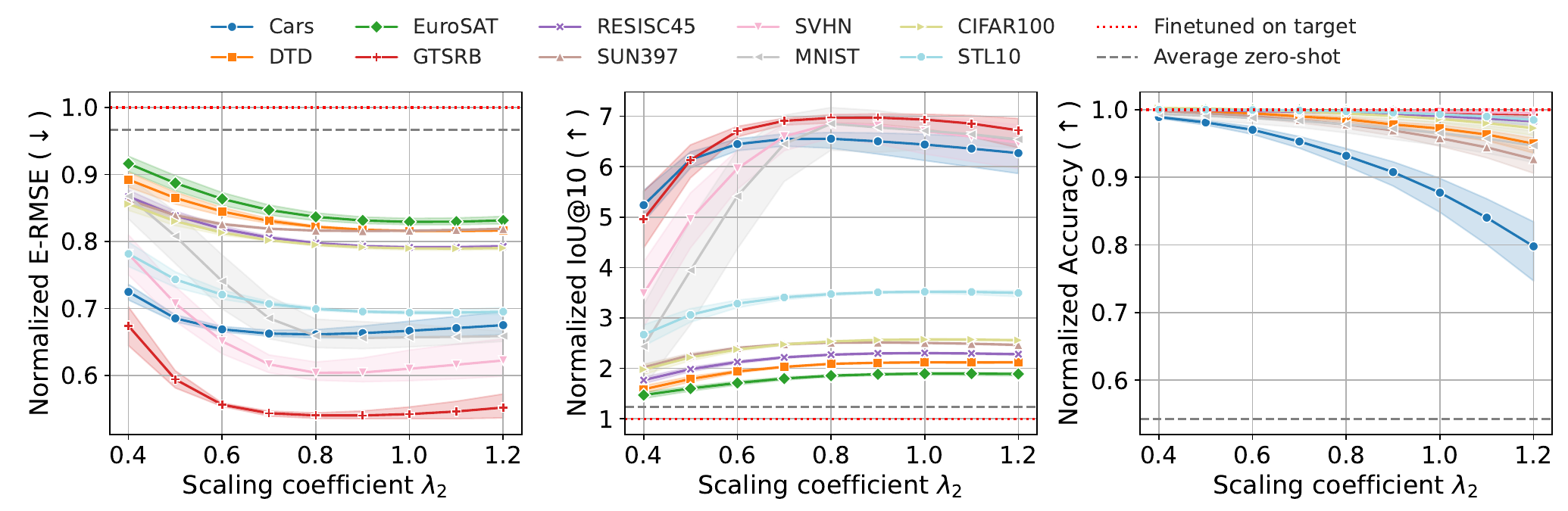}\\
\vspace{-2mm}
\footnotesize
\begin{minipage}[t]{0.3333\textwidth}\centering (a)\end{minipage}%
\begin{minipage}[t]{0.3333\textwidth}\centering (b)\end{minipage}%
\begin{minipage}[t]{0.3333\textwidth}\centering (c)\end{minipage}%
\normalsize
\vspace{-1.5mm}
\caption{Normalized (a) E-RMSE, (b) IoU@10, and (c) accuracy for each target dataset over the scaling coefficient for explainability $\lambda_2$.
Lines show the mean over the nine models trained with $\alpha \in \{0.1, 0.2, \ldots, 0.9\}$.
Shaded areas show a 95\% bootstrap interval of that mean, obtained by resampling the nine $\alpha$ values.}
\label{fig:experiment:imagenet:scaling_coef}
\end{figure}

\subsubsection{Impact of Scaling Coefficient}

We investigate the impact of the scaling coefficient for explainability $\lambda_2$ for each target dataset.
Figure~\ref{fig:experiment:imagenet:scaling_coef} shows the normalized E-RMSE, IoU@10, and accuracy for each target dataset over $\lambda_2 \in \{0.4,0.5,\ldots,1.2\}$.
The scores are normalized by dividing them by the corresponding scores of $\btheta^{\rm T}_{\rm ft}$.
Although trends vary by dataset, the explainability scores in Figures~\ref{fig:experiment:imagenet:scaling_coef}(a) and (b) improve for all 10 target datasets, with improvements plateauing around $\lambda_2 \approx 0.8$.
Figure~\ref{fig:experiment:imagenet:scaling_coef}(c) shows that classification accuracy tends to decrease as $\lambda_2$ increases, but the degradation is kept within 5\% of $\btheta^{\rm T}_{\rm ft}$ when $\lambda_2 \leq 1.0$, except for Cars.

The theoretical analysis in Appendix~\ref{sec:appendix:theory} shows that such a drop can occur independently of the gain in explanation quality.
For Cars the drop is noticeable, but the explanation performance is already high at $\lambda_2 \leq 0.6$, so an appropriate $\lambda_2$ balances the two.
The fixed setting $\lambda_2 = 0.7$ used uniformly across the 10 targets in Table~\ref{tab:experiment:imagenet} already attains substantial improvements without any target-side explanation supervision.
$\lambda_2$ can also be selected without target explanation labels.
Among the values whose mean accuracy on the other targets stays within a tolerance of the best, we take the one with the best mean explanation metric on those targets.
With a tolerance of $0.5$ points, this rule selects $0.7 \pm 0.1$ for 9 of 10 held-out targets (Appendix~\ref{sec:appendix:lambda2}).

\begin{figure}[t]
\centering
\includegraphics[width=\textwidth]{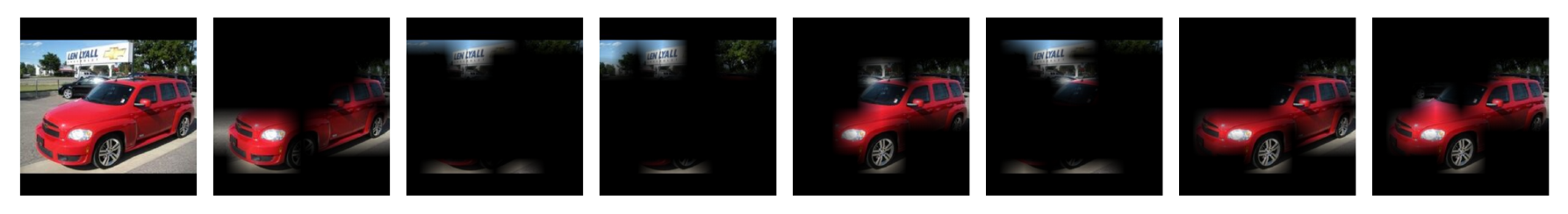}\\
\vspace{-0.2cm}
\includegraphics[width=\textwidth]{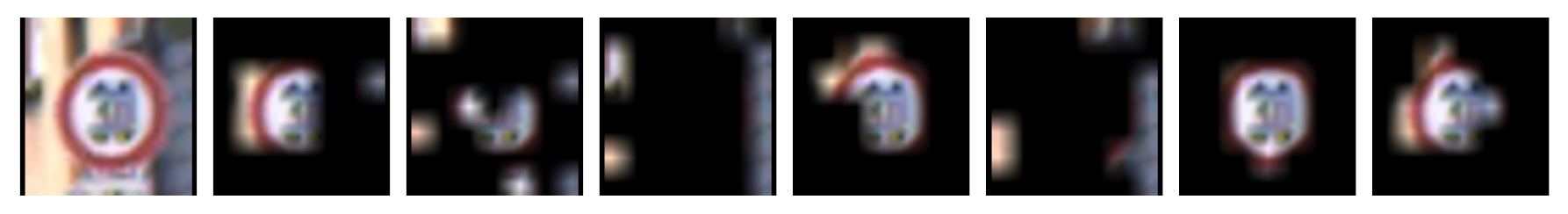}\\
\vspace{-0.2cm}
\includegraphics[width=\textwidth]{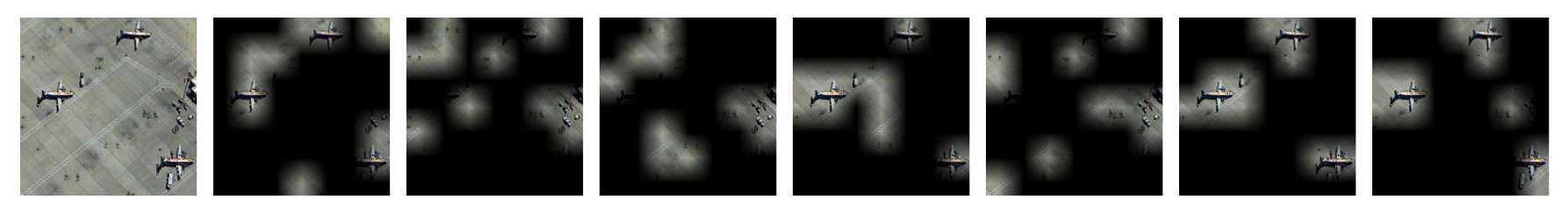}\\
\vspace{-2mm}
\noindent\scriptsize
\begin{minipage}[t]{0.125\textwidth}\centering (a) Input\end{minipage}%
\begin{minipage}[t]{0.125\textwidth}\centering (b) Ground-truth\end{minipage}%
\begin{minipage}[t]{0.125\textwidth}\centering (c) Zero-shot\end{minipage}%
\begin{minipage}[t]{0.125\textwidth}\centering (d) Finetuned on ImageNet\end{minipage}%
\begin{minipage}[t]{0.125\textwidth}\centering (e) Finetuned on ImageNet+X\end{minipage}%
\begin{minipage}[t]{0.125\textwidth}\centering (f) Finetuned on target\end{minipage}%
\begin{minipage}[t]{0.125\textwidth}\centering (g) Proposed\end{minipage}%
\begin{minipage}[t]{0.125\textwidth}\centering (h) Finetuned on target+X\end{minipage}%
\normalsize
\caption{Visualization examples highlighting the top-10 patches in the predicted patch-level attributions.
Rows show (from top to bottom) the results for Cars, GTSRB, and RESISC45 datasets.}
\label{fig:experiment:imagenet:example}
\end{figure}

\subsubsection{Qualitative Evaluation}

To evaluate the predicted patch-level attributions qualitatively, we visualize the top-10 patches in the attributions produced by the proposed method and the other methods in Figure~\ref{fig:experiment:imagenet:example}, which corresponds to the IoU@10 evaluation.
The patches highlighted by the proposed method (Figure~\ref{fig:experiment:imagenet:example}(g)) are more similar to the GT ones (Figure~\ref{fig:experiment:imagenet:example}(b)) than those highlighted by the three baseline methods that do not use any explanation supervision (Figures~\ref{fig:experiment:imagenet:example}(c), (d), and (f)), and visually approach the target+X oracle (Figure~\ref{fig:experiment:imagenet:example}(h)).
The model finetuned on ImageNet+X (Figure~\ref{fig:experiment:imagenet:example}(e)) also generates high-quality explanations similar to the proposed method, consistent with the quantitative evaluation. However, due to its low classification performance, it is not practical to use this single model for both classification and explanation simultaneously.
Additional randomly selected examples for each of the 10 target datasets are provided in Appendix~\ref{sec:appendix:additional_examples}.

\subsubsection{Comparison with Kernel SHAP}\label{sec:experiment:imagenet:vsshap}
\begin{wrapfigure}[18]{r}{0.35\columnwidth}
\vspace{-1.0em}
\centering
\includegraphics[width=0.35\columnwidth]{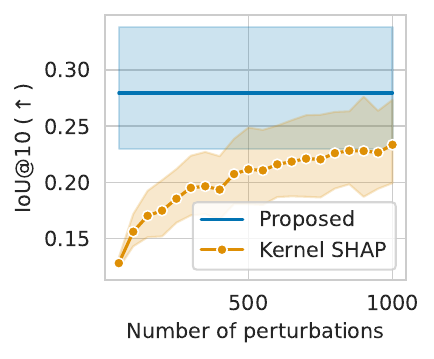}
\vspace{-1.5em}
\caption{IoU@10 of the proposed method and Kernel SHAP applied to $\btheta^{\rm T}_{\rm ft}$ across the 10 target datasets, with varying number of perturbations $P$ for SHAP. The shaded area represents the 95\% bootstrap confidence interval. The full five-metric grid is in Figure~\ref{fig:appendix:imagenet:vsshap} in Appendix~\ref{sec:appendix:vs_shap}.}
\label{fig:experiment:imagenet:vsshap}
\end{wrapfigure}
The GT patch-level attributions for ViT-B/32 are generated using Kernel SHAP applied to $\btheta^{\rm T}_{\rm ft}$ with $P = 500$ perturbations (SHAP-500) per image, which produces high-quality explanations but is computationally expensive (computational complexity is detailed in Appendix~\ref{sec:appendix:complexity}).
Once $\tilde{\btheta}^{\rm T}_{\rm ft\star}$ is obtained, the proposed method instead produces patch-level attributions in a single forward pass.
To quantify this efficiency advantage, we compute Kernel SHAP attributions on $\btheta^{\rm T}_{\rm ft}$ with varying $P$ (using independent perturbation samples from those used for GT) and evaluate them against the GT alongside the proposed method's deterministic attributions.

Figure~\ref{fig:experiment:imagenet:vsshap} reports the average IoU@10 across the 10 target datasets, complementing the top-$10$ visualization in the previous subsection (the full five-metric grid is in Appendix~\ref{sec:appendix:vs_shap}).
The proposed method attains IoU@10 $= 0.279$ against the GT, whereas Kernel SHAP on $\btheta^{\rm T}_{\rm ft}$ reaches only $0.233$ even at $P = 1{,}000$ perturbations.
This gap reflects the sampling variance of Kernel SHAP at the GT budget.
An independent SHAP-500 run on the same model attains IoU@10 $= 0.211$ against the GT, so any single SHAP-500 realization is a noisy estimator.
The learned attributions may average out this noise, at the cost of a single forward pass.

\subsubsection{Faithfulness, Specificity, and Comparison with External Explainers}\label{sec:experiment:imagenet:faithfulness}

GT agreement does not show whether the highlighted patches matter for the transferred model's own predictions, whether the effect is specific to $\btau^{\rm S}_{\star}$, or how the method compares with other ways of explaining a finetuned classifier.
We address these questions with the pipeline of Table~\ref{tab:experiment:imagenet} on a 10-target set in which CIFAR10 replaces SUN397.
On this set, the mean accuracy is $0.914$ for $\btheta^{\rm T}_{\rm ft}$ and $0.911$ for the proposed method.

\paragraph{Within-model faithfulness}
We test whether the patches highlighted by the transferred model $\tilde{\btheta}^{\rm T}_{\rm ft\star}$ actually drive its own predictions.
We mask the patches in order of attributed importance and measure how the predicted-class probability of $\tilde{\btheta}^{\rm T}_{\rm ft\star}$ changes on the correctly classified images among $500$ class-stratified test images per target.
We quantify this change with four metrics: Fidelity$+$ and Fidelity$-$~\cite{wang2024tve}, and deletion and insertion scores~\cite{petsiuk2018rise}.
We compare three attributions scored on the same model.
They are the self-explanation of $\tilde{\btheta}^{\rm T}_{\rm ft\star}$, Kernel SHAP applied to $\tilde{\btheta}^{\rm T}_{\rm ft\star}$ itself (number of perturbations $P = 500$) as a model-specific reference, and a random attribution.
The self-explanation matches or exceeds the model-specific SHAP reference on $36$ of the $40$ comparisons ($10$ targets $\times$ $4$ metrics) and beats random on all $40$ (Appendix~\ref{sec:appendix:faithfulness}).
The highlighted patches are thus the ones that move the model's own predictions.

\begin{wraptable}[13]{r}{0.44\textwidth}
\vspace{-4mm}
\caption{Accuracy and explanation quality when each vector is added in place of $\btau^{\rm S}_{\star}$ in~\bref{eq:proposed:theta_init}, averaged over the 10 targets. Per-target results are in Appendix~\ref{sec:appendix:controls}.}
\label{tab:experiment:controls}
\centering
\footnotesize
\setlength{\tabcolsep}{3pt}
\begin{tabular}{@{}lccc@{}}
\toprule
Added vector & Acc. & Rank corr. & IoU@1 \\
\midrule
Proposed & $0.911$ & $0.204$ & $0.132$ \\
\midrule
No vector & $0.914$ & $-0.147$ & $0.009$ \\
Equal-norm random & $0.909$ & $-0.144$ & $0.010$ \\
Front-6 blocks & $0.912$ & $-0.147$ & $0.009$ \\
Back-6 blocks & $0.913$ & $0.148$ & $0.109$ \\
Shuffled supervision & $0.909$ & $0.190$ & $0.079$ \\
\bottomrule
\end{tabular}
\vspace{-3mm}
\end{wraptable}
\paragraph{Specificity of the effect}
We test whether the explanation improvement is caused by $\btau^{\rm S}_{\star}$ itself rather than by adding any vector to the model.
To this end, we replace $\btau^{\rm S}_{\star}$ in~\bref{eq:proposed:theta_init} by control vectors, keep the rest of the pipeline unchanged, and compare the resulting models (Table~\ref{tab:experiment:controls}).
Adding an equal-norm random direction or the vector restricted to the first six transformer blocks gives almost the same results as adding no vector.
In contrast, the last six blocks alone recover $84\%$ of the Rank corr.\ improvement and $82\%$ of the IoU@1 improvement of $\btau^{\rm S}_{\star}$ over no vector.
These results show that the effect depends on the direction of the vector and comes mostly from the later blocks (Appendices~\ref{sec:appendix:controls} and~\ref{sec:appendix:circuit_overlap:mechanism}).
A vector retrained with explanation maps shuffled across source images recovers $96\%$ of the Rank corr.\ improvement but only $57\%$ of the IoU@1 improvement (Appendix~\ref{sec:appendix:controls}).
This suggests that the vector carries both an attribution prior that does not depend on the content of the supervision and a component that does.

\paragraph{Comparison with external explainers}
We also compare the proposed method with two external explainers: a separate model trained with the explanation loss alone ($\alpha = 0$ in~\bref{eq:proposed:objective}) that attributes the predictions of $\btheta^{\rm T}_{\rm ft}$, and TVE~\cite{wang2024tve}, which in our evaluation with its public checkpoint explains a linear probe on its own frozen encoder (Appendix~\ref{sec:appendix:alternatives}).
Both produce non-random explanations for the classifier they accompany.
Unlike them, the proposed method produces the explanation within the finetuned target itself, in one forward pass with no extra parameters.
TVE's probe is $13.5$ points less accurate than $\btheta^{\rm T}_{\rm ft}$ on this set.
When the deployed model must not be modified, post-hoc explainers such as Kernel SHAP remain the appropriate choice (Appendix~\ref{sec:limitations}).
\space%
\subsection{Empirical Test of Transfer-Success Prediction}\label{sec:experiment:predict}

We empirically test the predictability examined in Section~\ref{sec:proposed:predict} with ImageNet+X as the source and the 10 targets of Section~\ref{sec:experiment:imagenet}, with CIFAR10~\cite{Krizhevsky2009LearningML} in place of CIFAR100.
Let $q(\cdot)$ denote a function that maps a model's parameters to a scalar explanation-quality score evaluated on $\set{D}^{\rm T}$, instantiated with $q \in \{\text{E-MSE}, \text{IoU@1}, \text{IoU@5}, \text{IoU@10}, \text{Rank corr.}\}$.
E-MSE is the squared E-RMSE, $\set{L}_{\rm exp}$ in~\bref{eq:proposed:exp_loss}.
All scores use 512 fixed random test images per target.
For each metric, let $b_q = q(\btheta^{\rm T}_{\rm ft})$ denote the baseline value before explainability transfer and $a_q = q(\tilde{\btheta}^{\rm T}_{\rm ft\star})$ the value after transfer.
For source dataset $\set{D}^{\rm S}$ and target dataset $\set{D}^{\rm T}$, we define the predictor target as the (sign-aligned) raw improvement in explanation quality: $\Delta_q(S, T) = s_q \cdot (a_q - b_q)$,
where $s_q \in \{+1, -1\}$ aligns the sign so that larger $\Delta_q$ indicates better transfer ($s_q = -1$ for E-MSE and $s_q = +1$ otherwise).
We obtain $N = 10$ source--target pairs (one per target) and perform leave-one-target-out (LOTO) evaluation: each fold withholds one target, fits Lasso or Ridge regression on the remaining nine pairs with inner-LOTO regularization selection, and predicts $\Delta_q$ for the held-out target.
For each metric, we use the better of Lasso and Ridge and report the held-out Pearson $r$ between predicted and actual $\Delta_q$ over the 10 targets.
A one-sided $500$-permutation test reruns all steps, including this choice.

\begin{wraptable}{r}{0.45\textwidth}
\vspace{-1.0em}
\centering
\caption{Held-out Pearson correlation (LOTO, $N = 10$) between predicted and actual $\Delta_q$ on ViT-B/32 for three representative metrics. The full five-metric ViT-B/32 grid and ViT-L/14 results are in Appendix~\ref{sec:appendix:predict}.
Best per metric is in \textbf{bold}.}
\label{tab:experiment:predict}
\small
\begin{tabular}{@{}lccc@{}}
\toprule
Method & E-MSE & IoU@10 & Rank corr. \\
\midrule
Lasso & $\mathbf{+0.92}$ & $+0.81$ & $+0.77$ \\
Ridge & $+0.91$ & $\mathbf{+0.85}$ & $\mathbf{+0.82}$ \\
\bottomrule
\end{tabular}
\vspace{-1.0em}
\end{wraptable}

\paragraph{Result}
Table~\ref{tab:experiment:predict} shows that a regression on the six candidate features strongly correlates with the actual transfer outcome on the three reported metrics.
On E-MSE, Lasso attains $r = +0.92$ across the 10 held-out targets (uncorrected permutation $p = 0.006$), with a bootstrap 95\% interval of $[+0.84, +0.98]$.
The intervals on Rank corr.\ are wider, at $[+0.56, +0.92]$ for Lasso and $[+0.70, +0.94]$ for Ridge (Appendix~\ref{sec:appendix:pair:predictor}).
Over all five metrics (full grid in Appendix~\ref{sec:appendix:predict}), the regression reaches $r \geq +0.70$ on four, namely E-MSE, Rank corr., IoU@5, and IoU@10 (uncorrected permutation $p \leq 0.022$).
The remaining metric, IoU@1, is comparatively weak ($r = +0.33$ at best).
We revisit this metric- and backbone-specific unevenness in Appendix~\ref{sec:limitations}.
A held-out evaluation of the same predictor on 56 cross-domain pairs is provided in Appendix~\ref{sec:appendix:pair}.
The claim is restricted to a fixed source with held-out targets.
Under leave-one-source-out evaluation on the 56 pairs, none of 40 predictor configurations meets our criteria (Appendix~\ref{sec:appendix:pair:predictor}).

\section{Conclusion}
We proposed a task-arithmetic method to transfer visual explainability from self-explaining to prediction-only VLM-based ViTs across domains, without target-side explanation supervision.
A single explainability vector learned on ImageNet+X improved explanation quality on 10 diverse target datasets while largely preserving accuracy.
The transferred explanations are faithful to the model's own predictions, and comparisons with control vectors (e.g., random vectors of the same norm) show that the effect is specific to the vector's direction, largely mediated by the later blocks, and partly dependent on the content of the source supervision.
For a fixed source, transfer success on held-out targets is predictable to a meaningful degree from a small set of model-internal features alone.
Limitations and broader impacts are in Appendices~\ref{sec:limitations} and~\ref{sec:appendix:impact_statement}.

\begin{ack}
This work was supported by JSPS KAKENHI Grant Number JP26K14968 and by a joint research fund from ZOZO Research.
The authors declare no other competing interests.
\end{ack}

\bibliographystyle{plainnat}
\bibliography{bibref.bib}

\newpage
\appendix

\section*{Appendix Contents}
\begingroup
\renewcommand{\arraystretch}{1.15}
\noindent\begin{tabular}{@{}p{1.2em}p{0.92\columnwidth}@{}}
\ref{sec:appendix:shared_vs_separated}. & \hyperref[sec:appendix:shared_vs_separated]{Ablation Study: Shared Head vs.\ Separated Head} \\
\ref{sec:appendix:circuit_overlap}. & \hyperref[sec:appendix:circuit_overlap]{Circuit Overlap between Classification and Explanation} \\
   & \quad \ref{sec:appendix:circuit_overlap:rq1} \hyperref[sec:appendix:circuit_overlap:rq1]{Classification and Explanation Heads Largely Overlap} \\
   & \quad \ref{sec:appendix:circuit_overlap:rq2} \hyperref[sec:appendix:circuit_overlap:rq2]{Circuit Overlap Is Present before Explanation Training} \\
   & \quad \ref{sec:appendix:circuit_overlap:mechanism} \hyperref[sec:appendix:circuit_overlap:mechanism]{Where the Transferred Effect Is Mediated} \\
\ref{sec:appendix:predict}. & \hyperref[sec:appendix:predict]{Feature Pool and ViT-L/14 Predictability Analysis} \\
\ref{sec:appendix:environment}. & \hyperref[sec:appendix:environment]{Computing Environment for Experiments} \\
\ref{sec:appendix:vitl}. & \hyperref[sec:appendix:vitl]{Experiments with ViT-L/14} \\
   & \quad \ref{sec:appendix:vitl:setup} \hyperref[sec:appendix:vitl:setup]{Setup} \\
   & \quad \ref{sec:appendix:vitl:quantitative} \hyperref[sec:appendix:vitl:quantitative]{Quantitative Results} \\
   & \quad \ref{sec:appendix:vitl:qualitative} \hyperref[sec:appendix:vitl:qualitative]{Qualitative Evaluation} \\
\ref{sec:appendix:pair}. & \hyperref[sec:appendix:pair]{Transferring Explainability Across Source and Target Datasets} \\
   & \quad \ref{sec:appendix:pair:tradeoff} \hyperref[sec:appendix:pair:tradeoff]{Accuracy and Explanation Error Trade-off} \\
   & \quad \ref{sec:appendix:pair:predictor} \hyperref[sec:appendix:pair:predictor]{Predictability Analysis on 56 Pairs} \\
\ref{sec:appendix:theory}. & \hyperref[sec:appendix:theory]{Theoretical Interpretation} \\
\ref{sec:appendix:vs_shap}. & \hyperref[sec:appendix:vs_shap]{Results on Comparison with Kernel SHAP} \\
\ref{sec:appendix:complexity}. & \hyperref[sec:appendix:complexity]{Discussion on Computational Complexity of Explanation Generation} \\
\ref{sec:appendix:faithfulness}. & \hyperref[sec:appendix:faithfulness]{Faithfulness of the Transferred Self-Explanations} \\
\ref{sec:appendix:controls}. & \hyperref[sec:appendix:controls]{Control-Vector Ablations} \\
\ref{sec:appendix:alternatives}. & \hyperref[sec:appendix:alternatives]{Comparison with Amortized and Transferable Explainers} \\
   & \quad \ref{sec:appendix:alternatives:amortized} \hyperref[sec:appendix:alternatives:amortized]{Amortized Explainer with the Same Supervision} \\
   & \quad \ref{sec:appendix:alternatives:tve} \hyperref[sec:appendix:alternatives:tve]{Transferable Vision Explainer} \\
\ref{sec:appendix:lambda2}. & \hyperref[sec:appendix:lambda2]{Selecting $\lambda_2$ without Target-Side Explanation Supervision} \\
\ref{sec:appendix:stress}. & \hyperref[sec:appendix:stress]{Task-Dependent Explanations on Two-Task Images} \\
\ref{sec:limitations}. & \hyperref[sec:limitations]{Limitations} \\
\ref{sec:appendix:impact_statement}. & \hyperref[sec:appendix:impact_statement]{Impact Statement} \\
\ref{sec:appendix:additional_examples}. & \hyperref[sec:appendix:additional_examples]{Additional Visualization Examples} \\
\end{tabular}
\endgroup
\medskip
\clearpage

\section{Ablation Study: Shared Head vs.\ Separated Head}\label{sec:appendix:shared_vs_separated}

To investigate whether the shared-head design of our self-explaining model harms classification and explanation quality, we compare the proposed shared-head architecture with a separated-head variant.
In the shared-head model (proposed), the domain-specific head $g_{\bW}$ is shared between classification (mapping the $\mathtt{[CLS]}$ token to logits) and explanation (mapping patch tokens to attributions), and is kept frozen from the pretrained model.
In the separated-head model, a dedicated explanation head $g_{\bW_{\rm exp}}$ is introduced separately from the classification head $g_{\bW}$, where both heads are initialized from the pretrained $g_{\bW}$. The entire model, including the feature extractor and both heads, is then trained jointly.

Table~\ref{tab:appendix:shared_vs_separated} shows the results averaged across the 10 target datasets when finetuning on each target dataset with explanation supervision (target+X).
At $\alpha = 0.8$ (used as the default in the main paper), the shared-head model attains a lower E-RMSE and higher Rank corr., IoU@5, and IoU@10 than the separated-head model at every $\alpha \in \{0.0, 0.1, \ldots, 1.0\}$, with comparable classification accuracy.
No separated-head setting matches the shared-head model on accuracy and explanation quality at once.
The settings whose accuracy shows no significant difference from the best ($\alpha \geq 0.8$) are significantly worse on E-RMSE, Rank corr., and IoU@10, and the settings that are close on explanation quality ($\alpha \leq 0.5$) are significantly less accurate.
On IoU@1, the separated-head model at $\alpha = 0$ is slightly higher ($0.250$ vs.\ $0.248$), with no significant difference.
The shared-head model attains this without any training of the head, whereas the separated-head model trains $g_{\bW_{\rm exp}}$ with explanation supervision.

\paragraph{Why the gap reflects head trainability, not architecture}
Because $g_{\bW}$ and $g_{\bW_{\rm exp}}$ are initialized to the same pretrained weights and both heads are linear, freezing both heads in the separated-head model makes it mathematically equivalent to the shared-head model.
The two heads then remain weight-identical throughout training, producing the same outputs and the same gradient flow into the feature extractor.
The shared-head row in Table~\ref{tab:appendix:shared_vs_separated} can therefore be read as the architecture-only counterpart of the separated-head condition with both heads frozen, and the gap between this row and the trainable-head separated rows isolates the effect of head trainability rather than the architectural choice of separating the heads.

\begingroup
\renewcommand{\arraystretch}{1.2} %
\begin{table}[t!]
\caption{Comparison of shared-head (proposed) and separated-head architectures when finetuning on each target dataset with explanation supervision (target+X), averaged across the 10 target datasets. The shared-head model keeps $g_{\bW}$ frozen from the pretrained model, while the separated-head model has separate classification and explanation heads, both initialized from the pretrained $g_{\bW}$, and trains the entire model including the feature extractor and both heads.
\textbf{\underline{Bold and underlined values}} indicate the best among all settings for each metric, while \textbf{bold values} indicate that there is no statistically significant difference from the best, as determined by a paired $t$-test at a significance level of 0.05.
}
\label{tab:appendix:shared_vs_separated}
\centering
\resizebox{1.0\textwidth}{!}{%
\begin{tabular}{@{}llrrrrrr@{}}
\toprule
Head & $\alpha$ & Accuracy ($\uparrow$) & E-RMSE ($\downarrow$) & Rank corr.\ ($\uparrow$) & IoU@1 ($\uparrow$) & IoU@5 ($\uparrow$) & IoU@10 ($\uparrow$) \\ \midrule
Separated & 0.0 & $0.474 \pm 0.322$ & $\bf 1.058 \pm 0.242$ & $\bf 0.218 \pm 0.125$ & \underline{$\bf 0.250 \pm 0.227$} & $\bf 0.280 \pm 0.198$ & $\bf 0.309 \pm 0.150$ \\
Separated & 0.1 & $0.886 \pm 0.122$ & $\bf 1.066 \pm 0.244$ & $\bf 0.211 \pm 0.126$ & $\bf 0.246 \pm 0.227$ & $\bf 0.275 \pm 0.199$ & $\bf 0.305 \pm 0.150$ \\
Separated & 0.2 & $0.891 \pm 0.118$ & $\bf 1.067 \pm 0.244$ & $0.211 \pm 0.126$ & $0.243 \pm 0.228$ & $\bf 0.275 \pm 0.199$ & $\bf 0.305 \pm 0.149$ \\
Separated & 0.3 & $0.893 \pm 0.116$ & $\bf 1.068 \pm 0.244$ & $0.210 \pm 0.126$ & $\bf 0.244 \pm 0.226$ & $\bf 0.274 \pm 0.199$ & $\bf 0.305 \pm 0.150$ \\
Separated & 0.4 & $0.896 \pm 0.114$ & $\bf 1.069 \pm 0.243$ & $0.210 \pm 0.125$ & $\bf 0.243 \pm 0.228$ & $\bf 0.274 \pm 0.198$ & $\bf 0.304 \pm 0.149$ \\
Separated & 0.5 & $0.896 \pm 0.112$ & $\bf 1.069 \pm 0.242$ & $0.210 \pm 0.125$ & $\bf 0.243 \pm 0.225$ & $\bf 0.274 \pm 0.198$ & $\bf 0.304 \pm 0.149$ \\
Separated & 0.6 & $0.899 \pm 0.111$ & $1.069 \pm 0.242$ & $0.210 \pm 0.124$ & $\bf 0.241 \pm 0.225$ & $\bf 0.273 \pm 0.198$ & $0.305 \pm 0.149$ \\
Separated & 0.7 & $0.900 \pm 0.110$ & $1.069 \pm 0.241$ & $0.209 \pm 0.123$ & $\bf 0.239 \pm 0.223$ & $\bf 0.273 \pm 0.196$ & $0.304 \pm 0.148$ \\
Separated & 0.8 & $\bf 0.901 \pm 0.110$ & $1.066 \pm 0.243$ & $0.211 \pm 0.122$ & $\bf 0.239 \pm 0.224$ & $\bf 0.273 \pm 0.196$ & $0.306 \pm 0.149$ \\
Separated & 0.9 & $\bf 0.902 \pm 0.109$ & $1.062 \pm 0.239$ & $0.214 \pm 0.118$ & $\bf 0.235 \pm 0.222$ & $0.274 \pm 0.194$ & $0.308 \pm 0.147$ \\
Separated & 1.0 & \underline{$\bf 0.910 \pm 0.099$} & $1.562 \pm 0.080$ & $-0.146 \pm 0.063$ & $0.008 \pm 0.005$ & $0.033 \pm 0.013$ & $0.076 \pm 0.020$ \\
Shared (proposed) & 0.8 & $\bf 0.909 \pm 0.100$ & \underline{$\bf 1.028 \pm 0.224$} & \underline{$\bf 0.242 \pm 0.104$} & $\bf 0.248 \pm 0.217$ & \underline{$\bf 0.291 \pm 0.189$} & \underline{$\bf 0.328 \pm 0.140$} \\ \bottomrule
\end{tabular}
}
\end{table}
\endgroup

\section{Circuit Overlap between Classification and Explanation}\label{sec:appendix:circuit_overlap}

Our self-explaining model in Section~\ref{sec:proposed:model} produces the classification logits and the patch-level attributions in parallel from the backbone features.
A natural concern is whether this parallel-output structure is only a surface-level coupling at the final head, while classification and explanation are actually computed by disjoint internal sub-circuits of the backbone.
In that case, the two capabilities would not interact mechanistically inside the model.
In this appendix, we test this concern by quantifying the head-level overlap of classification and explanation circuits in self-explaining ViT image classifiers, and further examine whether this overlap is induced by the explanation supervision $\set{L}_{\rm exp}$ or is already a property of the architecture before any explanation training.
Appendices~\ref{sec:appendix:circuit_overlap:rq1} and~\ref{sec:appendix:circuit_overlap:rq2} use, on both backbones, the 10 targets of Section~\ref{sec:experiment:imagenet} with CIFAR10 in place of CIFAR100, as in Section~\ref{sec:experiment:predict}.
The target sets of Appendix~\ref{sec:appendix:circuit_overlap:mechanism} are stated there.

\paragraph{Head-level effect}
Let $L$ and $H$ denote the number of transformer layers and the number of attention heads per layer, respectively (e.g., $L = H = 12$ for ViT-B/32 and $L = 24, H = 16$ for ViT-L/14).
We follow the standard multi-head attention notation: within each transformer block, the attention sublayer concatenates its $H$ per-head outputs (each in $\mathbb{R}^{d_h}$, with $d_h = d_{\rm model} / H$ and $d_{\rm model}$ the residual-stream dimension) and applies an output projection $\bW_O \in \mathbb{R}^{d_{\rm model} \times d_{\rm model}}$ to produce the post-projection attention output of the block~\citep{vaswani2017attention}.
For each dataset $D$ and each loss $\set{L} \in \{ \set{L}_{\rm cls}, \set{L}_{\rm exp} \}$, we attach a forward hook to the post-projection attention output $\boldsymbol{y}$ of every block and back-propagate $\set{L}$ over a fixed batch of $256$ samples.
With the row-vector convention $\boldsymbol{y} = \boldsymbol{z} \bW_O^{\top} + \boldsymbol{b}_O$, we recover the pre-projection concatenated-head activation $\boldsymbol{z} = (\boldsymbol{y} - \boldsymbol{b}_O)(\bW_O^{\top})^{+}$ with the pseudoinverse and its gradient $\nabla_{\boldsymbol{z}} \set{L} = (\nabla_{\boldsymbol{y}} \set{L}) \bW_O$.
Reshaping both to $\mathbb{R}^{H \times d_h}$ gives the activation $\boldsymbol{z}_{b,s,h}$ of head $h$ at sample $b$ and token $s$.
We average the absolute per-token attribution score of \citet{michel2019sixteen} over the $B$ samples and $S$ tokens:
\begin{equation}
e_D[\ell,h] = \frac{1}{BS} \sum_{b,s} \bigl| \boldsymbol{z}_{b,s,h} \cdot \nabla_{\boldsymbol{z}_{b,s,h}} \set{L} \bigr|,
\label{eq:appendix:circuit_overlap:head_effect}
\end{equation}
and write $e^{\rm cls}_D$ and $e^{\rm exp}_D$ for $\set{L}_{\rm cls}$ and $\set{L}_{\rm exp}$.
The absolute inner product is a first-order estimate of the magnitude of the change in $\set{L}$ when that head's output at that token is zeroed.
Taking the absolute value per token before averaging differs from the head-level gate of \citet{michel2019sixteen}, whose gradient sums the token contributions before the absolute value.
Rescaling each gradient by the corresponding activation suppresses the early-layer mass that raw gradient norms accumulate via back-propagation chaining and is a common choice for per-head importance estimation in the mechanistic interpretability literature.

\paragraph{Coupling metric}
For each dataset $D$, let $\boldsymbol{e}^{\rm cls}_D, \boldsymbol{e}^{\rm exp}_D \in \mathbb{R}^{LH}$ denote the flattened per-head effects for $\set{L}_{\rm cls}$ and $\set{L}_{\rm exp}$.
We define the {\it circuit-overlap coefficient} as the Pearson correlation
\begin{equation}
\rho_{\rm circ}[D] = \mathrm{Pearson}(\boldsymbol{e}^{\rm cls}_D, \boldsymbol{e}^{\rm exp}_D).
\label{eq:appendix:circuit_overlap:rho}
\end{equation}
A value close to $1$ indicates that the same heads carry both classification and explanation signal;
a value close to $0$ indicates head-level specialization between the two objectives.
We further define a {\it per-layer} version
\begin{equation}
\rho_{\rm circ}^{(\ell)}[D] = \mathrm{Pearson}\bigl(\boldsymbol{e}^{\rm cls}_D[\ell, :],\; \boldsymbol{e}^{\rm exp}_D[\ell, :]\bigr), \qquad \ell \in \{1, \ldots, L\},
\label{eq:appendix:circuit_overlap:rho_layer}
\end{equation}
which restricts the correlation to the $H$ heads within a single transformer layer $\ell$ and is computed from the same per-head attribution effects as $\rho_{\rm circ}[D]$.
\space%
\subsection{Classification and Explanation Heads Largely Overlap}\label{sec:appendix:circuit_overlap:rq1}

Computing $\rho_{\rm circ}[D]$ on the self-explaining checkpoint $\btheta^D_{\rm ft\star}$ across the 10 target datasets yields a positive coupling on every dataset, with mean $\bar{\rho}_{\rm circ} = 0.49$ on ViT-B/32 (range $[0.22, 0.75]$) and $\bar{\rho}_{\rm circ} = 0.90$ on ViT-L/14 (range $[0.83, 0.96]$).
The only marginally low value is GTSRB on ViT-B/32 ($\rho_{\rm circ} = 0.22$).
It is fully resolved on ViT-L/14 ($\rho_{\rm circ} = 0.96$), consistent with capacity scaling subsuming the apparent specialization.
This comparison also differs in explanation supervision (Kernel SHAP for ViT-B/32 vs.\ Grad-CAM for ViT-L/14, see Appendix~\ref{sec:appendix:vitl}).
Grad-CAM by construction couples the explanation circuit to the classification gradients and may therefore contribute to the higher overlap on ViT-L/14.

\begin{figure}[t]
\centering
\includegraphics[width=0.7\textwidth]{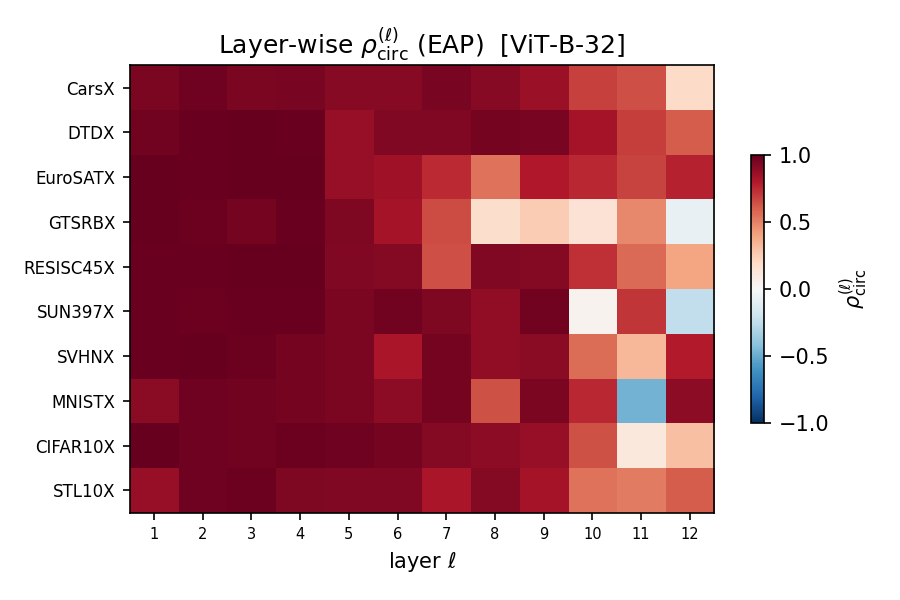}
\caption{Layer-wise circuit-overlap profile $\rho_{\rm circ}^{(\ell)}$ (per-head attribution) on the ViT-B/32 self-explaining checkpoint. Each row corresponds to one of the 10 target datasets and each column to one of the 12 transformer layers; cell color encodes $\rho_{\rm circ}^{(\ell)}$ for that (dataset, layer) pair. The heatmap is uniformly dark in the early-to-mid layers and lightens toward the final layers ($\ell \in \{10, 11, 12\}$). This is the early-shared / late-diverged structure described in the text.}
\label{fig:circuit_overlap:layerwise}
\end{figure}

\paragraph{Layer-wise profile}
Computing $\rho_{\rm circ}^{(\ell)}$ within each transformer layer reveals a consistent pattern across both backbones: early-to-mid layers exhibit near-perfect coupling, while late layers progressively decouple.
On ViT-B/32, layers $\ell \in \{1, \ldots, 6\}$ attain $\rho_{\rm circ}^{(\ell)} \in [0.90, 0.98]$, after which coupling drops to $\rho_{\rm circ}^{(12)} = 0.43$ (Figure~\ref{fig:circuit_overlap:layerwise}).
The same pattern holds on ViT-L/14.
Layers $\ell \in \{1, \ldots, 9\}$ stay above $0.97$, coupling then declines gradually, and the last two blocks reach $\rho_{\rm circ}^{(23)} = 0.43$ and $\rho_{\rm circ}^{(24)} = 0.49$ (figure omitted for space).
This early-shared / late-diverged structure is consistent with the finding of \citet{gandelsman2024interpreting} that CLS-token direct effects in CLIP-trained ViTs concentrate in the last attention layers.
This suggests that the network reuses a shared backbone for both objectives until late blocks, where head-level specialization between the classification and explanation circuits emerges.

\begin{figure}[t]
\centering
\begin{minipage}{\textwidth}
  \centering
  \includegraphics[width=0.95\textwidth]{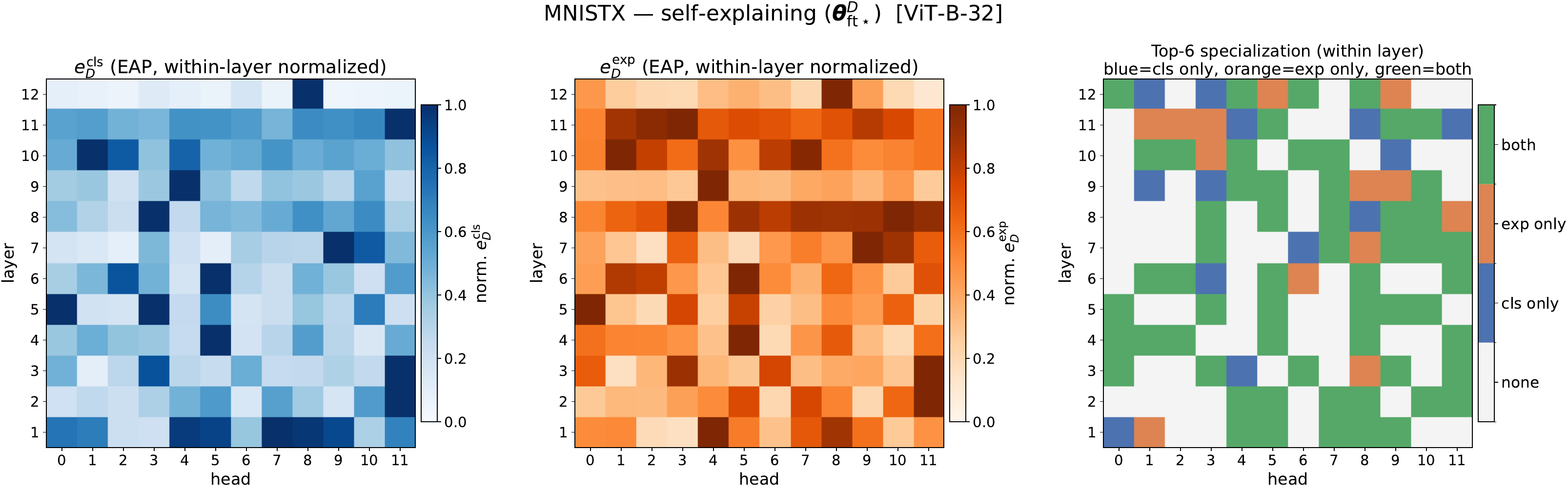}\\
  (a) MNIST, the highest-coupling target on ViT-B/32 ($\rho_{\rm circ} = 0.75$).
\end{minipage}
\par\vspace{1mm}
\begin{minipage}{\textwidth}
  \centering
  \includegraphics[width=0.95\textwidth]{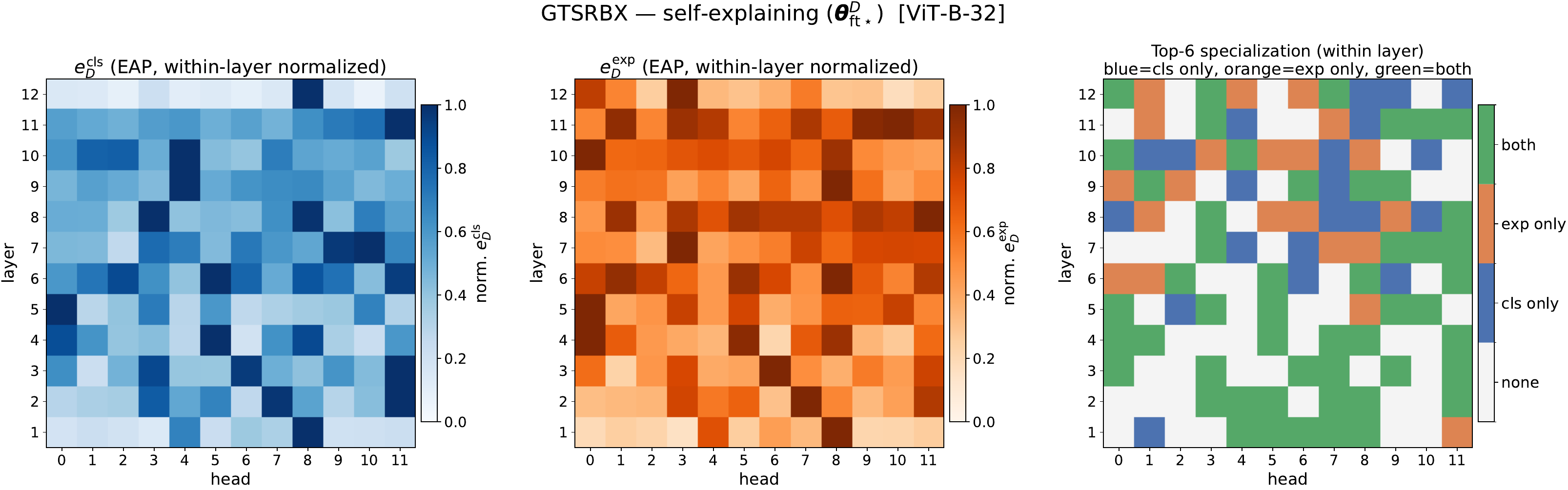}\\
  (b) GTSRB, the lowest-coupling target on ViT-B/32 ($\rho_{\rm circ} = 0.22$).
\end{minipage}
\caption{Per-head attribution effects on the ViT-B/32 self-explaining checkpoint for the two extreme-coupling target datasets. Left and middle panels: within-layer normalized $e^{\rm cls}_D[\ell, h]$ and $e^{\rm exp}_D[\ell, h]$ over the $L = 12$ layers and $H = 12$ heads. Right panel: within-layer top-$6$ specialization map (blue: cls-only, orange: exp-only, green: both, white: neither).}
\label{fig:circuit_overlap:headlevel:vitb}
\end{figure}

\paragraph{Per-dataset head-level visualizations}
Figure~\ref{fig:circuit_overlap:headlevel:vitb} shows the per-head attribution effects $e^{\rm cls}_D[\ell, h]$ and $e^{\rm exp}_D[\ell, h]$ for the highest- and lowest-coupling target datasets on ViT-B/32 (MNIST and GTSRB, respectively).
The third panel of each row marks, for each transformer layer, the heads that fall into the within-layer top-$k$ for $e^{\rm cls}_D$ only, $e^{\rm exp}_D$ only, or both, with $k = 6$ chosen for visual clarity.
These maps show where the shared (green) heads lie.
They are not by themselves evidence of overlap, because two random top-$6$ sets out of $12$ heads share at least one head with probability above $0.99$, and the quantitative evidence is $\rho_{\rm circ}$.
The analogous maps for ViT-L/14 are omitted for space, and we compare the backbones with the layer-wise $\rho_{\rm circ}^{(\ell)}$ instead.
GTSRB inverts its rank between backbones.
It is the lowest-coupling dataset on ViT-B/32 ($\rho_{\rm circ} = 0.22$) but the highest-coupling one on ViT-L/14 ($\rho_{\rm circ} = 0.96$).
On ViT-B/32, its layer-wise coupling falls below $0.5$ from layer $8$ onward (Figure~\ref{fig:circuit_overlap:headlevel:vitb}(b)).
On ViT-L/14, it stays at or above $0.74$ up to layer $22$ and drops only in the last two blocks ($-0.10$ and $0.28$).

\subsection{Circuit Overlap Is Present before Explanation Training}\label{sec:appendix:circuit_overlap:rq2}

\begin{figure}[t]
\centering
\includegraphics[width=0.85\textwidth]{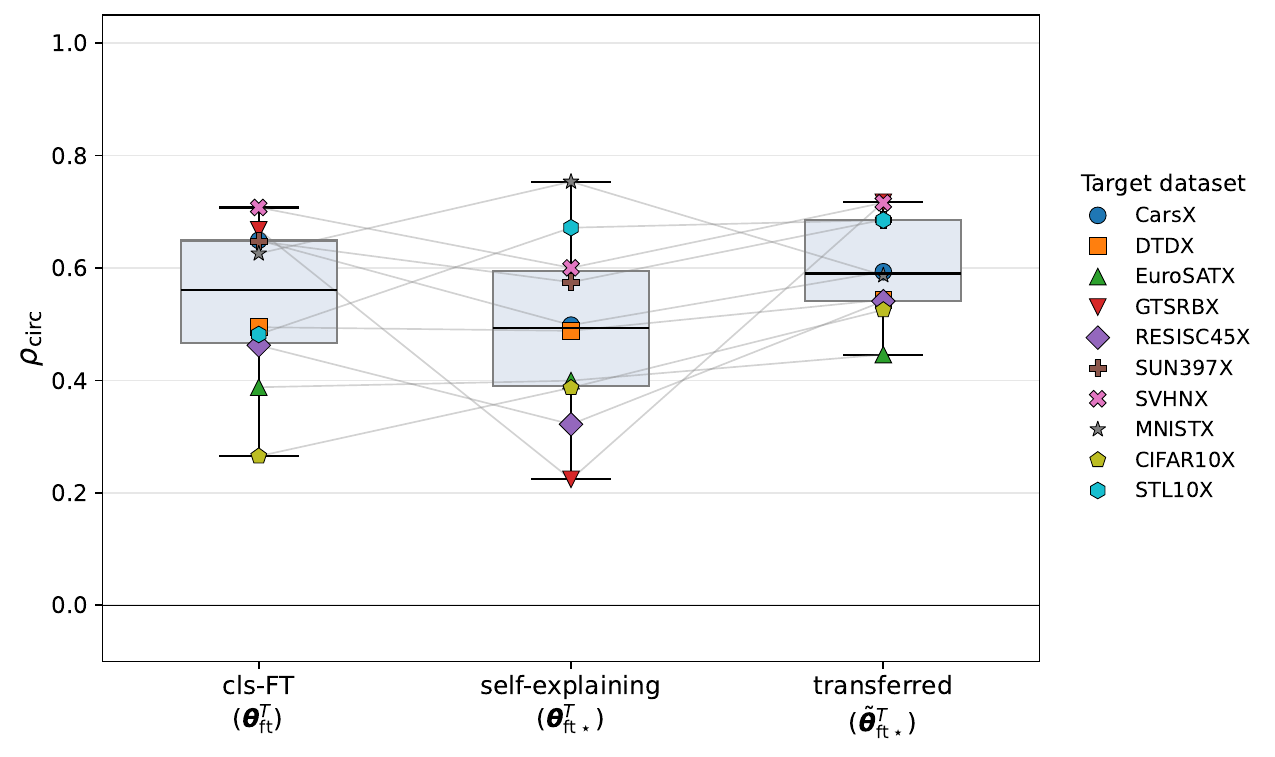}
\caption{Circuit-overlap coefficient $\rho_{\rm circ}$ on ViT-B/32 measured under per-head attribution for three training conditions: cls-only finetuned $\btheta^T_{\rm ft}$, self-explaining $\btheta^T_{\rm ft\star}$, and transferred $\tilde{\btheta}^T_{\rm ft\star}$ (with source fixed to $S = \text{ImageNet+X}$). Each marker color/shape encodes one of the 10 target datasets (legend) and gray lines connect the same target across conditions. Conditions (i) and (ii) show no detectable difference, while the transferred condition has a slightly higher mean ($0.60$) than both and than the ImageNet-source coupling level ($\bar\rho_{\rm circ}^{S = \text{ImageNet+X}} = 0.56$). The 10 targets of this analysis include CIFAR10 in place of CIFAR100 of Table~\ref{tab:experiment:imagenet}.}
\label{fig:circuit_overlap:conditions}
\end{figure}

A natural hypothesis is that the strong cls--exp coupling reported above is itself induced by jointly optimizing $\set{L}_{\rm cls}$ and $\set{L}_{\rm exp}$ during finetuning.
We test this hypothesis by repeating the measurement on three checkpoints per dataset:
(i) the classification-only finetuned model $\btheta^D_{\rm ft}$, which has never seen explanation supervision;
(ii) the self-explaining model $\btheta^D_{\rm ft\star}$, which has been jointly optimized for classification and explanation; 
and (iii) the transferred model $\tilde{\btheta}^{T}_{\rm ft\star}$ obtained by task arithmetic~\bref{eq:proposed:theta_init}, with the source fixed to ImageNet+X for every target $T$.
Figure~\ref{fig:circuit_overlap:conditions} reports $\rho_{\rm circ}$ on ViT-B/32; the corresponding ViT-L/14 figure is omitted for space and the L/14 statistics are summarized in text below.
On both backbones, conditions (i) and (ii) show no detectable difference (B-32: $\bar\rho_{\rm circ} = 0.54$ vs.\ $0.49$, paired Wilcoxon $p = 0.63$; L-14: $0.90$ vs.\ $0.90$, $p = 0.70$).
The cls--exp coupling thus already exists in the classification-only checkpoint, and this evaluation detects no additional coupling from joint optimization.
The transferred condition (iii), in contrast, changes the coupling.
On ViT-L/14, the target checkpoints attain $\bar\rho_{\rm circ} = 0.90$ in both (i) and (ii), but the ImageNet-source self-explaining checkpoint attains only $0.58$.
The mean of the transferred condition drops to $0.77$, toward the source ($p = 0.006$ vs.\ (i) and $p = 0.002$ vs.\ (ii)).
On ViT-B/32, the source coupling ($0.56$) is close to the target means, and the transferred mean rises to $0.60$ ($p = 0.06$ vs.\ (i) and $p = 0.05$ vs.\ (ii)).
This rise passes the source value, so a pull toward the source is clear only on ViT-L/14, where the source and target coupling levels differ.
This source-pull effect on ViT-L/14 is consistent with the parameter-arithmetic interpretation of task arithmetic, in which $\tilde{\btheta}^{T}_{\rm ft\star}$ inherits part of the head-level structure of the source.

Since the coupling is present before any explanation supervision, it is not created by training the explanation output, which is consistent with both tokens flowing through a shared attention backbone.
This answers the concern raised at the start of this appendix at a descriptive level, since the two parallel outputs share much of their head-level computation inside the backbone.
Causal evidence on where the transferred effect is mediated comes from the parameter interventions of Appendices~\ref{sec:appendix:circuit_overlap:mechanism} and~\ref{sec:appendix:controls}.
\space%
\subsection{Where the Transferred Effect Is Mediated}\label{sec:appendix:circuit_overlap:mechanism}

The analyses above describe the self-explaining checkpoints.
This subsection asks where the explainability vector $\btau^{\rm S}_{\star}$ itself acts when it is added to a target.
We use three analyses.
The first measures how the norm of $\btau^{\rm S}_{\star}$ is distributed over the blocks.
The second adds only part of $\btau^{\rm S}_{\star}$ to the target and checks how much of the improvement remains, using the layer-restricted controls of Appendix~\ref{sec:appendix:controls}.
The third measures how the attributions of individual heads change when $\btau^{\rm S}_{\star}$ is added.
The first and third show where the change is large, but not whether that change matters for the explanations.
Only the second tests this directly.

\paragraph{Distribution of the vector norm}
Decomposing the squared norm of $\btau^{\rm S}_{\star}$ ($\alpha = 0.8$, source ImageNet+X) exactly over transformer blocks and module types, the last six blocks carry $59.2\%$ of $\|\btau^{\rm S}_{\star}\|^2$ (first six blocks $39.4\%$, non-block parameters $1.4\%$).
Table~\ref{tab:appendix:mechanism:modules} shows that this back-loading comes mainly from the pre-MLP LayerNorm.
$89.1\%$ of the squared norm of its change sits in the last six blocks, concentrated in blocks $10$--$11$, and it accounts for $81.6\%$ of the difference in squared norm between the two halves.
Without it, the last six blocks hold $52.3\%$ of the squared norm of the remaining block parameters.
The attention projections are close to an even split.

\begin{table}[t]
\centering
\caption{Share of each module type's squared $\ell_2$ norm of $\btau^{\rm S}_{\star}$ that lies in the last six transformer blocks of ViT-B/32 (an even split would give $50\%$). The two norm columns are $\ell_2$ norms, and the share is computed from their squares. Modules are the pre-MLP and pre-attention LayerNorms, the MLP down- and up-projections, and the attention input (QKV) and output projections.}
\label{tab:appendix:mechanism:modules}
\small
\begin{tabular}{@{}lccc@{}}
\toprule
Module & Norm, blocks 1--6 & Norm, blocks 7--12 & Share in blocks 7--12 \\
\midrule
LayerNorm before MLP & $1.572$ & $4.500$ & $89.1\%$ \\
LayerNorm before attention & $1.319$ & $1.959$ & $68.8\%$ \\
MLP down-projection & $2.917$ & $3.392$ & $57.5\%$ \\
MLP up-projection & $3.854$ & $4.033$ & $52.3\%$ \\
Attention QKV projection & $3.594$ & $3.300$ & $45.7\%$ \\
Attention output projection & $1.669$ & $1.520$ & $45.3\%$ \\
\bottomrule
\end{tabular}
\end{table}

\paragraph{Layer-restricted intervention}
Adding only the part of $\btau^{\rm S}_{\star}$ that lies in the last six blocks recovers $84.0\%$ of the Rank corr.\ improvement and $81.9\%$ of the IoU@10 improvement of the full vector over the no-vector model on 10-target means.
These means are over the CIFAR10 set of Appendix~\ref{sec:appendix:controls}, in which CIFAR10 replaces SUN397.
MNIST is the outlier at $58\%$ and $51\%$.
The part in the first six blocks, in contrast, recovers essentially nothing and is indistinguishable from adding no vector (Table~\ref{tab:appendix:controls:means}).
On 10-target means, neither restricted vector changes classification accuracy by more than $0.3$ points.
Most of the transferred effect is therefore mediated by parameter changes in the later blocks.
The two restricted vectors do not add up to the full vector, however.
The first six blocks recover nothing alone, yet the full vector exceeds the last six blocks by $16.0\%$ of the Rank corr.\ improvement and $18.1\%$ of the IoU@10 improvement.
We do not separate this remainder into the contributions of the first six blocks, the non-block parameters, and their interactions.

\paragraph{Per-head attribution changes}
Using the per-head attribution effect $e_D[\ell, h]$ in~\bref{eq:appendix:circuit_overlap:head_effect}, we measured how the effects change when $0.7\,\btau^{\rm S}_{\star}$ is added to $\btheta^{\rm T}_{\rm ft}$ on three targets (Cars, GTSRB, and RESISC45), using $256$ samples each.
Table~\ref{tab:appendix:mechanism:heads} reports, for each loss, the share of the summed absolute change that falls in the last six blocks, in the last two blocks, and in the five most-changed of the $144$ heads.
The change in the explanation-loss effects leans mildly toward the later blocks.
The last six blocks hold $53.5\%$ of it on average, against $50\%$ for a uniform spread, and the last two blocks hold $19.7\%$, against $16.7\%$.
The five most-changed heads account for $10.8\%$ of the change, against $3.5\%$ for a uniform spread.
For GTSRB and RESISC45, four of these five heads lie in the last block, and for Cars two do.
Most of the change is therefore spread over many heads, although the largest individual changes sit in the last block.
The classification-loss effects shift mildly toward the earlier blocks ($43.6\%$ in the last six).
On GTSRB and RESISC45, their summed absolute change is three to four orders of magnitude smaller than that of the explanation-loss effects, so their shares on these two targets should not be over-read.
This per-head picture is weaker and less clean than the norm and layer-restriction results above, and we read it as partial corroboration rather than as an independent localization.

\begin{table}[t]
\centering
\caption{Where the per-head attribution effects $e_D[\ell, h]$ change when $0.7\,\btau^{\rm S}_{\star}$ is added to $\btheta^{\rm T}_{\rm ft}$ ($256$ samples per target). Each entry is the share of the summed absolute change over the $144$ heads that falls in the last six blocks (blocks 7--12), in the last two blocks (blocks 11--12), or in the five heads with the largest absolute change. The last row gives the shares for a uniform spread.}
\label{tab:appendix:mechanism:heads}
\small
\begin{tabular}{@{}lcccccc@{}}
\toprule
 & \multicolumn{3}{c}{Explanation loss} & \multicolumn{3}{c}{Classification loss} \\
\cmidrule(lr){2-4}\cmidrule(l){5-7}
Target & Blocks 7--12 & Blocks 11--12 & Top-5 heads & Blocks 7--12 & Blocks 11--12 & Top-5 heads \\
\midrule
Cars & $47.7\%$ & $15.1\%$ & $8.7\%$ & $43.4\%$ & $8.7\%$ & $6.8\%$ \\
GTSRB & $54.0\%$ & $20.9\%$ & $11.1\%$ & $45.2\%$ & $13.8\%$ & $10.1\%$ \\
RESISC45 & $58.8\%$ & $23.1\%$ & $12.5\%$ & $42.3\%$ & $8.7\%$ & $8.0\%$ \\
\midrule
Mean & $53.5\%$ & $19.7\%$ & $10.8\%$ & $43.6\%$ & $10.4\%$ & $8.3\%$ \\
Uniform & $50.0\%$ & $16.7\%$ & $3.5\%$ & $50.0\%$ & $16.7\%$ & $3.5\%$ \\
\bottomrule
\end{tabular}
\end{table}

\paragraph{Interpretation}
The part of $\btau^{\rm S}_{\star}$ in the early blocks has no measurable effect on its own, and these blocks are shared between classification and explanation (Appendix~\ref{sec:appendix:circuit_overlap:rq1}).
The explanation mechanism is concentrated in the later blocks, where most of the vector's effect arises, on top of the target's own representations.
This is consistent with the source-pull effect of Appendix~\ref{sec:appendix:circuit_overlap:rq2}, where, on ViT-L/14, adding the vector shifts the transferred model's head-level coupling toward that of the source, and with the two-task experiment of Appendix~\ref{sec:appendix:stress}, where the same vector yields task-appropriate attributions on two tasks.
These analyses establish where the transferred effect is mediated, not a complete causal account of how the attributions are computed.
\space%

\section{Feature Pool and ViT-L/14 Predictability Analysis}\label{sec:appendix:predict}

This appendix characterizes when and how the predictability examined in Section~\ref{sec:experiment:predict} holds, with (i) formal definitions for the six features, (ii) the hypothesis-level rationale for the six features and the collinearity check that consolidates redundant $\ell_2$/cosine variants, and (iii) results on ViT-L/14 with an oracle-normalized target $R_q$ that complements the raw $\Delta_q$ target.

\paragraph{Formal definition of the predictor features}
The classification-circuit features (rows 1--3 of Table~\ref{tab:appendix:predict:tau_def}) build on the head-level attribution effect vectors $\boldsymbol{e}^{\rm cls}(\btheta;\set{D}), \boldsymbol{e}^{\rm exp}(\btheta;\set{D}) \in \mathbb{R}^{LH}$ defined in Appendix~\ref{sec:appendix:circuit_overlap} (same forward-hook setup, $256$ samples, per-head attribution of \citet{michel2019sixteen}, flattened in row-major order), made explicit in the model parameter $\btheta$ and the dataset $\set{D}$ used to compute the loss.
We use $\boldsymbol{a} = \boldsymbol{e}^{\rm cls}(\tilde{\btheta}^{\rm T}_{\rm ft\star};\set{D}^{\rm T})$, $\boldsymbol{b} = \boldsymbol{e}^{\rm cls}(\btheta^{\rm T}_{\rm ft};\set{D}^{\rm T})$, and $\boldsymbol{c} = \boldsymbol{e}^{\rm exp}(\btheta^{\rm S}_{\rm ft\star};\set{D}^{\rm S})$ as the transferred-cls, target-cls, and source-exp effect vectors respectively, where $\tilde{\btheta}^{\rm T}_{\rm ft\star}$ is the transferred model of \bref{eq:proposed:theta_init}, write $\mathrm{Top}_k(\cdot)$ for the index set of the $k$ largest entries with $k = 16$ for ViT-B/32 and $k = 32$ for ViT-L/14, and use the standard Jaccard index $J(A, B) = |A \cap B|/|A \cup B|$.
The $\btau$-output-shift features (rows 4--6) build on $\mathtt{[CLS]}$ feature shifts on unlabeled target images.
Let $f^{\mathtt{[CLS]}}_{\btheta}(\bx) \in \mathbb{R}^{K}$ denote the $\mathtt{[CLS]}$ output of the feature extractor with parameters $\btheta$ on image $\bx$, and let $\set{X}^{\rm T}_{\rm unlab} = \{\bx_i\}_{i=1}^{N_T}$ be unlabeled target images.
Writing $\bh^{\rm base}_i = f^{\mathtt{[CLS]}}_{\btheta_{\rm base}}(\bx_i)$ for the zero-shot backbone and $\bh^{\rm taued}_i = f^{\mathtt{[CLS]}}_{\btheta_{\rm base} + \btau^{\rm S}_{\star}}(\bx_i)$ for the $\btau$-perturbed backbone, define the per-image shift $\bdelta_i = \bh^{\rm taued}_i - \bh^{\rm base}_i$ and the image-mean centroid $\bar{\bh}^{\bullet} = \frac{1}{N_T}\sum_{i=1}^{N_T} \bh^{\bullet}_i$ for $\bullet \in \{\rm base, taued\}$.
Row 6 averages features across images first and then takes the cosine, factoring out per-image directional noise and isolating the systematic $\btau$-induced bias.

\begin{table}[t]
\centering
\caption{Formal definition of the six predictor features used in Section~\ref{sec:proposed:predict}. Classification-circuit features compare flattened per-head attribution effect vectors. $\btau$-output-shift features compare $\mathtt{[CLS]}$ representations of unlabeled target images under a frozen zero-shot backbone with and without adding $\btau^{\rm S}_{\star}$. The index $i \in \{1, \ldots, N_T\}$ runs over the unlabeled target images $\bx_i \in \set{X}^{\rm T}_{\rm unlab}$. We use $k = 16$ (ViT-B/32) and $k = 32$ (ViT-L/14) for the top-$k$ Jaccard.}
\label{tab:appendix:predict:tau_def}
\small
\begin{tabular}{@{}cllc@{}}
\toprule
\# & Class & Identifier & Definition \\
\midrule
1 & cls-circuit  & \texttt{cos\_tr\_vs\_T\_ft\_cls} & $\cos(\boldsymbol{a}, \boldsymbol{b})$ \\
2 & cls-circuit  & \texttt{jacc\_tr\_vs\_T\_ft\_cls} & $J\bigl(\mathrm{Top}_k(\boldsymbol{a}),\, \mathrm{Top}_k(\boldsymbol{b})\bigr)$ \\
3 & cls-circuit  & \texttt{cos\_S\_ftstar\_exp\_vs\_T\_ft\_cls} & $\cos(\boldsymbol{c}, \boldsymbol{b})$ \\
\midrule
4 & $\tau$-shift & \texttt{tau\_feat\_shift\_l2\_mean}         & $\frac{1}{N_T}\sum_{i=1}^{N_T} \|\bdelta_i\|_2$ \\
5 & $\tau$-shift & \texttt{tau\_feat\_shift\_l2\_std}          & $\mathrm{std}_i\bigl(\|\bdelta_i\|_2\bigr)$ \\
6 & $\tau$-shift & \texttt{tau\_centroid\_cos}                 & $\cos(\bar{\bh}^{\rm base}, \bar{\bh}^{\rm taued})$ \\
\bottomrule
\end{tabular}
\end{table}

\paragraph{Why these six features}
The classification-circuit features (rows 1--3) test how the head-level classification structure of the target relates to the transferred model and to the source explanation circuit.
They are the transferred--target cls-effect cosine, the top-$k$ Jaccard overlap, and the source-exp vs.\ target-cls cross-cosine, all motivated by the head-level overlap of Appendix~\ref{sec:appendix:circuit_overlap}.
The $\btau$-output-shift features (rows 4--6) test whether $\btau^{\rm S}_{\star}$ non-trivially perturbs target image content.
They are the per-image $\ell_2$ shift mean and std on the zero-shot backbone, plus the centroid-cosine that filters per-image directional noise.
Within each family we considered $\ell_2$-magnitude and cosine variants in parallel.
We collapsed pairs with $|r| \gtrsim 0.8$, where $r$ is the Pearson correlation between two candidate features computed once on the 10 pairs of each backbone (no target variable is used).
For example, $\cos$ tracks $-\ell_2$ at $|r| \approx 0.80$--$0.96$ across backbones, and the complement-of-top-$k$ tracks the Jaccard at $r = -1.00$ by construction.
Keeping one representative per cluster yields the 3+3 features of Table~\ref{tab:appendix:predict:tau_def}.
The pruning uses feature--feature correlations only and is therefore target-independent.
The target $\Delta_q$ is used only for regression fitting and regularization-strength selection, both of which use only the training targets of each leave-one-target-out fold.
As a post-hoc observation, feeding all variants to the regularizer instead of the pruned pool gives $r = +0.80$ rather than $+0.92$ for ViT-B/32 Lasso on E-MSE and no non-trivial ViT-L/14 result under $R_q$, which we attribute to near-duplicate features splitting the regularization budget.
This observation was not used to choose the pool.

\paragraph{Full ViT-B/32 results}
For completeness, Table~\ref{tab:appendix:predict:vitb_full} reports held-out Pearson $r$ for ViT-B/32 across all five explanation-quality metrics.
The main paper (Table~\ref{tab:experiment:predict}) shows three representative metrics under the page budget.

\begin{table}[t]
\centering
\caption{Held-out Pearson correlation (LOTO, $N = 10$) on ViT-B/32, full grid corresponding to Table~\ref{tab:experiment:predict}.}
\label{tab:appendix:predict:vitb_full}
\small
\begin{tabular}{@{}lccccc@{}}
\toprule
Method & E-MSE & IoU@1 & IoU@5 & IoU@10 & Rank corr. \\
\midrule
Lasso & $\mathbf{+0.92}$ & $+0.19$ & $\mathbf{+0.70}$ & $+0.81$ & $+0.77$ \\
Ridge & $+0.91$ & $\mathbf{+0.33}$ & $+0.63$ & $\mathbf{+0.85}$ & $\mathbf{+0.82}$ \\
\bottomrule
\end{tabular}
\end{table}

\paragraph{ViT-L/14 results with an oracle-normalized target $R_q$}
On ViT-L/14, the raw $\Delta_q$ target is dominated by per-target variance, with GTSRB as an outlier whose large $\Delta_q$ disproportionately drives the regression.
We therefore evaluate ViT-L/14 with an oracle-normalized target that absorbs per-target scale.
For each $(S, T)$ pair, in addition to $b_q = q(\btheta^{\rm T}_{\rm ft})$ and $a_q = q(\tilde{\btheta}^{\rm T}_{\rm ft\star})$ from Section~\ref{sec:experiment:predict}, let $o_q = q(\btheta^{\rm T}_{\rm ft\star})$ denote the target oracle, which is the explanation-quality score of the target self-explaining model trained with explanation supervision (the ``target+X'' model of Section~\ref{sec:experiment:imagenet}).
We define
\begin{equation}
R_q(S, T) = \frac{b_q - a_q}{b_q - o_q},
\label{eq:appendix:predict:R}
\end{equation}
the fraction of the explanation-quality gap from the target classification-only baseline $\btheta^{\rm T}_{\rm ft}$ to the target oracle $\btheta^{\rm T}_{\rm ft\star}$ that is closed by the transferred model $\tilde{\btheta}^{\rm T}_{\rm ft\star}$.
When the oracle improves on the baseline, which holds for every target and metric here, $R_q = 1$ matches the oracle, $R_q = 0$ indicates no improvement, $R_q < 0$ indicates degradation, and $R_q > 1$ exceeds the oracle.
The oracle score $o_q$ enters only when the predictor is fitted and evaluated.
Predicting the outcome on a new target uses the six model-internal features alone and requires no target-side explanation supervision.

Table~\ref{tab:appendix:predict:vitl} reports held-out Pearson $r$ between predicted and actual $R_q$ on ViT-L/14 with the same 6-feature pool and the same nested-LOTO protocol as Section~\ref{sec:experiment:predict}.
Three of the five explanation-quality metrics are statistically significant under $R_q$ at uncorrected one-sided permutation $p < 0.05$, computed as in Section~\ref{sec:experiment:predict}: IoU@1 (Lasso $r = +0.82$, $p = 0.016$), E-MSE (Ridge $r = +0.81$, $p = 0.034$), and Rank corr.\ (Lasso $r = +0.70$, $p = 0.046$).
The remaining two metrics, IoU@5 and IoU@10, are not predictable under $R_q$ on this backbone.
This residual backbone--metric specificity is discussed in Appendix~\ref{sec:limitations}.

\begin{table}[t]
\centering
\caption{Held-out Pearson correlation (LOTO, $N = 10$) on ViT-L/14, using the oracle-normalized target $R_q$ defined in Eq.~(\ref{eq:appendix:predict:R}). Best per metric is in \textbf{bold}.}
\label{tab:appendix:predict:vitl}
\small
\begin{tabular}{@{}lccccc@{}}
\toprule
Method & E-MSE & IoU@1 & IoU@5 & IoU@10 & Rank corr. \\
\midrule
Lasso & $+0.81$ & $\mathbf{+0.82}$ & $+0.21$ & $-0.06$ & $\mathbf{+0.70}$ \\
Ridge & $\mathbf{+0.81}$ & $+0.81$ & $\mathbf{+0.23}$ & $\mathbf{-0.03}$ & $+0.49$ \\
\bottomrule
\end{tabular}
\end{table}

\section{Computing Environment for Experiments}\label{sec:appendix:environment}
The experiments were conducted in parallel on Rocky Linux machines with two Intel Xeon Platinum 8558 2.1GHz (48 cores) CPUs, 2,048 GB of RAM, and eight NVIDIA H200 SXM GPUs with 141GB of memory.
The code was implemented in Python 3.9.21 using PyTorch 2.6.
The additional experiments in Appendices~\ref{sec:appendix:circuit_overlap:mechanism} and~\ref{sec:appendix:faithfulness}--\ref{sec:appendix:stress} and the leave-one-source-out evaluation in Appendix~\ref{sec:appendix:pair:predictor} were run on NVIDIA RTX 6000 Ada GPUs with 48GB of memory, using the same code base.
\space%

\section{Experiments with ViT-L/14}\label{sec:appendix:vitl}
To validate the effectiveness of the proposed method with a larger vision transformer, we conduct experiments using ViT-L/14 as the backbone of the self-explaining image classifier.
\space%
\subsection{Setup}\label{sec:appendix:vitl:setup}
The model input is a 224$\times$224 image, which is split into 16$\times$16 patches of size 14$\times$14.
Therefore, the number of patches $M$ is 256.

The GT patch-level attributions for ImageNet and the 10 target datasets, Cars~\cite{krause20133d}, DTD~\cite{cimpoi2014describing}, EuroSAT~\cite{helber2019eurosat}, GTSRB~\cite{stallkamp2011german}, RESISC45~\cite{cheng2017remote}, SUN397~\cite{xiao2016sun}, SVHN~\cite{netzer2011reading}, MNIST~\cite{lecun1998mnist}, CIFAR100~\cite{Krizhevsky2009LearningML}, and STL10~\cite{coates2011analysis}, are obtained by applying Grad-CAM~\cite{Selvaraju2020-fx}, implemented by \texttt{pytorch-grad-cam}~\cite{jacobgilpytorchcam}, to the finetuned ViT-L/14-based image classifiers for each dataset.
Specifically, we apply Grad-CAM to the layer normalization of the final residual block in the Vision Transformer encoder.
The obtained attributions are pixel-wise values.
We therefore aggregate them into patch-level attribution values by averaging the pixel-wise attribution values within each patch.

\subsection{Quantitative Results}\label{sec:appendix:vitl:quantitative}
Table~\ref{tab:experiment:imagenet:vitl} shows the average performances of the explainability transferred by the proposed method with $\alpha=0.8$ in \bref{eq:proposed:objective} and $\lambda_1=1.0, \lambda_2=0.7$ in \bref{eq:proposed:theta_init}, in comparison to the baseline models with the parameters $\btheta_{\rm base}$, $\btheta^{\rm S}_{\rm ft}$, $\btheta^{\rm S}_{\rm ft\star}$, and $\btheta^{\rm T}_{\rm ft}$.
For reference, we also show the performances of the model finetuned on the target dataset with explanation supervision, i.e., the oracle $\btheta^{\rm T}_{\rm ft\star}$.
The table shows that compared to the baselines, the proposed method achieves the second-best accuracy while achieving the best or second-best explainability performances.
The accuracy gap to the oracle $\btheta^{\rm T}_{\rm ft\star}$ is $0.015$ ($\approx 1.6\%$ relative), and the E-RMSE gap is $\approx 1.4\%$ relative. The proposed method thus transfers explainability acquired on ImageNet+X while largely preserving classification accuracy.
The method with the best accuracy is the model $\btheta^{\rm T}_{\rm ft}$ finetuned on the target dataset without explanation supervision.
However, this model shows the worst or second-worst explainability performances.
In contrast, the best Rank corr. is achieved by the model $\btheta^{\rm S}_{\rm ft\star}$ finetuned on ImageNet+X, but this model has the lowest accuracy.
The proposed method combines the high target-domain accuracy of $\btheta^{\rm T}_{\rm ft}$ with the explainability vector extracted from $\btheta^{\rm S}_{\rm ft\star}$, attaining classification accuracy close to that of $\btheta^{\rm T}_{\rm ft}$ ($0.928$ vs.\ $0.940$) and explanation quality close to the oracle $\btheta^{\rm T}_{\rm ft\star}$.

Overall, the classification accuracy is higher than that of the ViT-B/32-based models shown in Table~\ref{tab:experiment:imagenet}, consistent with the larger model capacity of ViT-L/14.
On the other hand, the explainability performance scores are lower than those of the ViT-B/32-based models.
The two settings differ in the backbone, the GT generator (Grad-CAM instead of Kernel SHAP), and the number of patches ($256$ instead of $49$), so the scores are not directly comparable.
Possible causes include the different GT generator and the finer patch grid, which also lowers the chance level of IoU@1 from $1/49$ to $1/256$.

\begingroup
\renewcommand{\arraystretch}{1.5} %
\begin{table}[t]
\caption{
Average performances and their standard deviations of ViT-L/14-based self-explaining models over the 10 target datasets when transferring explainability learned on ImageNet+X, compared with those of baselines.
The proposed method's results are reported at $\alpha = 0.8$ and $\lambda_2 = 0.7$ (with $\lambda_1 = 1.0$ as in the common setup).
Here, the last row shows the oracle performances achieved by finetuning on the target datasets with explanation supervision (target+X).
\textbf{\underline{Bold and underlined values}} indicate the best among the proposed method and the baselines for each metric, while \textbf{bold values} indicate that there is no statistically significant difference from the best, as determined by a paired $t$-test at a significance level of 0.05.
}
\label{tab:experiment:imagenet:vitl}
\centering
\resizebox{1.0\textwidth}{!}{%
\begin{NiceTabular}{@{}rlcccccc@{}}
\toprule
 & Param. & Accuracy ($\uparrow$) & E-RMSE ($\downarrow$) & Rank corr. ($\uparrow$) & IoU@1 ($\uparrow$) & IoU@5 ($\uparrow$) & IoU@10 ($\uparrow$) \\ \midrule
Zero-shot & $\btheta_{\rm base}$ & $0.630 \pm 0.151$ & $1.470 \pm 0.043$ & $-0.050 \pm 0.028$ & $\bf 0.003 \pm 0.002$ & $\bf 0.007 \pm 0.004$ & $0.014 \pm 0.006$ \\
Finetuned on ImageNet & $\btheta^{\rm S}_{\rm ft}$ & $0.542 \pm 0.190$ & $1.486 \pm 0.052$ & $-0.072 \pm 0.042$ & $0.002 \pm 0.001$ & $0.005 \pm 0.003$ & $0.011 \pm 0.005$ \\
Finetuned on ImageNet+X & $\btheta^{\rm S}_{\rm ft\star}$ & $0.501 \pm 0.197$ & $\bf 1.256 \pm 0.078$ & \underline{$\bf 0.121 \pm 0.059$} & $\bf 0.011 \pm 0.007$ & $\bf 0.030 \pm 0.015$ & $\bf 0.053 \pm 0.024$ \\
Finetuned on target & $\btheta^{\rm T}_{\rm ft}$ & \underline{$\bf 0.940 \pm 0.065$} & $1.493 \pm 0.058$ & $-0.069 \pm 0.038$ & $0.001 \pm 0.001$ & $0.005 \pm 0.003$ & $0.010 \pm 0.005$ \\
Proposed & $\tilde{\btheta}^{\rm T}_{\rm ft\star}$ & $0.928 \pm 0.072$ & \underline{$\bf 1.253 \pm 0.106$} & $\bf 0.117 \pm 0.072$ & \underline{$\bf 0.015 \pm 0.017$} & \underline{$\bf 0.042 \pm 0.050$} & \underline{$\bf 0.067 \pm 0.063$} \\ \dotrule[1.5pt]
Finetuned on target+X & $\btheta^{\rm T}_{\rm ft\star}$ & $0.943 \pm 0.069$ & $1.236 \pm 0.086$ & $0.123 \pm 0.049$ & $0.016 \pm 0.012$ & $0.045 \pm 0.032$ & $0.075 \pm 0.048$ \\ \bottomrule
\end{NiceTabular}
}
\end{table}
\endgroup
\space%
\subsection{Qualitative Evaluation}\label{sec:appendix:vitl:qualitative}

\begin{figure}[t]
\centering
\includegraphics[width=\textwidth]{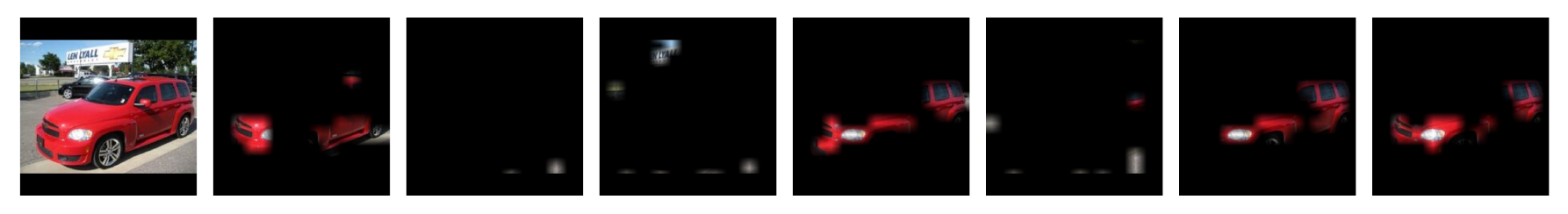}\\
\vspace{-0.2cm}
\includegraphics[width=\textwidth]{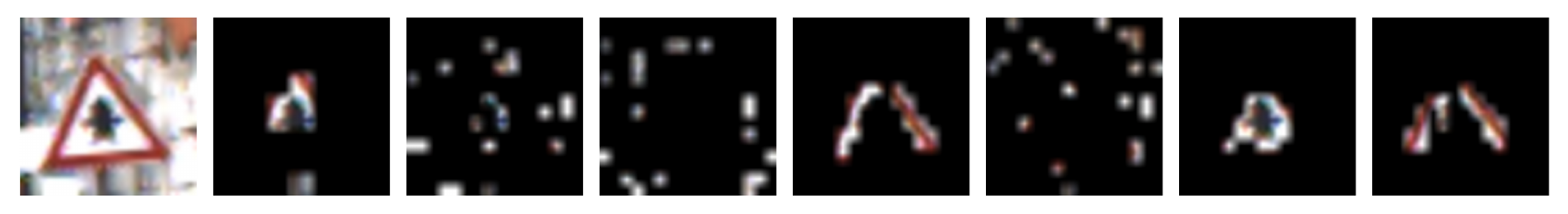}\\
\vspace{-0.2cm}
\includegraphics[width=\textwidth]{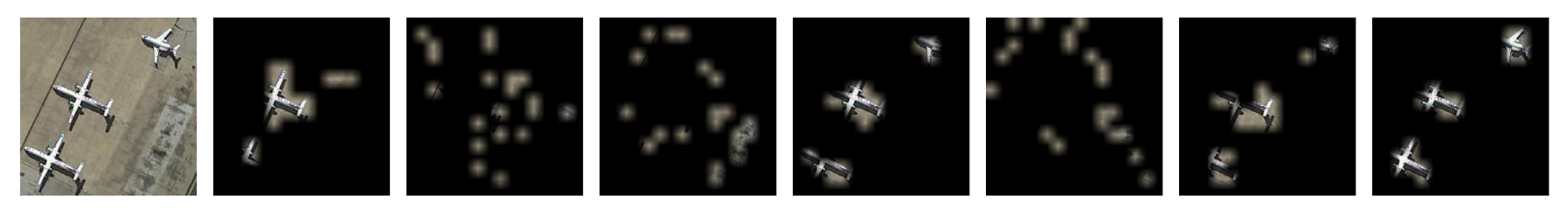}\\
\vspace{-2mm}
\noindent\scriptsize
\begin{minipage}[t]{0.125\textwidth}\centering (a) Input\end{minipage}%
\begin{minipage}[t]{0.125\textwidth}\centering (b) Ground-truth\end{minipage}%
\begin{minipage}[t]{0.125\textwidth}\centering (c) Zero-shot\end{minipage}%
\begin{minipage}[t]{0.125\textwidth}\centering (d) Finetuned on ImageNet\end{minipage}%
\begin{minipage}[t]{0.125\textwidth}\centering (e) Finetuned on ImageNet+X\end{minipage}%
\begin{minipage}[t]{0.125\textwidth}\centering (f) Finetuned on target\end{minipage}%
\begin{minipage}[t]{0.125\textwidth}\centering (g) Proposed\end{minipage}%
\begin{minipage}[t]{0.125\textwidth}\centering (h) Finetuned on target+X\end{minipage}%
\normalsize
\caption{Visualization examples highlighting the top-20 patches of each attribution produced using ViT-L/14 self-explaining image classifiers.
Rows show (from top to bottom) the results for Cars, GTSRB, and RESISC45 datasets.}
\label{fig:experiment:imagenet:example:vitl}
\end{figure}

To evaluate the predicted patch-level attributions qualitatively, we visualize the top-20 patches in the attributions produced by the proposed method and the other methods in Figure~\ref{fig:experiment:imagenet:example:vitl}.
The visualization results are similar to those of ViT-B/32 shown in Figure~\ref{fig:experiment:imagenet:example}.
In particular, the patches highlighted by the proposed method (depicted in Figure~\ref{fig:experiment:imagenet:example:vitl}(g)) are more similar to the GT ones (depicted in Figure~\ref{fig:experiment:imagenet:example:vitl}(b)) than those highlighted by the three baseline methods that do not use any explanation supervision (depicted in Figures~\ref{fig:experiment:imagenet:example:vitl}(c), (d), and (f)).
Also, the model finetuned on ImageNet+X (Figure~\ref{fig:experiment:imagenet:example:vitl}(e)) also generates high-quality explanations, similar to the proposed method, as in the quantitative evaluation.
However, due to its low classification performance shown in Table~\ref{tab:experiment:imagenet:vitl}, it is not practical to use this single model for both classification and explanation simultaneously.
Although the model finetuned on the target dataset with explanation supervision (Figure~\ref{fig:experiment:imagenet:example:vitl}(h)) produces the explanations most similar to the GT ones overall, there are cases where the proposed method produces better explanations, such as in the GTSRB example shown in the second row of Figure~\ref{fig:experiment:imagenet:example:vitl}.

\section{Transferring Explainability Across Source and Target Datasets}\label{sec:appendix:pair}

In this preliminary experiment, we use eight labeled image datasets, namely Cars~\cite{krause20133d}, DTD~\cite{cimpoi2014describing}, EuroSAT~\cite{helber2019eurosat}, GTSRB~\cite{stallkamp2011german}, RESISC45~\cite{cheng2017remote}, SUN397~\cite{xiao2016sun}, SVHN~\cite{netzer2011reading}, and MNIST~\cite{lecun1998mnist}.
We extend each of these datasets to include GT patch-level attributions generated using Kernel SHAP~\cite{Lundberg2017-ii,noauthor_undated-ca}, which explain the classification results of the CLIP image classifier finetuned on each dataset.
Here, the number of perturbations for each image in Kernel SHAP is set to 500.
Similar methods for generating explanation supervision have been used in previous studies~\cite{yang2023efficient,monteiro2024selective}.
We create 56 dataset pairs from the eight datasets, where one dataset in each pair is used as the source dataset and the other as the target dataset.

In the proposed method, we first learn the three types of model parameters $\btheta^{\rm S}_{\rm ft}$, $\btheta^{\rm S}_{\rm ft\star}$, and $\btheta^{\rm T}_{\rm ft}$ for each dataset pair, and then obtain the model parameters $\tilde{\btheta}^{\rm T}_{\rm ft\star}$ by applying the proposed explainability transfer~\bref{eq:proposed:theta_init}.
We evaluate the quality of $\tilde{\btheta}^{\rm T}_{\rm ft\star}$ in terms of the classification accuracy and the explanation error using explanation root mean squared error (E-RMSE), $\sqrt{\set{L}_{\rm exp}(\btheta; \set{D}^{\rm T})}$, where $\set{L}_{\rm exp}$ is defined in~\bref{eq:proposed:exp_loss} and $\set{D}^{\rm T}$ is the target dataset with explanation supervision.

\subsection{Accuracy and Explanation Error Trade-off}\label{sec:appendix:pair:tradeoff}

\begin{figure}[t]
\centering
\includegraphics[width=1.0\textwidth]{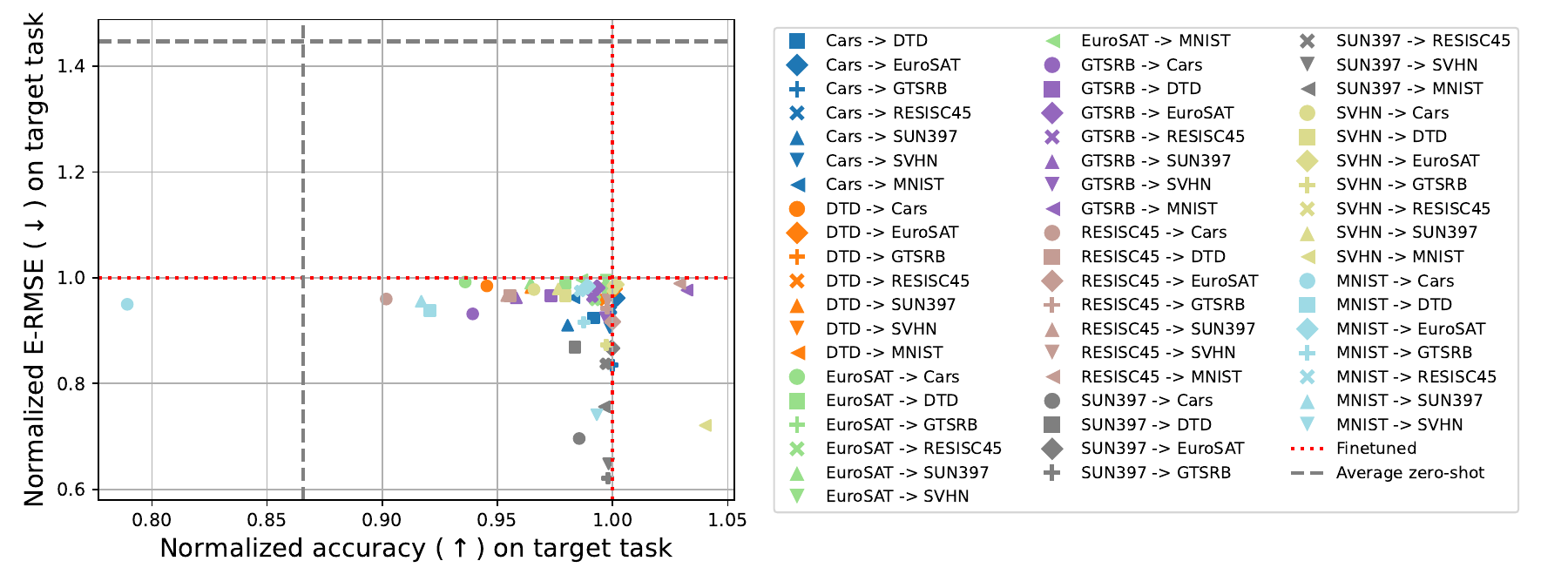}
\vspace{-3mm}
\caption{
Normalized accuracies (larger is better) and normalized E-RMSEs (lower is better) in all the pairs of the eight benchmark datasets when fixed in $\alpha=0.8$, $\lambda_1=1.0$, and $\lambda_2=0.7$.
The color and shape of each marker indicate the source and target domains of the corresponding dataset pair, respectively.
}
\label{fig:appendix:pair_analogy}
\end{figure}

The first result, illustrated in Figure~\ref{fig:appendix:pair_analogy}, shows the normalized accuracies and the normalized E-RMSEs for all the dataset pairs.
Here, the normalized scores are obtained by dividing by the corresponding scores of $\btheta^{\rm T}_{\rm ft}$.
The red dotted line indicates the performances of $\btheta^{\rm T}_{\rm ft}$ (both the accuracy and E-RMSE are one), while the gray dashed line indicates those of the zero-shot classifier parameterized by $\btheta_{\rm base}$.
In successful cases, the E-RMSE decreases from one while the accuracy stays close to one.
The E-RMSE decreases on all 56 pairs, and 55 of them keep at least 90\% of the accuracy.
In particular, the effectiveness when using SUN397 as the source dataset is notable.
This is likely because SUN397 is a large-scale dataset containing diverse scene images, allowing the model to learn generalizable explanatory capabilities for various datasets.
The only pair below 90\% is the transfer from handwritten digit recognition (MNIST) to vehicle type recognition (Cars), which keeps $79\%$ of the accuracy.
When transferring between the two digit recognition tasks (MNIST and SVHN), the accuracy is maintained or improved while the E-RMSE improves substantially.
This suggests that transfer between very different domains can have adverse effects on accuracy.

\subsection{Predictability Analysis on 56 Pairs}\label{sec:appendix:pair:predictor}
The 56 source--target pairs above also provide a held-out test of the predictability finding of Section~\ref{sec:experiment:predict}, independent of the $N=10$ ImageNet-source leave-one-target-out (LOTO) evaluation in the main paper.
We refit the same six-feature regression (three classification-circuit features and three $\btau$-output-shift features, defined in Appendix~\ref{sec:appendix:predict}) with the same definition of $\Delta_q$ (here with E-MSE as the error metric).
We use the same nested-LOTO protocol across the eight target datasets.
Each outer fold withholds one target together with its seven source pairings and trains on the remaining 49 pairs, which gives eight outer folds in total, with seven inner folds for regularization-strength selection.
A simple variance decomposition explains why 6 features alone are insufficient on this larger pair set.
Source identity already accounts for 55.2\% of the variance of $\Delta_{\text{E-MSE}}$ and 53.1\% of $\Delta_{\text{IoU}@10}$.
The per-source mean $\Delta_{\text{E-MSE}}$ ranges from $0.022$ (EuroSAT+X) to $0.391$ (SUN397+X).
No source-universal coefficient vector can fit this $18.1\times$ spread.
We therefore augment the design matrix with seven source dummy variables, which are one-hot indicators of the eight source datasets with Cars+X as the reference.
They act as source fixed effects and give each source its own intercept.
Table~\ref{tab:appendix:pair:predict_56} reports both conditions.
Without source dummies, the six-feature regression collapses on all five explanation-quality metrics (Pearson $|r| \le 0.5$, and $r \le -0.41$ for Lasso on IoU@5 and IoU@10).
With source dummies, five of five explanation-quality metrics achieve held-out $r \in [+0.51, +0.66]$ for both Lasso and Ridge, and the resulting regression reduces held-out MSE by $1.3\times$--$1.8\times$ relative to a predictor that outputs the mean of $\Delta_q$.
On this larger pair set, source-mean variation thus needs to enter the model as an intercept rather than through the six features themselves.

\begin{table}[t]
\centering
\caption{56-pair held-out test of transfer-success predictability (ViT-B/32, $\Delta_q$ target, nested LOTO over the eight target datasets). Each cell reports held-out Pearson $r$. ``No source dummies'' uses the 6 features only. ``With source dummies'' adds the seven source dummy variables as fixed effects.}
\label{tab:appendix:pair:predict_56}
\small
\begin{tabular}{@{}lcccccc@{}}
\toprule
& \multicolumn{2}{c}{\textbf{no source dummies (6 features)}} & & \multicolumn{2}{c}{\textbf{with source dummies (6+7)}} \\
\cmidrule(lr){2-3} \cmidrule(lr){5-6}
Metric & Lasso & Ridge & & Lasso & Ridge \\
\midrule
E-MSE   & $-0.08$ & $+0.04$ & & $+0.59$ & $+0.60$ \\
IoU@1   & $-0.04$ & $+0.09$ & & $+0.65$ & $+0.66$ \\
IoU@5   & $-0.46$ & $-0.07$ & & $+0.55$ & $+0.51$ \\
IoU@10  & $-0.41$ & $-0.00$ & & $+0.60$ & $+0.58$ \\
Rank corr. & $+0.05$ & $+0.07$ & & $+0.60$ & $+0.61$ \\
\bottomrule
\end{tabular}
\end{table}

The model with source dummies evaluated above absorbs per-source intercepts that are observed during training.
Extrapolating to a genuinely unseen source dataset would require either (a) a leave-one-source-out protocol with explicit source-property features, or (b) committing to the 6-feature-only column of Table~\ref{tab:appendix:pair:predict_56} and accepting that source-mean variation is then unmodelled.
We treat this as a limitation of the present regression setup rather than as a conclusion about the underlying signal in the six features.

\paragraph{Bootstrap intervals and leave-one-source-out evaluation}
To make the scope of the predictability finding explicit, we rerun the predictor pipeline behind Table~\ref{tab:appendix:pair:predict_56} with only the splits changed.
First, for the $N = 10$ leave-one-target-out setting of Section~\ref{sec:experiment:predict}, Table~\ref{tab:appendix:pair:loto_ci} adds bootstrap 95\% intervals ($10^4$ resamples of the held-out (actual, predicted) pairs, seed 42) to the held-out correlations.
The intervals exclude zero for E-MSE, Rank corr., IoU@10, and IoU@5 with Lasso, and include zero for IoU@1, matching the significance pattern reported there.
Second, we evaluate leave-one-source-out (LOSO) on the 56 pairs.
It uses eight outer folds, each withholding the seven pairs that share one source dataset, with the regularization strength selected by inner LOSO on the remaining 49 pairs.
Source dummies cannot be used here, since the held-out source has no dummy, so LOSO is the analogue of the no-source-dummies column of Table~\ref{tab:appendix:pair:predict_56}.
Because the seven predictions of a fold share one fitted model, we report a source-level cluster bootstrap interval (resampling whole held-out sources) next to the pair-level interval.
The cluster interval is the appropriate one for this design and is wider.
Table~\ref{tab:appendix:pair:loso} shows the result for the six base features.
Lasso collapses to a constant predictor on all five metrics, and its large negative pooled $r$ is a leave-one-out mean-shift artifact rather than an anti-predictive signal.
Ridge is flat, with every cluster interval including zero, and no cell beats the mean squared error of a constant predictor.
Adding meta-features does not change this.
We try the task-vector cosine $\cos(\btau^{\rm S}_{\rm ft}, \btau^{\rm T}_{\rm ft})$, a class-text similarity between the template-averaged CLIP text embeddings of the source and target class names, and, as an upper-bound probe only, five oracle features that depend on the target's own $\btau^{\rm T}_{\star}$ and are unavailable before transfer.
Across $4$ feature sets $\times$ $5$ metrics $\times$ $2$ regressors, none of the $40$ configurations is simultaneously non-degenerate, significant at $p < 0.05$, positive with both bootstrap intervals excluding zero, and better than the constant predictor.
The predictability claim of this paper is therefore restricted to a fixed source with held-out targets.
Whether held-out-source prediction is learnable with other features remains open.

\begin{table}[t]
\centering
\caption{Held-out Pearson correlation of the transfer-success predictor on ViT-B/32 (LOTO, $N = 10$, $\Delta_q$ target) with bootstrap 95\% intervals ($10^4$ resamples of the held-out predictions). Point estimates match Table~\ref{tab:appendix:predict:vitb_full}.}
\label{tab:appendix:pair:loto_ci}
\small
\begin{tabular}{@{}lcccc@{}}
\toprule
 & \multicolumn{2}{c}{Lasso} & \multicolumn{2}{c}{Ridge} \\
\cmidrule(lr){2-3}\cmidrule(lr){4-5}
Metric & $r$ & 95\% CI & $r$ & 95\% CI \\
\midrule
E-MSE & $+0.92$ & $[+0.84, +0.98]$ & $+0.91$ & $[+0.82, +0.98]$ \\
IoU@1 & $+0.19$ & $[-0.43, +0.55]$ & $+0.33$ & $[-0.02, +0.85]$ \\
IoU@5 & $+0.70$ & $[+0.48, +0.94]$ & $+0.63$ & $[+0.33, +0.91]$ \\
IoU@10 & $+0.81$ & $[+0.66, +0.94]$ & $+0.85$ & $[+0.71, +0.95]$ \\
Rank corr. & $+0.77$ & $[+0.56, +0.92]$ & $+0.82$ & $[+0.69, +0.94]$ \\
\bottomrule
\end{tabular}
\space%
\end{table}

\begin{table}[t]
\centering
\caption{Leave-one-source-out evaluation on the 56 pairs (ViT-B/32, $\Delta_q$ target, six base features, no source dummies). Pooled $r$ over the 56 held-out predictions with pair-level and source-level cluster bootstrap 95\% intervals. ``Degenerate'' marks folds in which Lasso selects no feature and predicts a constant, whose negative pooled $r$ is a leave-one-out mean-shift artifact. No cell beats the mean squared error of a constant predictor.}
\label{tab:appendix:pair:loso}
\footnotesize
\begin{tabular}{@{}llcccl@{}}
\toprule
Metric & Regressor & Pooled $r$ & Pair 95\% CI & Cluster 95\% CI & Note \\
\midrule
E-MSE & Lasso & $-0.74$ & $[-0.88, -0.54]$ & $[-0.85, -0.28]$ & degenerate \\
 & Ridge & $-0.21$ & $[-0.41, -0.01]$ & $[-0.53, +0.15]$ & worse than constant \\
IoU@1 & Lasso & $-0.76$ & $[-0.89, -0.55]$ & $[-0.82, -0.22]$ & degenerate \\
 & Ridge & $-0.16$ & $[-0.34, +0.03]$ & $[-0.47, +0.21]$ & worse than constant \\
IoU@5 & Lasso & $-0.14$ & $[-0.71, -0.01]$ & $[-0.76, -0.02]$ & degenerate \\
 & Ridge & $-0.12$ & $[-0.32, +0.04]$ & $[-0.35, +0.20]$ & worse than constant \\
IoU@10 & Lasso & $-0.72$ & $[-0.87, -0.50]$ & $[-0.80, -0.32]$ & degenerate \\
 & Ridge & $-0.14$ & $[-0.33, +0.03]$ & $[-0.45, +0.14]$ & worse than constant \\
Rank corr. & Lasso & $-0.75$ & $[-0.88, -0.55]$ & $[-0.88, -0.31]$ & degenerate \\
 & Ridge & $-0.26$ & $[-0.46, -0.05]$ & $[-0.59, +0.10]$ & worse than constant \\
\bottomrule
\end{tabular}
\space%
\end{table}

\section{Theoretical Interpretation}\label{sec:appendix:theory}

In this section, we provide a theoretical interpretation of the explainability vectors $\btau^{\rm S}_{\star}$ and $\btau^{\rm T}_{\star}$.

\paragraph{$\btau_{\star}$ as Optimal Perturbation}
First, we show that the explainability vector $\btau^{\rm S}_{\star}$ can be interpreted as the optimal perturbation under a local quadratic approximation around $\btheta^{\rm S}_{\rm ft}$ of the loss function in~\bref{eq:proposed:objective}.
Note that the same derivation can be applied to $\btau^{\rm T}_{\star}$.
The loss function in~\bref{eq:proposed:objective} consists of the classification loss $\set{L}_{\rm cls}$ and the explanation loss $\set{L}_{\rm exp}$.
We perform a local expansion around $\btheta^{\rm S}_{\rm ft}$ for $\set{L}_{\rm cls}$.
Let $\btheta = \btheta^{\rm S}_{\rm ft} + \btau$ where $\btau \in \mathbb{R}^D$ is a small perturbation.
Here, $\bar{\set{D}}^{\rm S} = \{(\bx^{(n)}, y^{(n)})\}_{n=1}^{N^{\rm S}}$ denotes the classification-only subset of $\set{D}^{\rm S}$, i.e., $\set{D}^{\rm S}$ with the GT patch-level attributions $\bphi^{(n)}$ removed.
This allows the same derivation to apply to the target domain, where explanation supervision is unavailable, by replacing $\bar{\set{D}}^{\rm S}$ with $\bar{\set{D}}^{\rm T}$.
Since $\btheta^{\rm S}_{\rm ft}$ approximately satisfies $\nabla \set{L}_{\rm cls}(\btheta^{\rm S}_{\rm ft}; \bar{\set{D}}^{\rm S}) = \bzero$, a second-order Taylor expansion yields
\begin{equation}
\set{L}_{\rm cls}(\btheta^{\rm S}_{\rm ft} + \btau; \bar{\set{D}}^{\rm S}) \approx \set{L}_{\rm cls}(\btheta^{\rm S}_{\rm ft}; \bar{\set{D}}^{\rm S}) + \frac{1}{2} \btau^{\top} \bH^{\rm S} \btau,
\label{eq:appendix:theory:cls_expansion}
\end{equation}
where $\bH^{\rm S} = \nabla^2 \set{L}_{\rm cls}(\btheta^{\rm S}_{\rm ft}; \bar{\set{D}}^{\rm S})$ is the Hessian matrix of $\set{L}_{\rm cls}$ at $\btheta^{\rm S}_{\rm ft}$.
Similarly, we perform a first-order Taylor expansion of $\set{L}_{\rm exp}$ around $\btheta^{\rm S}_{\rm ft}$ as follows:
\begin{equation}
\set{L}_{\rm exp}(\btheta^{\rm S}_{\rm ft} + \btau; \set{D}^{\rm S}) \approx \set{L}_{\rm exp}(\btheta^{\rm S}_{\rm ft}; \set{D}^{\rm S}) + (\bg^{\rm S})^{\top} \btau,
\label{eq:appendix:theory:exp_expansion}
\end{equation}
where $\bg^{\rm S} = \nabla \set{L}_{\rm exp}(\btheta^{\rm S}_{\rm ft}; \set{D}^{\rm S})$ is the gradient vector of $\set{L}_{\rm exp}$ at $\btheta^{\rm S}_{\rm ft}$.
Substituting \bref{eq:appendix:theory:cls_expansion} and \bref{eq:appendix:theory:exp_expansion} into \bref{eq:proposed:objective}, and solving $\btau$ that minimizes the approximated loss function under the assumptions $\alpha > 0$ and a positive definite $\bH^{\rm S}$, which make the minimizer unique, we obtain:
\begin{equation}
\btau^{\rm S}_{\star} \approx - \frac{1-\alpha}{\alpha} (\bH^{\rm S})^{-1} \bg^{\rm S}.
\label{eq:appendix:theory:tau_star}
\end{equation}
The negative gradient $-\bg^{\rm S}$ represents the direction in which the explanation loss $\set{L}_{\rm exp}$ decreases most rapidly around $\btheta^{\rm S}_{\rm ft}$.
However, simply moving $\btheta^{\rm S}_{\rm ft}$ in that direction could significantly degrade classification performance.
The Hessian matrix $\bH^{\rm S}$ represents the curvature of the classification loss $\set{L}_{\rm cls}$, and preconditioning $\bg^{\rm S}$ with its inverse $(\bH^{\rm S})^{-1}$ damps movement along parameter directions where the classification loss has high curvature.
The hyperparameter $\alpha$ controls the magnitude of $\btau^{\rm S}_{\star}$.
As $\alpha$ approaches one, the factor $(1-\alpha)/\alpha$ and hence $\btau^{\rm S}_{\star}$ shrink.
The transferred parameters~\bref{eq:proposed:theta_init} then stay close to $\btheta_{\rm base} + \lambda_1 \btau^{\rm T}_{\rm ft}$, so classification accuracy is prioritized.

\paragraph{Risk of Classification Accuracy Drop}

Because $\btheta^{\rm T}_{\rm ft}$ is a local minimum of the target classification loss $\set{L}_{\rm cls}$, no parameter shift, including the addition of the source-domain explainability vector $\btau^{\rm S}_{\star}$, decreases the classification loss under the local quadratic approximation.
The increase comes from the components of the shift along high-curvature directions of the target loss.
Formally, based on the approximated classification loss \bref{eq:appendix:theory:cls_expansion}, the increase in classification loss when adding the source-domain explainability vector $\btau^{\rm S}_{\star}$ to the target-domain parameters $\btheta^{\rm T}_{\rm ft}$ is proportional to
\begin{equation}
R(\btau^{\rm S}_{\star}; \bH^{\rm T})
= (\btau^{\rm S}_{\star})^{\top} \bH^{\rm T} \btau^{\rm S}_{\star},
\label{eq:appendix:theory:risk}
\end{equation}
which equals twice the second-order term $\frac{1}{2} (\btau^{\rm S}_{\star})^{\top} \bH^{\rm T} \btau^{\rm S}_{\star}$ in the local quadratic approximation around $\btheta^{\rm T}_{\rm ft}$.
With $\lambda_1 = 1$ in~\bref{eq:proposed:theta_init}, the transferred parameters are $\btheta^{\rm T}_{\rm ft} + \lambda_2 \btau^{\rm S}_{\star}$.
The increase is then $\frac{\lambda_2^2}{2} R(\btau^{\rm S}_{\star}; \bH^{\rm T})$, so $\lambda_2$ scales this risk quadratically.
To understand the property of this function, we perform eigenvalue decomposition of $\bH^{\rm T}$ as
\begin{align}
\bH^{\rm T} = \bU \, \mathrm{diag}(\nu_1, \nu_2, \ldots, \nu_D) \, \bU^{\top},\\
(\nu_1 \geq \nu_2 \geq \cdots \geq \nu_D \geq 0). \nonumber
\end{align}
Here, $\bU = [\bu_1, \bu_2, \ldots, \bu_D]$ is the matrix of orthonormal eigenvectors of $\bH^{\rm T}$, and $\nu_1, \ldots, \nu_D$ are the corresponding eigenvalues.
Expanding $\btau^{\rm S}_{\star}$ in the eigenvector basis of $\bH^{\rm T}$ yields
\begin{equation}
\btau^{\rm S}_{\star} = \sum_{i=1}^{D} a_i \bu_i,\quad a_i = (\btau^{\rm S}_{\star})^{\top} \bu_i.
\end{equation}
Then \bref{eq:appendix:theory:risk} can be rewritten as
\begin{equation}
R(\btau^{\rm S}_{\star}; \bH^{\rm T}) = \sum_{i=1}^{D} \nu_i a_i^2.
\end{equation}
This expression indicates that if $\btau^{\rm S}_{\star}$ has a large component in the direction of the eigenvectors associated with large eigenvalues of $\bH^{\rm T}$, the increase in classification loss $R(\btau^{\rm S}_{\star}; \bH^{\rm T})$ will be significant.

If a precomputed Hessian matrix $\bH^{\rm T}$ is available, we can select the explainability vector $\btau^{\rm S}_{\star}$ that minimizes $R(\btau^{\rm S}_{\star}; \bH^{\rm T})$ among candidates learned in different source domains, without requiring any target-domain explanation supervision (the Hessian itself is precomputed once on $\bar{\set{D}}^{\rm T}$).
However, storing the precomputed $\bH^{\rm T}$, a dense matrix whose size scales quadratically with the number of model parameters, is infeasible.
In practice, it is therefore reasonable to approximate it with a diagonal matrix.

\section{Results on Comparison with Kernel SHAP}\label{sec:appendix:vs_shap}

\begin{figure}[t]
\centering
\begin{tabular}{@{}cc@{}}
\includegraphics[width=0.30\columnwidth]{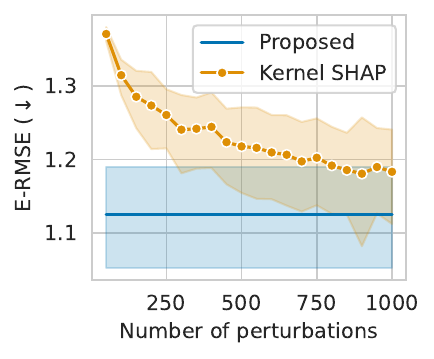} &
\includegraphics[width=0.30\columnwidth]{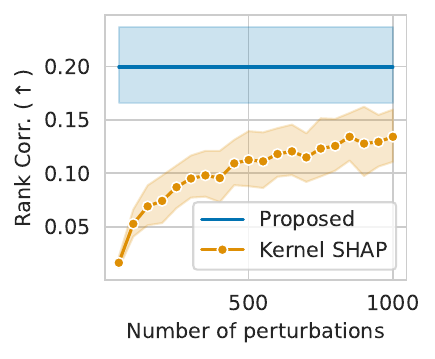} \\[-2mm]
\footnotesize (a) E-RMSE & \footnotesize (b) Rank corr.
\end{tabular}\\[2mm]
\begin{tabular}{@{}ccc@{}}
\includegraphics[width=0.30\columnwidth]{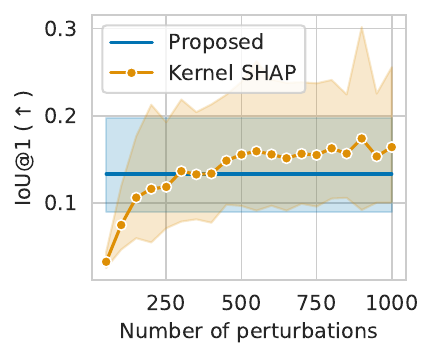} &
\includegraphics[width=0.30\columnwidth]{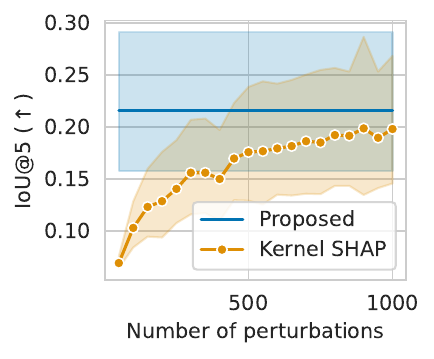} &
\includegraphics[width=0.30\columnwidth]{fig/analogy_imagenet_vs_shap/analogy_imagenet_vs_shap_iou_10.pdf} \\[-2mm]
\footnotesize (c) IoU@1 & \footnotesize (d) IoU@5 & \footnotesize (e) IoU@10
\end{tabular}
\caption{Explainability scores of the proposed method and Kernel SHAP averaged across the target datasets with varying number of perturbations.
The shaded area represents the 95\% confidence interval.
}
\label{fig:appendix:imagenet:vsshap}
\end{figure}

Figure~\ref{fig:appendix:imagenet:vsshap} shows the explainability scores of the proposed method and Kernel SHAP with varying number of perturbations $P$ when transferring explainability learned on ImageNet+X to the 10 target datasets.
Kernel SHAP is applied to $\btheta^{\rm T}_{\rm ft}$ with perturbation samples independent of those used for the GT, which is itself SHAP-500 (Section~\ref{sec:experiment:imagenet:vsshap}).
For Rank corr., the proposed method remains clearly higher than Kernel SHAP even at 1,000 perturbations ($0.200$ vs.\ $0.134$).
For IoU@5 and IoU@10, the proposed method is higher at every number of perturbations, but the gap narrows to $0.02$ and $0.05$ at 1,000 perturbations and the 95\% confidence intervals overlap for large numbers of perturbations.
For E-RMSE, the proposed method is lower than Kernel SHAP at every number of perturbations, but the gap narrows to $0.06$ at 1,000 perturbations and the 95\% confidence intervals overlap for large numbers of perturbations.
For IoU@1, the proposed method matches Kernel SHAP with about 300 perturbations, and Kernel SHAP is higher with more perturbations.

\section{Discussion on Computational Complexity of Explanation Generation}\label{sec:appendix:complexity}
As shown in Figure~\ref{fig:model}, the difference between our self-explaining model, i.e., ``prediction-with-explanation'' model, and the ``prediction-only'' model used when generating patch-level attributions with Kernel SHAP is that the former applies the domain-specific head $g_{\bW}$ to both the $\mathtt{[CLS]}$ token and the $M$ image patches, while the latter applies it only to the $\mathtt{[CLS]}$ token.
The domain-specific head $g_{\bW}$ calculates matrix multiplication $\bW^{\top}\bu$ for input feature vector $\bu \in \mathbb{R}^{K}$ and weight matrix $\bW \in \mathbb{R}^{K \times C}$, where $C$ is the number of classes.
The computational complexity of this matrix multiplication is $\mathcal{O}(KC)$.
Denoting the computational complexity of the feature extractor $f_{\btheta}$ as $F$, the computational complexity of our self-explaining image classifier per explanation is $\mathcal{O}(F + MKC)$ where $M$ is the number of patches in an image.
In practice, since $MKC$ is typically much smaller than $F$, the computational complexity can be approximated as $\mathcal{O}(F)$.

In contrast, the computational complexity of Kernel SHAP per explanation is $\mathcal{O}\left(PF + PM^2\right)$, where $P$ is the number of perturbations and $PM^2$ is associated with solving the least-squares problem.
In practice, since $PF$ is much larger than $PM^2$, the computational complexity can be approximated as $\mathcal{O}(PF)$.

In summary, the per-explanation complexities of our self-explaining image classifier and Kernel SHAP both scale linearly with the feature-extractor cost $F$, but Kernel SHAP multiplies it by $P$.

\paragraph{Costs and amortization}
Table~\ref{tab:appendix:complexity:cost} attaches measured and estimated costs to this analysis (CLIP ViT-B/32, one NVIDIA RTX 6000 Ada GPU unless noted).
The source side is a one-time cost.
Generating the Kernel SHAP supervision for the $1.28$M training images of ImageNet-1k at $P = 500$ takes $91.4$ single-GPU hours, extrapolated from the measured $0.257$ s per image at $P = 500$.
The original run took about $29.2$ hours of wall-clock time, reconstructed from file timestamps, on an unrecorded number of GPUs.
Training $\btheta^{\rm S}_{\rm ft}$ and $\btheta^{\rm S}_{\rm ft\star}$ takes $9.0$ GPU-hours, estimated as twice the $4.5$ GPU-hours measured for one training run with the identical protocol.
The source side thus costs $100.4$ GPU-hours in total.
As with other pretrained transferable explainers, this cost is borne once by the provider of the explainability vector.
Transferring the vector to a target is a parameter addition that takes under a second on a CPU, and the target side pays no training cost.
At inference, our single joint forward pass produces the prediction and the explanation in about $0.39$ ms per image (a proxy measured on the single forward pass of $\btheta^{\rm T}_{\rm ft}$, which has the same architecture), against $257$ ms for Kernel SHAP with $P = 500$ on $\btheta^{\rm T}_{\rm ft}$ (a ratio of about $664\times$, larger than the asymptotic $P = 500$, whose causes we do not separate), $0.8$ ms for the amortized explainer of Appendix~\ref{sec:appendix:alternatives} (two forwards), and $7.2$ ms for TVE (three forwards).
These are per-image averages under each evaluation's own batch setting (batched forwards for the proposed method and the amortized explainer, one image at a time for Kernel SHAP), not single-image latencies, so the ratio depends on these conditions.
The one-time cost breaks even against running Kernel SHAP per explanation after $1.41$M served explanations in total.
For $T = 10$ targets this is about $141$k explanations per target, or $1.6$ days if each target serves one image per second, after which every further explanation is essentially free relative to a prediction.
At the scale of our evaluation alone ($10$ targets $\times$ $500$ images for the faithfulness metrics) the source cost is not recouped.
The argument for amortization is a deployment-scale one.

\begin{table}[t]
\centering
\caption{Costs for CLIP ViT-B/32 (NVIDIA RTX 6000 Ada). GPU-hours for the source side are extrapolated or estimated single-GPU equivalents (see text). Per-image times are for prediction plus explanation, averaged per image under each evaluation's batch setting (not single-image latencies), and the time of the proposed method is a proxy.}
\label{tab:appendix:complexity:cost}
\footnotesize
\begin{tabular}{@{}lp{0.56\textwidth}r@{}}
\toprule
Stage & Item & Cost \\
\midrule
Source (one-time) & Kernel SHAP supervision for the ImageNet-1k training set ($1.28$M images, $P = 500$) & $91.4$ GPU-h \\
 & Training $\btheta^{\rm S}_{\rm ft}$ and $\btheta^{\rm S}_{\rm ft\star}$ & $9.0$ GPU-h \\
 & Total & $100.4$ GPU-h \\
\midrule
Per target & Transfer by parameter addition~\bref{eq:proposed:theta_init} (CPU) & $0.8$ s \\
\midrule
Per image & Proposed (one forward, prediction and explanation) & $0.39$ ms \\
 & $\btheta^{\rm T}_{\rm ft}$ + amortized explainer (two forwards) & $0.8$ ms \\
 & TVE (three forwards) & $7.2$ ms \\
 & $\btheta^{\rm T}_{\rm ft}$ + Kernel SHAP, $P = 500$ ($\approx 500$ forwards) & $257$ ms \\
\midrule
Break-even vs.\ Kernel SHAP & Served explanations in total ($T = 10$: per target) & $1.41$M ($141$k) \\
\bottomrule
\end{tabular}
\end{table}
\space%

\section{Faithfulness of the Transferred Self-Explanations}\label{sec:appendix:faithfulness}

This appendix gives the procedure and the per-target results of the within-model faithfulness evaluation summarized in Section~\ref{sec:experiment:imagenet:faithfulness}.
GT agreement (Table~\ref{tab:experiment:imagenet}) measures how well the transferred model reproduces the explanation target chosen by the user.
Faithfulness asks a different question, namely whether the patches highlighted by the transferred model matter for that model's own predictions.
Since faithfulness is a property of an (explanation, model) pair, every attribution below is scored on the predicted probabilities of the same transferred model $\tilde{\btheta}^{\rm T}_{\rm ft\star}$.

\paragraph{Target set}
Appendices~\ref{sec:appendix:faithfulness}, \ref{sec:appendix:controls}, \ref{sec:appendix:lambda2}, and~\ref{sec:appendix:circuit_overlap:mechanism}, and Appendix~\ref{sec:appendix:alternatives} except Table~\ref{tab:appendix:alternatives:amortized}, use a 10-target set in which CIFAR10 replaces SUN397 of Table~\ref{tab:experiment:imagenet}.
The nine shared targets use the same checkpoints and the same evaluation pipeline as Table~\ref{tab:experiment:imagenet}.
On this set, the mean accuracy of $\btheta^{\rm T}_{\rm ft}$ is $0.914$ and that of the transferred model is $0.911$, which is why 10-target means in these appendices differ from those of Table~\ref{tab:experiment:imagenet}.
We refer to it as the CIFAR10 set where the distinction matters.

\paragraph{Procedure}
For each target we draw $500$ test images by class-proportional stratified sampling (seed $42$, at least one image per class) and keep the images that $\tilde{\btheta}^{\rm T}_{\rm ft\star}$ classifies correctly ($N = 338$--$492$ per target, Table~\ref{tab:appendix:faithfulness:full}), so that the attribution for the true class $y$ is also that for the predicted class.
Four attributions are compared:
(i) the self-explanation, i.e., the attribution $\tilde{\bPhi}_{c}$ for the predicted class $c$, obtained in the same forward pass as the prediction;
(ii) Kernel SHAP on $\tilde{\btheta}^{\rm T}_{\rm ft\star}$ itself with $P = 500$ perturbations, following the GT-generation protocol of Section~\ref{sec:experiment:imagenet} (the model-specific reference);
(iii) the GT attribution of Section~\ref{sec:experiment:imagenet}, i.e., Kernel SHAP applied to the separate model $\btheta^{\rm T}_{\rm ft}$;
and (iv) a random attribution (standard normal entries, seed $42$).
Each attribution is scored with four metrics computed on the predicted probability of $\tilde{\btheta}^{\rm T}_{\rm ft\star}$ for the predicted class.
The first two are Fidelity$+$ and Fidelity$-$ adapted from the fidelity metrics of TVE~\cite{wang2024tve}, i.e., the probability drop when the top-$k\%$ patches are masked (higher is better) and when the bottom-$k\%$ patches are masked (lower is better), averaged over $k \in \{10, 20, 30, 40, 50\}$.
The other two are deletion and insertion scores in the style of \citet{petsiuk2018rise}, i.e., the probability averaged over the steps at which the top $1, 11, 21, \ldots$ patches by attribution (every $10$ patches below the number of patches) are removed (lower is better) or are the only patches kept (higher is better).
For CLIP ViT-B/32 with $49$ patches these are five steps ($1, 11, 21, 31, 41$), and for the $196$-patch grid of TVE in Appendix~\ref{sec:appendix:alternatives} they are $20$ steps ($1, 11, \ldots, 191$).
Masked patches are set to zero in the normalized input, and unlike the original AUC the two endpoints of the curve are not included.

\paragraph{Results}
Table~\ref{tab:appendix:faithfulness:full} lists the raw values.
The self-explanation matches or exceeds the reference on $36$ of the $40$ target--metric comparisons and beats random on all $40$.
Comparisons use unrounded values, so the two cells that tie at three decimals in Table~\ref{tab:appendix:faithfulness:full} count as a match (Del on Cars) and a miss (Ins on EuroSAT).
The random attribution is the worst row on every metric for every target, and the GT attribution of the separate model $\btheta^{\rm T}_{\rm ft}$ also beats random everywhere.
Kernel SHAP with $P = 500$ is a finite-sample Monte Carlo estimate rather than a faithfulness ceiling, so a single-forward-pass explanation can exceed it.
The result thus indicates that, on $36$ of the $40$ comparisons, the transferred self-explanation is at least as faithful to the model it accompanies as the reference, with one forward pass instead of $P = 500$.
On the unstratified first $500$ test images per target, the self-explanation matches or exceeds the reference on $31$ of the $36$ comparisons for the nine targets other than EuroSAT and beats random on $36$ of $40$.
All four misses against random are on EuroSAT, whose first $500$ test images contain only two of its 10 classes, which is why we use stratified sampling.

\begin{table}[t]
\centering
\caption{Raw faithfulness scores on the transferred model $\tilde{\btheta}^{\rm T}_{\rm ft\star}$ (CIFAR10 set, class-stratified $500$-image subsets restricted to correctly classified images, with $N$ in parentheses). All four attributions are scored on the predicted probabilities of $\tilde{\btheta}^{\rm T}_{\rm ft\star}$. Best value per target and metric in bold.}
\label{tab:appendix:faithfulness:full}
\footnotesize
\begin{tabular}{@{}llcccc@{}}
\toprule
Dataset ($N$) & Attribution & Fid$+$ ($\uparrow$) & Fid$-$ ($\downarrow$) & Ins ($\uparrow$) & Del ($\downarrow$) \\
\midrule
Cars (338) & Self-explanation (ours) & $\mathbf{0.820}$ & $\mathbf{0.094}$ & $\mathbf{0.493}$ & $0.148$ \\
 & Kernel SHAP on $\tilde{\btheta}^{\rm T}_{\rm ft\star}$ & $0.789$ & $0.255$ & $0.353$ & $0.148$ \\
 & GT (Kernel SHAP on $\btheta^{\rm T}_{\rm ft}$) & $0.799$ & $0.173$ & $0.422$ & $\mathbf{0.147}$ \\
 & Random & $0.508$ & $0.515$ & $0.175$ & $0.311$ \\
\midrule
CIFAR10 (492) & Self-explanation (ours) & $\mathbf{0.409}$ & $\mathbf{0.029}$ & $\mathbf{0.737}$ & $\mathbf{0.492}$ \\
 & Kernel SHAP on $\tilde{\btheta}^{\rm T}_{\rm ft\star}$ & $0.316$ & $0.046$ & $0.677$ & $0.553$ \\
 & GT (Kernel SHAP on $\btheta^{\rm T}_{\rm ft}$) & $0.280$ & $0.049$ & $0.677$ & $0.585$ \\
 & Random & $0.147$ & $0.140$ & $0.555$ & $0.682$ \\
\midrule
CIFAR100 (445) & Self-explanation (ours) & $\mathbf{0.570}$ & $\mathbf{0.110}$ & $\mathbf{0.575}$ & $\mathbf{0.351}$ \\
 & Kernel SHAP on $\tilde{\btheta}^{\rm T}_{\rm ft\star}$ & $0.531$ & $0.115$ & $0.551$ & $0.381$ \\
 & GT (Kernel SHAP on $\btheta^{\rm T}_{\rm ft}$) & $0.450$ & $0.118$ & $0.536$ & $0.426$ \\
 & Random & $0.262$ & $0.255$ & $0.399$ & $0.538$ \\
\midrule
DTD (395) & Self-explanation (ours) & $0.343$ & $\mathbf{0.147}$ & $\mathbf{0.532}$ & $\mathbf{0.482}$ \\
 & Kernel SHAP on $\tilde{\btheta}^{\rm T}_{\rm ft\star}$ & $\mathbf{0.351}$ & $0.174$ & $0.501$ & $0.496$ \\
 & GT (Kernel SHAP on $\btheta^{\rm T}_{\rm ft}$) & $0.318$ & $0.163$ & $0.518$ & $0.512$ \\
 & Random & $0.260$ & $0.247$ & $0.432$ & $0.561$ \\
\midrule
EuroSAT (491) & Self-explanation (ours) & $\mathbf{0.181}$ & $0.067$ & $0.672$ & $\mathbf{0.698}$ \\
 & Kernel SHAP on $\tilde{\btheta}^{\rm T}_{\rm ft\star}$ & $0.156$ & $0.044$ & $0.672$ & $0.714$ \\
 & GT (Kernel SHAP on $\btheta^{\rm T}_{\rm ft}$) & $0.166$ & $\mathbf{0.033}$ & $\mathbf{0.699}$ & $0.701$ \\
 & Random & $0.076$ & $0.085$ & $0.632$ & $0.765$ \\
\midrule
GTSRB (491) & Self-explanation (ours) & $\mathbf{0.893}$ & $\mathbf{0.006}$ & $\mathbf{0.789}$ & $\mathbf{0.211}$ \\
 & Kernel SHAP on $\tilde{\btheta}^{\rm T}_{\rm ft\star}$ & $0.798$ & $0.034$ & $0.717$ & $0.260$ \\
 & GT (Kernel SHAP on $\btheta^{\rm T}_{\rm ft}$) & $0.781$ & $0.035$ & $0.736$ & $0.265$ \\
 & Random & $0.287$ & $0.266$ & $0.415$ & $0.543$ \\
\midrule
MNIST (446) & Self-explanation (ours) & $\mathbf{0.768}$ & $\mathbf{0.057}$ & $\mathbf{0.664}$ & $\mathbf{0.280}$ \\
 & Kernel SHAP on $\tilde{\btheta}^{\rm T}_{\rm ft\star}$ & $0.665$ & $0.250$ & $0.460$ & $0.323$ \\
 & GT (Kernel SHAP on $\btheta^{\rm T}_{\rm ft}$) & $0.661$ & $0.212$ & $0.493$ & $0.320$ \\
 & Random & $0.418$ & $0.430$ & $0.326$ & $0.449$ \\
\midrule
RESISC45 (482) & Self-explanation (ours) & $\mathbf{0.468}$ & $0.077$ & $\mathbf{0.598}$ & $\mathbf{0.434}$ \\
 & Kernel SHAP on $\tilde{\btheta}^{\rm T}_{\rm ft\star}$ & $0.384$ & $\mathbf{0.071}$ & $0.597$ & $0.494$ \\
 & GT (Kernel SHAP on $\btheta^{\rm T}_{\rm ft}$) & $0.335$ & $0.083$ & $0.592$ & $0.522$ \\
 & Random & $0.179$ & $0.186$ & $0.475$ & $0.612$ \\
\midrule
STL10 (492) & Self-explanation (ours) & $\mathbf{0.439}$ & $\mathbf{0.027}$ & $\mathbf{0.743}$ & $\mathbf{0.473}$ \\
 & Kernel SHAP on $\tilde{\btheta}^{\rm T}_{\rm ft\star}$ & $0.287$ & $0.065$ & $0.658$ & $0.579$ \\
 & GT (Kernel SHAP on $\btheta^{\rm T}_{\rm ft}$) & $0.325$ & $0.038$ & $0.699$ & $0.556$ \\
 & Random & $0.147$ & $0.145$ & $0.555$ & $0.676$ \\
\midrule
SVHN (481) & Self-explanation (ours) & $\mathbf{0.710}$ & $\mathbf{0.032}$ & $\mathbf{0.727}$ & $\mathbf{0.323}$ \\
 & Kernel SHAP on $\tilde{\btheta}^{\rm T}_{\rm ft\star}$ & $0.545$ & $0.040$ & $0.657$ & $0.399$ \\
 & GT (Kernel SHAP on $\btheta^{\rm T}_{\rm ft}$) & $0.582$ & $0.039$ & $0.683$ & $0.380$ \\
 & Random & $0.190$ & $0.185$ & $0.481$ & $0.598$ \\
\bottomrule
\end{tabular}
\space%
\end{table}
\space%
\section{Control-Vector Ablations}\label{sec:appendix:controls}

This appendix details the control experiment summarized in Table~\ref{tab:experiment:controls}, which asks whether the subtraction $\btau^{\rm S}_{\star} = \btheta^{\rm S}_{\rm ft\star} - \btheta^{\rm S}_{\rm ft}$ specifically isolates explanation ability, or whether a similar effect could be produced by a perturbation of the same magnitude, by source-specific optimization effects, or by an attribution prior that does not depend on the content of the explanation supervision.
Every control vector $\btau_{\rm ctrl}$ enters the transfer rule~\bref{eq:proposed:theta_init} in place of $\btau^{\rm S}_{\star}$ with the same coefficients ($\lambda_1 = 1.0$, $\lambda_2 = 0.7$), and the resulting model is evaluated with the pipeline of Table~\ref{tab:experiment:imagenet} (full test split, GT-agreement metrics) on the CIFAR10 set (Appendix~\ref{sec:appendix:faithfulness}).

\paragraph{Control vectors}
\begin{itemize}
  \setlength{\itemsep}{1pt}
  \item[(a)] Equal-norm random: a Gaussian random direction over all parameters, rescaled so that its global $\ell_2$ norm equals $\|\btau^{\rm S}_{\star}\| = 10.48$ (seed $42$). Any effect would come from the magnitude of the perturbation alone.
  \item[(b)] Layer-restricted: $\btau^{\rm S}_{\star}$ restricted to transformer blocks $1$--$6$ (front-6) or $7$--$12$ (back-6), with all other parameters set to zero and no rescaling (norms $6.58$ and $8.07$). This localizes where the effect resides.
  \item[(c)] Full finetuning vector, raw: $\btheta^{\rm S}_{\rm ft\star} - \btheta_{\rm base}$, i.e., the source vector without the subtraction of $\btau^{\rm S}_{\rm ft}$ (norm $26.6$, $2.5\times$ that of $\btau^{\rm S}_{\star}$). If the subtraction were unnecessary, this vector should behave like $\btau^{\rm S}_{\star}$.
  \item[(d)] Shuffled supervision: the explanation maps of ImageNet+X are permuted across images (seed $42$), which destroys the correspondence between an image and its explanation while preserving the distribution and format of the maps.
The source model is retrained on this supervision with the same hyperparameters and initialization as $\btheta^{\rm S}_{\rm ft\star}$, and the control vector is $\btheta^{\rm S\prime}_{\rm ft\star} - \btheta^{\rm S}_{\rm ft}$. Whatever this vector transfers is a content-free attribution prior. This control is a single retraining with one permutation seed.
  \item[(e)] No vector: $\btau_{\rm ctrl} = \bzero$, i.e., $\btheta^{\rm T}_{\rm ft}$ itself, the baseline for all control vectors.
  \item[(f)] Full finetuning vector, norm-matched: (c) rescaled to $\|\btau^{\rm S}_{\star}\|$, which separates the direction of (c) from its magnitude.
\end{itemize}

\paragraph{Results}
Table~\ref{tab:appendix:controls:means} reports the 10-target means on all five metrics.
Table~\ref{tab:experiment:controls} in the main text shows three of them for six of the eight vectors.
Table~\ref{tab:appendix:controls:full} lists the per-target results.

\begin{table}[t]
\caption{Control vectors: 10-target means on the CIFAR10 set (full test split). Each row adds the indicated vector to $\btheta_{\rm base} + \btau^{\rm T}_{\rm ft}$ with $\lambda_2 = 0.7$. The first row is the proposed method. Labels (a)--(f) refer to the control vectors defined above.}
\label{tab:appendix:controls:means}
\centering
\footnotesize
\begin{tabular}{@{}lccccc@{}}
\toprule
Added vector & Acc.\ ($\uparrow$) & E-RMSE ($\downarrow$) & Rank corr.\ ($\uparrow$) & IoU@1 ($\uparrow$) & IoU@10 ($\uparrow$) \\
\midrule
Proposed ($\btau^{\rm S}_{\star}$) & $0.911$ & $1.118$ & $0.204$ & $0.132$ & $0.281$ \\
\midrule
(e) No vector & $0.914$ & $1.563$ & $-0.147$ & $0.009$ & $0.077$ \\
(a) Equal-norm random & $0.909$ & $1.560$ & $-0.144$ & $0.010$ & $0.078$ \\
(b) Front-6 blocks & $0.912$ & $1.563$ & $-0.147$ & $0.009$ & $0.077$ \\
(b) Back-6 blocks & $0.913$ & $1.189$ & $0.148$ & $0.109$ & $0.244$ \\
(c) Full FT, raw & $0.872$ & $1.109$ & $0.215$ & $0.132$ & $0.283$ \\
(f) Full FT, norm-matched & $0.911$ & $1.327$ & $0.046$ & $0.053$ & $0.167$ \\
(d) Shuffled supervision & $0.909$ & $1.154$ & $0.190$ & $0.079$ & $0.264$ \\
\bottomrule
\end{tabular}
\space%
\end{table}

We define the recovery of a control vector on a metric as its improvement over no vector (e) divided by the improvement of $\btau^{\rm S}_{\star}$ over (e), both computed on unrounded 10-target means.
The control vectors behave as follows.
\begin{itemize}
  \setlength{\itemsep}{1pt}
  \item The equal-norm random vector (a) and the front-6 vector (b) are indistinguishable from no vector (e) on every metric.
A perturbation of the same norm, or of the right direction confined to the first six blocks, contributes no explanation-specific signal. Note that $0.7\|\btau^{\rm S}_{\star}\| = 7.3$ is $2.4$--$3.0\times$ the norm of the targets' own task vectors $\btau^{\rm T}_{\rm ft}$ (measured on Cars, GTSRB, and RESISC45), so this is not a small perturbation.
  \item The back-6 vector (b) recovers $84.0\%$ of the Rank corr.\ improvement, $81.5\%$ of the IoU@1 improvement, and $81.9\%$ of the IoU@10 improvement. The per-target recovery is not uniform.
MNIST is the clear outlier ($58\%$ Rank corr.\ and $51\%$ IoU@10), while the other nine targets lie between $75\%$ and $95\%$ on Rank corr. Together with the front-6 result, this is a causal localization of the transferred effect in the later blocks (Appendix~\ref{sec:appendix:circuit_overlap:mechanism}).
  \item The raw full finetuning vector (c) explains as well as $\btau^{\rm S}_{\star}$ (its 10-target means are within $0.011$ of the proposed method on every explanation metric) but costs $3.9$ points of accuracy on average, concentrated on Cars ($-19.6$ points). Norm-matching it (f) removes the accuracy loss but collapses explanation quality (Rank corr.\ $0.046$ vs.\ $0.204$). The explanation effect of the raw vector therefore rides on its surplus norm, and the subtraction of $\btau^{\rm S}_{\rm ft}$ extracts the explanation component at an accuracy-preserving norm.
  \item The shuffled-supervision vector (d) recovers most of the coarse ranking structure ($96\%$ of the Rank corr.\ and $92\%$ of the IoU@10 improvement), so a considerable part of the transferred effect is a content-free attribution prior. Content specificity shows up at the top-1 patch.
The vector recovers only $57.4\%$ of the IoU@1 improvement and is below $\btau^{\rm S}_{\star}$ on IoU@1 for $9$ of the $10$ targets (e.g., Cars $0.062$ vs.\ $0.169$, GTSRB $0.219$ vs.\ $0.366$). It ties or exceeds $\btau^{\rm S}_{\star}$ on Rank corr.\ or IoU@10 on three targets (DTD, GTSRB, and SVHN).
\end{itemize}

\begin{table}[t]
\centering
\caption{Per-target results of the control-vector ablation (CIFAR10 set, full test split, $\lambda_1 = 1.0$, $\lambda_2 = 0.7$). Each row adds the indicated vector to $\btheta_{\rm base} + \btau^{\rm T}_{\rm ft}$. The labels (a)--(f) refer to the list in the text. 10-target means are in Table~\ref{tab:appendix:controls:means}.}
\label{tab:appendix:controls:full}
\scriptsize
\begin{tabular}{@{}llccccc@{}}
\toprule
Dataset & Vector & Accuracy ($\uparrow$) & E-RMSE ($\downarrow$) & Rank corr.\ ($\uparrow$) & IoU@1 ($\uparrow$) & IoU@10 ($\uparrow$) \\
\midrule
Cars & Proposed ($\btau^{\rm S}_{\star}$) & $0.702$ & $1.068$ & $0.249$ & $0.169$ & $0.350$ \\
 & (e) No vector & $0.729$ & $1.618$ & $-0.195$ & $0.004$ & $0.053$ \\
 & (a) Equal-norm random & $0.713$ & $1.616$ & $-0.194$ & $0.004$ & $0.054$ \\
 & (b) Front-6 blocks & $0.717$ & $1.616$ & $-0.191$ & $0.004$ & $0.055$ \\
 & (b) Back-6 blocks & $0.720$ & $1.101$ & $0.227$ & $0.137$ & $0.328$ \\
 & (c) Full FT, raw & $0.506$ & $1.068$ & $0.250$ & $0.155$ & $0.348$ \\
 & (f) Full FT, norm-matched & $0.706$ & $1.269$ & $0.111$ & $0.062$ & $0.216$ \\
 & (d) Shuffled supervision & $0.704$ & $1.153$ & $0.207$ & $0.062$ & $0.286$ \\
\midrule
CIFAR10 & Proposed ($\btau^{\rm S}_{\star}$) & $0.976$ & $1.184$ & $0.172$ & $0.072$ & $0.221$ \\
 & (e) No vector & $0.976$ & $1.532$ & $-0.124$ & $0.012$ & $0.089$ \\
 & (a) Equal-norm random & $0.974$ & $1.531$ & $-0.123$ & $0.014$ & $0.089$ \\
 & (b) Front-6 blocks & $0.977$ & $1.533$ & $-0.126$ & $0.012$ & $0.089$ \\
 & (b) Back-6 blocks & $0.976$ & $1.235$ & $0.130$ & $0.059$ & $0.200$ \\
 & (c) Full FT, raw & $0.969$ & $1.173$ & $0.185$ & $0.077$ & $0.222$ \\
 & (f) Full FT, norm-matched & $0.977$ & $1.351$ & $0.031$ & $0.036$ & $0.146$ \\
 & (d) Shuffled supervision & $0.974$ & $1.226$ & $0.148$ & $0.044$ & $0.193$ \\
\midrule
CIFAR100 & Proposed ($\btau^{\rm S}_{\star}$) & $0.878$ & $1.223$ & $0.143$ & $0.099$ & $0.222$ \\
 & (e) No vector & $0.877$ & $1.525$ & $-0.105$ & $0.011$ & $0.090$ \\
 & (a) Equal-norm random & $0.874$ & $1.524$ & $-0.104$ & $0.011$ & $0.090$ \\
 & (b) Front-6 blocks & $0.879$ & $1.525$ & $-0.105$ & $0.011$ & $0.091$ \\
 & (b) Back-6 blocks & $0.880$ & $1.256$ & $0.116$ & $0.081$ & $0.205$ \\
 & (c) Full FT, raw & $0.844$ & $1.218$ & $0.146$ & $0.103$ & $0.223$ \\
 & (f) Full FT, norm-matched & $0.879$ & $1.336$ & $0.054$ & $0.051$ & $0.161$ \\
 & (d) Shuffled supervision & $0.874$ & $1.272$ & $0.111$ & $0.060$ & $0.190$ \\
\midrule
DTD & Proposed ($\btau^{\rm S}_{\star}$) & $0.769$ & $1.242$ & $0.141$ & $0.048$ & $0.172$ \\
 & (e) No vector & $0.778$ & $1.503$ & $-0.103$ & $0.011$ & $0.085$ \\
 & (a) Equal-norm random & $0.770$ & $1.499$ & $-0.098$ & $0.016$ & $0.086$ \\
 & (b) Front-6 blocks & $0.770$ & $1.499$ & $-0.098$ & $0.010$ & $0.087$ \\
 & (b) Back-6 blocks & $0.773$ & $1.289$ & $0.102$ & $0.039$ & $0.157$ \\
 & (c) Full FT, raw & $0.741$ & $1.232$ & $0.151$ & $0.054$ & $0.177$ \\
 & (f) Full FT, norm-matched & $0.779$ & $1.363$ & $0.032$ & $0.028$ & $0.129$ \\
 & (d) Shuffled supervision & $0.771$ & $1.230$ & $0.157$ & $0.034$ & $0.177$ \\
\midrule
EuroSAT & Proposed ($\btau^{\rm S}_{\star}$) & $0.986$ & $1.203$ & $0.143$ & $0.072$ & $0.189$ \\
 & (e) No vector & $0.986$ & $1.437$ & $-0.066$ & $0.015$ & $0.103$ \\
 & (a) Equal-norm random & $0.985$ & $1.436$ & $-0.065$ & $0.016$ & $0.104$ \\
 & (b) Front-6 blocks & $0.983$ & $1.438$ & $-0.068$ & $0.017$ & $0.104$ \\
 & (b) Back-6 blocks & $0.986$ & $1.262$ & $0.090$ & $0.053$ & $0.169$ \\
 & (c) Full FT, raw & $0.969$ & $1.197$ & $0.148$ & $0.071$ & $0.191$ \\
 & (f) Full FT, norm-matched & $0.983$ & $1.337$ & $0.021$ & $0.036$ & $0.144$ \\
 & (d) Shuffled supervision & $0.982$ & $1.212$ & $0.138$ & $0.050$ & $0.181$ \\
\bottomrule
\end{tabular}
\space%
\end{table}

\begin{table}[t]
\centering
\caption{Continuation of Table~\ref{tab:appendix:controls:full} for the remaining five targets.}
\label{tab:appendix:controls:full2}
\scriptsize
\begin{tabular}{@{}llccccc@{}}
\toprule
Dataset & Vector & Accuracy ($\uparrow$) & E-RMSE ($\downarrow$) & Rank corr.\ ($\uparrow$) & IoU@1 ($\uparrow$) & IoU@10 ($\uparrow$) \\
\midrule
GTSRB & Proposed ($\btau^{\rm S}_{\star}$) & $0.989$ & $0.889$ & $0.307$ & $0.366$ & $0.453$ \\
 & (e) No vector & $0.989$ & $1.638$ & $-0.187$ & $0.003$ & $0.065$ \\
 & (a) Equal-norm random & $0.987$ & $1.623$ & $-0.175$ & $0.003$ & $0.070$ \\
 & (b) Front-6 blocks & $0.988$ & $1.637$ & $-0.188$ & $0.002$ & $0.064$ \\
 & (b) Back-6 blocks & $0.989$ & $0.923$ & $0.280$ & $0.313$ & $0.433$ \\
 & (c) Full FT, raw & $0.983$ & $0.888$ & $0.310$ & $0.367$ & $0.453$ \\
 & (f) Full FT, norm-matched & $0.989$ & $1.202$ & $0.116$ & $0.132$ & $0.271$ \\
 & (d) Shuffled supervision & $0.988$ & $0.937$ & $0.303$ & $0.219$ & $0.470$ \\
\midrule
MNIST & Proposed ($\btau^{\rm S}_{\star}$) & $0.900$ & $1.077$ & $0.230$ & $0.075$ & $0.323$ \\
 & (e) No vector & $0.895$ & $1.657$ & $-0.258$ & $0.005$ & $0.045$ \\
 & (a) Equal-norm random & $0.887$ & $1.656$ & $-0.255$ & $0.006$ & $0.045$ \\
 & (b) Front-6 blocks & $0.893$ & $1.659$ & $-0.261$ & $0.003$ & $0.044$ \\
 & (b) Back-6 blocks & $0.898$ & $1.322$ & $0.025$ & $0.066$ & $0.187$ \\
 & (c) Full FT, raw & $0.826$ & $1.062$ & $0.264$ & $0.074$ & $0.318$ \\
 & (f) Full FT, norm-matched & $0.886$ & $1.483$ & $-0.110$ & $0.032$ & $0.076$ \\
 & (d) Shuffled supervision & $0.903$ & $1.091$ & $0.219$ & $0.087$ & $0.318$ \\
\midrule
RESISC45 & Proposed ($\btau^{\rm S}_{\star}$) & $0.955$ & $1.185$ & $0.175$ & $0.104$ & $0.227$ \\
 & (e) No vector & $0.957$ & $1.472$ & $-0.057$ & $0.017$ & $0.102$ \\
 & (a) Equal-norm random & $0.956$ & $1.470$ & $-0.055$ & $0.017$ & $0.103$ \\
 & (b) Front-6 blocks & $0.956$ & $1.471$ & $-0.056$ & $0.017$ & $0.103$ \\
 & (b) Back-6 blocks & $0.957$ & $1.218$ & $0.150$ & $0.082$ & $0.210$ \\
 & (c) Full FT, raw & $0.934$ & $1.176$ & $0.181$ & $0.109$ & $0.230$ \\
 & (f) Full FT, norm-matched & $0.956$ & $1.308$ & $0.080$ & $0.044$ & $0.164$ \\
 & (d) Shuffled supervision & $0.951$ & $1.207$ & $0.167$ & $0.074$ & $0.207$ \\
\midrule
STL10 & Proposed ($\btau^{\rm S}_{\star}$) & $0.978$ & $1.094$ & $0.236$ & $0.133$ & $0.278$ \\
 & (e) No vector & $0.979$ & $1.559$ & $-0.149$ & $0.011$ & $0.081$ \\
 & (a) Equal-norm random & $0.976$ & $1.557$ & $-0.146$ & $0.011$ & $0.081$ \\
 & (b) Front-6 blocks & $0.980$ & $1.562$ & $-0.154$ & $0.007$ & $0.078$ \\
 & (b) Back-6 blocks & $0.980$ & $1.132$ & $0.206$ & $0.113$ & $0.262$ \\
 & (c) Full FT, raw & $0.981$ & $1.083$ & $0.246$ & $0.141$ & $0.282$ \\
 & (f) Full FT, norm-matched & $0.984$ & $1.255$ & $0.112$ & $0.062$ & $0.198$ \\
 & (d) Shuffled supervision & $0.974$ & $1.161$ & $0.199$ & $0.040$ & $0.240$ \\
\midrule
SVHN & Proposed ($\btau^{\rm S}_{\star}$) & $0.974$ & $1.018$ & $0.241$ & $0.178$ & $0.377$ \\
 & (e) No vector & $0.974$ & $1.686$ & $-0.222$ & $0.003$ & $0.055$ \\
 & (a) Equal-norm random & $0.973$ & $1.685$ & $-0.222$ & $0.003$ & $0.055$ \\
 & (b) Front-6 blocks & $0.974$ & $1.687$ & $-0.226$ & $0.003$ & $0.053$ \\
 & (b) Back-6 blocks & $0.974$ & $1.158$ & $0.149$ & $0.148$ & $0.289$ \\
 & (c) Full FT, raw & $0.968$ & $0.994$ & $0.269$ & $0.174$ & $0.386$ \\
 & (f) Full FT, norm-matched & $0.974$ & $1.367$ & $0.013$ & $0.048$ & $0.165$ \\
 & (d) Shuffled supervision & $0.973$ & $1.047$ & $0.245$ & $0.124$ & $0.382$ \\
\bottomrule
\end{tabular}
\space%
\end{table}
\space%
\section{Comparison with Amortized and Transferable Explainers}\label{sec:appendix:alternatives}

Table~\ref{tab:appendix:alternatives:deployment} compares the deployment options discussed in Section~\ref{sec:experiment:imagenet:faithfulness}; this appendix details the two alternatives, an amortized explainer trained with the same supervision as our source model and Transferable Vision Explainer (TVE)~\cite{wang2024tve}.

\begin{table}[t]
\caption{Deployment options for explaining a finetuned target classifier (CIFAR10 set, 10-target means): the classifier whose predictions are explained; the accuracy of the classifier the user deploys (TVE: canonical / best of four probe pipelines); forward passes per image for prediction plus explanation; inference parameters beyond the deployed classifier; and wins over a random attribution on the classifier each method explains (10 targets $\times$ 4 faithfulness metrics; not a ranking across methods, since the explained classifiers differ).}
\label{tab:appendix:alternatives:deployment}
\centering
\small
\setlength{\tabcolsep}{4pt}
\begin{tabular}{@{}llcccc@{}}
\toprule
Method & Explains & Acc. & Fwd. & +Params & Non-random \\
\midrule
Proposed ($\tilde{\btheta}^{\rm T}_{\rm ft\star}$) & finetuned target itself & $0.911$ & $1$ & $0$ & $40/40$ \\
$\btheta^{\rm T}_{\rm ft}$ + Kernel SHAP ($P = 500$) & $\btheta^{\rm T}_{\rm ft}$ & $0.914$ & $\approx 500$ & $0$ & $40/40$ \\
$\btheta^{\rm T}_{\rm ft}$ + amortized ($\alpha = 0$) & $\btheta^{\rm T}_{\rm ft}$ & $0.914$ & $2$ & $87.8$M & $38/40$ \\
TVE~\cite{wang2024tve} (public checkpoint) & separate linear probe & $0.779$ / $0.817$ & $3$ & $244.9$M & $37/40$ \\
\bottomrule
\end{tabular}
\end{table}
We do not compare absolute faithfulness values across methods, because the four methods explain different classifiers on different backbones and patch grids.
The comparison across methods uses only the deployment characteristics and the wins over a random attribution.

\subsection{Amortized Explainer with the Same Supervision}\label{sec:appendix:alternatives:amortized}
The $\alpha = 0$ model of the objective~\bref{eq:proposed:objective} is trained on ImageNet+X with the explanation loss only and is therefore an amortized explainer with the same supervision and the same backbone as $\btheta^{\rm S}_{\rm ft\star}$.
We deploy it to each target by replacing the head $g_{\bW}$ with the target's class-text embeddings, exactly as for zero-shot classification, with no target-side training of any kind.
Its own classification accuracy is between $0.003$ and $0.110$ across the targets, so in Table~\ref{tab:appendix:alternatives:deployment} it is paired with $\btheta^{\rm T}_{\rm ft}$, which provides the prediction while the amortized model provides the attribution (two forward passes).
Our evaluation uses its attribution for the true class.
The visual tower and head of the second model add $87.8$M parameters.

\paragraph{GT agreement}
Table~\ref{tab:appendix:alternatives:amortized} evaluates the amortized explainer with the metrics and the 10 targets of Table~\ref{tab:experiment:imagenet} (including SUN397), next to the zero-shot model and the proposed method.
The expected ordering (zero-shot $<$ amortized $<$ proposed) holds on all four 10-target means and in $37$ of the $40$ target--metric cells.
The three exceptions are underlined in the table, where the amortized explainer exceeds the proposed method (Rank corr.\ and IoU@10 on DTD, IoU@1 on STL10).
On IoU@5, which the table omits, the ordering holds on all 10 targets.

\paragraph{Faithfulness}
We score the amortized attribution on the image subsets of Appendix~\ref{sec:appendix:faithfulness}, which keep the images that the transferred model classifies correctly.
For each image we use the attribution for the true class $y$ and compute the four metrics on the probability that $\btheta^{\rm T}_{\rm ft}$ assigns to $y$, as for the other attributions in Appendix~\ref{sec:appendix:faithfulness}.
It beats the random attribution on $38$ of the $40$ target--metric comparisons.
The two losses are Fid$+$ and Del on EuroSAT, where Kernel SHAP on $\btheta^{\rm T}_{\rm ft}$ beats random on all four metrics.

\begin{table}[t]
\centering
\caption{GT agreement of the $\alpha = 0$ amortized explainer (Amort.) on the 10 targets of Table~\ref{tab:experiment:imagenet}, next to the zero-shot model (ZS) and the proposed method (Prop.). Underlined cells violate the ordering ZS $<$ Amort.\ $<$ Prop.}
\label{tab:appendix:alternatives:amortized}
\scriptsize
\setlength{\tabcolsep}{3pt}
\begin{tabular}{@{}lcccccccccccc@{}}
\toprule
 & \multicolumn{3}{c}{E-RMSE ($\downarrow$)} & \multicolumn{3}{c}{Rank corr.\ ($\uparrow$)} & \multicolumn{3}{c}{IoU@1 ($\uparrow$)} & \multicolumn{3}{c}{IoU@10 ($\uparrow$)} \\
\cmidrule(lr){2-4}\cmidrule(lr){5-7}\cmidrule(lr){8-10}\cmidrule(lr){11-13}
Dataset & ZS & Amort. & Prop. & ZS & Amort. & Prop. & ZS & Amort. & Prop. & ZS & Amort. & Prop. \\
\midrule
Cars & $1.623$ & $1.118$ & $1.068$ & $-0.204$ & $0.225$ & $0.249$ & $0.002$ & $0.107$ & $0.169$ & $0.048$ & $0.305$ & $0.350$ \\
CIFAR100 & $1.508$ & $1.233$ & $1.223$ & $-0.087$ & $0.134$ & $0.143$ & $0.015$ & $0.085$ & $0.099$ & $0.095$ & $0.212$ & $0.222$ \\
DTD & $1.462$ & $1.246$ & $1.242$ & $-0.061$ & $\underline{0.145}$ & $0.141$ & $0.021$ & $0.037$ & $0.048$ & $0.102$ & $\underline{0.177}$ & $0.172$ \\
EuroSAT & $1.377$ & $1.243$ & $1.203$ & $-0.009$ & $0.109$ & $0.143$ & $0.018$ & $0.032$ & $0.072$ & $0.115$ & $0.170$ & $0.189$ \\
GTSRB & $1.543$ & $1.011$ & $0.889$ & $-0.142$ & $0.261$ & $0.307$ & $0.006$ & $0.241$ & $0.366$ & $0.073$ & $0.371$ & $0.453$ \\
MNIST & $1.536$ & $1.136$ & $1.077$ & $-0.117$ & $0.193$ & $0.230$ & $0.009$ & $0.064$ & $0.075$ & $0.081$ & $0.274$ & $0.323$ \\
RESISC45 & $1.427$ & $1.209$ & $1.185$ & $-0.016$ & $0.163$ & $0.175$ & $0.016$ & $0.080$ & $0.104$ & $0.116$ & $0.213$ & $0.227$ \\
STL10 & $1.548$ & $1.132$ & $1.094$ & $-0.133$ & $0.208$ & $0.236$ & $0.010$ & $\underline{0.141}$ & $0.133$ & $0.087$ & $0.256$ & $0.278$ \\
SUN397 & $1.519$ & $1.259$ & $1.248$ & $-0.108$ & $0.118$ & $0.130$ & $0.010$ & $0.079$ & $0.088$ & $0.081$ & $0.195$ & $0.205$ \\
SVHN & $1.534$ & $1.111$ & $1.018$ & $-0.074$ & $0.217$ & $0.241$ & $0.013$ & $0.106$ & $0.178$ & $0.104$ & $0.298$ & $0.377$ \\
\midrule
Mean & $1.508$ & $1.170$ & $1.125$ & $-0.095$ & $0.177$ & $0.200$ & $0.012$ & $0.097$ & $0.133$ & $0.090$ & $0.247$ & $0.279$ \\
\bottomrule
\end{tabular}
\space%
\end{table}

\subsection{Transferable Vision Explainer}\label{sec:appendix:alternatives:tve}
TVE pretrains an explainer on ImageNet with a self-supervised objective and produces attributions for classifiers built on the backbone encoder used in its pretraining.
Its training objective is self-supervised and does not use explanation supervision such as our source explanations.
We use the official implementation\footnote{\url{https://github.com/guanchuwang/TVE}} and checkpoint\footnote{\url{https://huggingface.co/Anonymous-researcher/leta}} (two ViT-Base backbones, $244.9$M parameters in total) and its canonical configuration, in which a per-target linear probe is trained on the frozen encoder.
In our evaluation TVE therefore explains this probe, not our finetuned CLIP classifier. Applying TVE to our classifier would require training an explainer for the CLIP backbone, which we do not do.

\paragraph{Probe protocol}
Probes are logistic-regression classifiers.
The L-BFGS probes select the inverse regularization strength $C \in \{10^{-3}, \ldots, 10^{3}\}$ on a $10\%$ validation split and are refit on the full training set.
The AdamW probes are trained for $50$ epochs with learning rate $10^{-3}$, a cosine schedule, and no weight decay.
To rule out an under-tuned probe, Table~\ref{tab:appendix:alternatives:probe} reports four pre-specified pipelines, which combine two optimizers (AdamW and L-BFGS) with two input normalizations.
One normalization uses the ViT-MAE statistics of TVE's explainer (canonical, since the explainer and the encoder share one input tensor), and the other uses the encoder's native statistics (TVE's best case).
The optimizer changes mean accuracy by at most $0.4$ points, and the normalization changes it by $3.8$ points.
Even in the best case ($0.817$) the probe is $9.7$ points below $\btheta^{\rm T}_{\rm ft}$ ($0.914$ on the same targets), and $13.5$ points below it in the canonical configuration ($0.779$).
This gap reflects our evaluation setting, in which the probe uses TVE's public frozen encoder instead of the finetuned CLIP backbone.

\paragraph{Faithfulness}
This check verifies that our TVE setup produces working explanations, so that the comparison in Table~\ref{tab:appendix:alternatives:deployment} is not driven by a broken baseline.
Table~\ref{tab:appendix:alternatives:tve} scores TVE's attributions (canonical configuration, with the attribution taken as the difference between the classifier logits of TVE's positive and negative outputs, as in the public TVE code) against a random attribution on the probe's own predicted class, with the four metrics of Appendix~\ref{sec:appendix:faithfulness}, on $500$ test images per target and TVE's $14 \times 14$ patch grid.
TVE beats random on $37$ of the $40$ comparisons (the three losses are Fid$-$ on MNIST and SVHN and insertion on SVHN).
TVE thus explains its probe better than a random attribution.
The table does not include our method, because faithfulness values depend on the classifier whose probabilities are scored, and the two methods explain different classifiers on different patch grids.
Our method is compared with TVE in Table~\ref{tab:appendix:alternatives:deployment}, where the difference lies in which classifier is explained and at what accuracy and cost.

\begin{table}[t]
\centering
\caption{Test accuracy of TVE's per-target linear probe under four pipelines (optimizer $\times$ input normalization). ``MAE + L-BFGS'' is the canonical configuration used in Tables~\ref{tab:appendix:alternatives:deployment} and~\ref{tab:appendix:alternatives:tve}; ``native'' is TVE's best case.}
\label{tab:appendix:alternatives:probe}
\footnotesize
\begin{tabular}{@{}lcccc@{}}
\toprule
Dataset & MAE+AdamW & MAE+L-BFGS & native+AdamW & native+L-BFGS \\
\midrule
Cars & $0.375$ & $0.379$ & $0.499$ & $0.510$ \\
CIFAR10 & $0.934$ & $0.938$ & $0.965$ & $0.970$ \\
CIFAR100 & $0.770$ & $0.791$ & $0.851$ & $0.863$ \\
DTD & $0.693$ & $0.686$ & $0.731$ & $0.740$ \\
EuroSAT & $0.953$ & $0.954$ & $0.958$ & $0.955$ \\
GTSRB & $0.643$ & $0.643$ & $0.722$ & $0.724$ \\
MNIST & $0.967$ & $0.979$ & $0.976$ & $0.978$ \\
RESISC45 & $0.866$ & $0.867$ & $0.867$ & $0.869$ \\
STL10 & $0.986$ & $0.987$ & $0.994$ & $0.994$ \\
SVHN & $0.562$ & $0.565$ & $0.565$ & $0.564$ \\
\midrule
Mean & $\mathbf{0.775}$ & $\mathbf{0.779}$ & $\mathbf{0.813}$ & $\mathbf{0.817}$ \\
\bottomrule
\end{tabular}
\space%
\end{table}

\begin{table}[t]
\centering
\caption{Faithfulness of TVE (canonical configuration) versus a random attribution, scored on the probe's own predicted class ($N = 500$ test images per dataset; first $500$ images of the test split, except $\dagger$EuroSAT, which uses a class-stratified subset because its first $500$ test images cover only two of 10 classes). TVE's value is in bold where it beats random.}
\label{tab:appendix:alternatives:tve}
\small
\begin{tabular}{@{}lccccc@{}}
\toprule
Dataset & Fid$+$ ($\uparrow$) & Fid$-$ ($\downarrow$) & Ins ($\uparrow$) & Del ($\downarrow$) & Probe acc. \\
 & TVE / random & TVE / random & TVE / random & TVE / random & \\
\midrule
Cars & $\mathbf{0.228}$ / $0.162$ & $\mathbf{0.111}$ / $0.163$ & $\mathbf{0.192}$ / $0.143$ & $\mathbf{0.114}$ / $0.151$ & $0.379$ \\
CIFAR10 & $\mathbf{0.272}$ / $0.154$ & $\mathbf{0.098}$ / $0.166$ & $\mathbf{0.652}$ / $0.564$ & $\mathbf{0.506}$ / $0.591$ & $0.938$ \\
CIFAR100 & $\mathbf{0.328}$ / $0.226$ & $\mathbf{0.161}$ / $0.221$ & $\mathbf{0.443}$ / $0.366$ & $\mathbf{0.318}$ / $0.381$ & $0.791$ \\
DTD & $\mathbf{0.129}$ / $0.113$ & $\mathbf{0.098}$ / $0.109$ & $\mathbf{0.610}$ / $0.563$ & $\mathbf{0.569}$ / $0.580$ & $0.686$ \\
EuroSAT$^{\dagger}$ & $\mathbf{0.148}$ / $0.111$ & $\mathbf{0.085}$ / $0.110$ & $\mathbf{0.651}$ / $0.613$ & $\mathbf{0.613}$ / $0.627$ & $0.954$ \\
GTSRB & $\mathbf{0.422}$ / $0.200$ & $\mathbf{0.142}$ / $0.202$ & $\mathbf{0.493}$ / $0.407$ & $\mathbf{0.262}$ / $0.421$ & $0.643$ \\
MNIST & $\mathbf{0.613}$ / $0.402$ & $0.460$ / $0.394$ & $\mathbf{0.428}$ / $0.421$ & $\mathbf{0.273}$ / $0.429$ & $0.979$ \\
RESISC45 & $\mathbf{0.227}$ / $0.139$ & $\mathbf{0.107}$ / $0.131$ & $\mathbf{0.587}$ / $0.536$ & $\mathbf{0.470}$ / $0.554$ & $0.867$ \\
STL10 & $\mathbf{0.072}$ / $0.032$ & $\mathbf{0.032}$ / $0.035$ & $\mathbf{0.784}$ / $0.736$ & $\mathbf{0.718}$ / $0.761$ & $0.987$ \\
SVHN & $\mathbf{0.201}$ / $0.116$ & $0.137$ / $0.113$ & $0.311$ / $0.318$ & $\mathbf{0.268}$ / $0.323$ & $0.565$ \\
\bottomrule
\end{tabular}
\space%
\end{table}
\space%
\section{Selecting \texorpdfstring{$\lambda_2$}{lambda2} without Target-Side Explanation Supervision}\label{sec:appendix:lambda2}

The experiments of Section~\ref{sec:experiment} use $\alpha = 0.8$ and $\lambda_2 = 0.7$ for all targets.
These values are a representative operating point on the accuracy--explanation trade-off rather than optimal values, because the right balance depends on the user's task.
$\alpha$ is a source-side training hyperparameter and can be selected on source validation data alone.
We fix $\alpha = 0.8$ as a representative source-side point rather than by such a rule, and Figure~\ref{fig:experiment:imagenet:scaling_coef} shows the sensitivity to $\alpha \in \{0.1, \ldots, 0.9\}$.
This appendix documents the sweep behind $\lambda_2$ and two held-out selection protocols that never use the explanation supervision of the target being selected for.

\paragraph{Sweep}
Table~\ref{tab:appendix:lambda2:sweep} reports the 10-target means (CIFAR10 set, $\lambda_1 = 1.0$, $\alpha = 0.8$) over the grid $\lambda_2 \in \{0.4, 0.5, \ldots, 1.2\}$.
Explanation metrics improve up to a plateau at $\lambda_2 \approx 1.0$--$1.1$ (Rank corr.\ spread $0.002$ over $[0.9, 1.1]$), while accuracy decreases monotonically over the whole grid ($2.0$ points from $0.4$ to $1.2$).
Accuracy falls by $0.3$ points from $0.4$ to $0.7$ and by $1.7$ points from $0.7$ to $1.2$.
The grid does not include $\lambda_2 = 0$, so neither protocol below can select ``no explainability vector''.

\begin{table}[t]
\centering
\caption{Sensitivity of the proposed method to $\lambda_2$ (10-target means on the CIFAR10 set, full test split, $\lambda_1 = 1.0$, $\alpha = 0.8$).}
\label{tab:appendix:lambda2:sweep}
\small
\begin{tabular}{@{}lcccc@{}}
\toprule
$\lambda_2$ & Accuracy ($\uparrow$) & Rank corr.\ ($\uparrow$) & IoU@10 ($\uparrow$) & E-RMSE ($\downarrow$) \\
\midrule
0.4 & $0.914$ & $0.120$ & $0.221$ & $1.233$ \\
0.5 & $0.914$ & $0.161$ & $0.255$ & $1.175$ \\
0.6 & $0.913$ & $0.186$ & $0.272$ & $1.138$ \\
0.7 (default) & $0.911$ & $0.204$ & $0.281$ & $1.118$ \\
0.8 & $0.909$ & $0.213$ & $0.285$ & $1.109$ \\
0.9 & $0.907$ & $0.217$ & $0.286$ & $1.104$ \\
1.0 & $0.903$ & $0.219$ & $0.287$ & $1.102$ \\
1.1 & $0.899$ & $0.219$ & $0.286$ & $1.102$ \\
1.2 & $0.894$ & $0.218$ & $0.285$ & $1.103$ \\
\bottomrule
\end{tabular}
\space%
\end{table}

\paragraph{Leave-one-dataset-out selection on explanation metrics}
For each held-out target, we choose the $\lambda_2$ that maximizes the mean explanation metric over the other nine targets (the metric is fixed in advance, and the held-out target's labels are never used) and evaluate the held-out target at that value.
The Explanation-only columns of Table~\ref{tab:appendix:lambda2:lodo} show that this protocol picks $1.0$--$1.1$ for every held-out target under Rank corr.\ and for nine of the 10 under IoU@10 (GTSRB picks $0.9$), and picks $0.7 \pm 0.1$ for none of them.
If $\lambda_2$ had been chosen by maximizing explanation metrics on the benchmark targets, the result would have been $0.9$--$1.1$, not $0.7$.
Compared with $\lambda_2 = 0.7$, the selected value raises held-out Rank corr.\ by $0.015$ on average and by at most $0.033$ (MNIST), and the value selected under IoU@10 changes held-out IoU@10 by at most $0.011$ (SVHN), so the conclusions of Table~\ref{tab:experiment:imagenet} do not depend on landing on a narrow optimum.
This argument rules out selection by argmax on target explanation metrics.
It does not rule out that the choice of $0.7$ was informed by the aggregate sweep, and we report it as a representative operating point.

\paragraph{Accuracy-constrained selection}
This protocol fixes in advance a tolerance $\varepsilon$, in accuracy points, on the accuracy loss that the deployment can accept.
Among the grid values of $\lambda_2$ whose accuracy is within $\varepsilon$ of the best accuracy on the grid, it selects the one with the best explanation metric on other datasets.
The rule thus needs no explanation labels on the target being selected for.
We run this rule leave-one-dataset-out on the benchmark records, with accuracy and Rank corr.\ both averaged over the other nine targets, so the held-out target's own labels are not used.
In deployment, the accuracy term would instead come from the target's validation classification labels, which we do not evaluate separately.
The Accuracy-constrained columns of Table~\ref{tab:appendix:lambda2:lodo} show that with $\varepsilon = 0.5$ points it selects exactly $0.7$ for eight targets and $0.7 \pm 0.1$ for nine of the 10, while $\varepsilon = 1.0$ points moves the selection to $0.9$ for most targets.
This procedure and the value $\varepsilon = 0.5$ points were both chosen after seeing the explanation-only result above, for illustration.
We include it as a practical selection rule that recovers the adopted operating point, not as evidence that $0.7$ was derived this way.

\begin{table}[t]
\centering
\caption{Leave-one-dataset-out selection of $\lambda_2$ on the CIFAR10 set. The Explanation-only columns give the $\lambda_2$ maximizing the mean Rank corr.\ or IoU@10 of the nine other targets. The Accuracy-constrained columns give the $\lambda_2$ maximizing the mean Rank corr.\ of the nine other targets among the values whose mean accuracy is within $\varepsilon$ of the best on the grid. The last row gives the mean held-out Rank corr.\ gain of the selected value over $\lambda_2 = 0.7$.}
\label{tab:appendix:lambda2:lodo}
\footnotesize
\begin{tabular}{@{}lcccc@{}}
\toprule
 & \multicolumn{2}{c}{Explanation-only} & \multicolumn{2}{c}{Accuracy-constrained} \\
\cmidrule(lr){2-3}\cmidrule(lr){4-5}
Held-out target & Rank corr. & IoU@10 & $\varepsilon = 0.5$pt & $\varepsilon = 1.0$pt \\
\midrule
Cars & 1.1 & 1.0 & 0.9 & 1.1 \\
CIFAR10 & 1.0 & 1.0 & 0.7 & 0.9 \\
CIFAR100 & 1.1 & 1.0 & 0.7 & 0.9 \\
DTD & 1.0 & 1.0 & 0.8 & 1.0 \\
EuroSAT & 1.1 & 1.0 & 0.7 & 0.9 \\
GTSRB & 1.1 & 0.9 & 0.7 & 0.9 \\
MNIST & 1.1 & 1.1 & 0.7 & 0.9 \\
RESISC45 & 1.0 & 1.0 & 0.7 & 0.9 \\
STL10 & 1.0 & 1.0 & 0.7 & 0.9 \\
SVHN & 1.1 & 1.0 & 0.7 & 0.9 \\
\midrule
Within $0.7 \pm 0.1$ & 0/10 & 0/10 & 9/10 & 0/10 \\
Mean (chosen $-$ 0.7) on Rank corr. & $+0.015$ & --- & $+0.001$ & $+0.014$ \\
\bottomrule
\end{tabular}
\space%
\end{table}
\space%
\section{Task-Dependent Explanations on Two-Task Images}\label{sec:appendix:stress}

Visual explanations depend on the task.
The same image calls for attention to different regions depending on the task, for example when it is classified by object category or by scene.
A natural concern is therefore whether a single explainability vector learned on one source can yield task-appropriate explanations on a target whose task differs.
This appendix tests this directly by constructing images that carry two classification tasks at once.

\paragraph{Setup}
We composite an MNIST digit~\cite{lecun1998mnist} and a FashionMNIST clothing item~\cite{xiao2017fashionmnist} at non-overlapping random positions of a $224 \times 224$ canvas ($12{,}000$ training and $2{,}000$ test images), so that every image has both a digit label and a clothing label, together with the ground-truth set of patches covering each object ($7.5$ patches on average for the digit, $10.9$ for the clothing item, out of $M = 49$).
Two prediction-only models are finetuned with classification supervision only, one for the digit task and one for the clothing task (test accuracy $0.9925$ and $0.9195$, respectively, using the MNIST prompt template for digits and the generic ``a photo of a \{class\}'' template for clothing).
The same explainability vector $\btau^{\rm S}_{\star}$ learned on ImageNet+X is then added to both models with $\lambda_2 = 0.7$, as in Section~\ref{sec:experiment:imagenet}.
On the shared $2{,}000$ test images, we measure which object region the top-$k$ patches of each model's self-explanation fall into.
The self-explanation is the attribution for the predicted class, and $k = \min(5, |\text{region}|)$ handles small digit regions.
The chance rate is the region's share of the $49$ patches, which is $15.2\%$ for the digit region and $22.2\%$ for the clothing region.

\begin{figure}[t]
\centering
\includegraphics[width=0.78\textwidth]{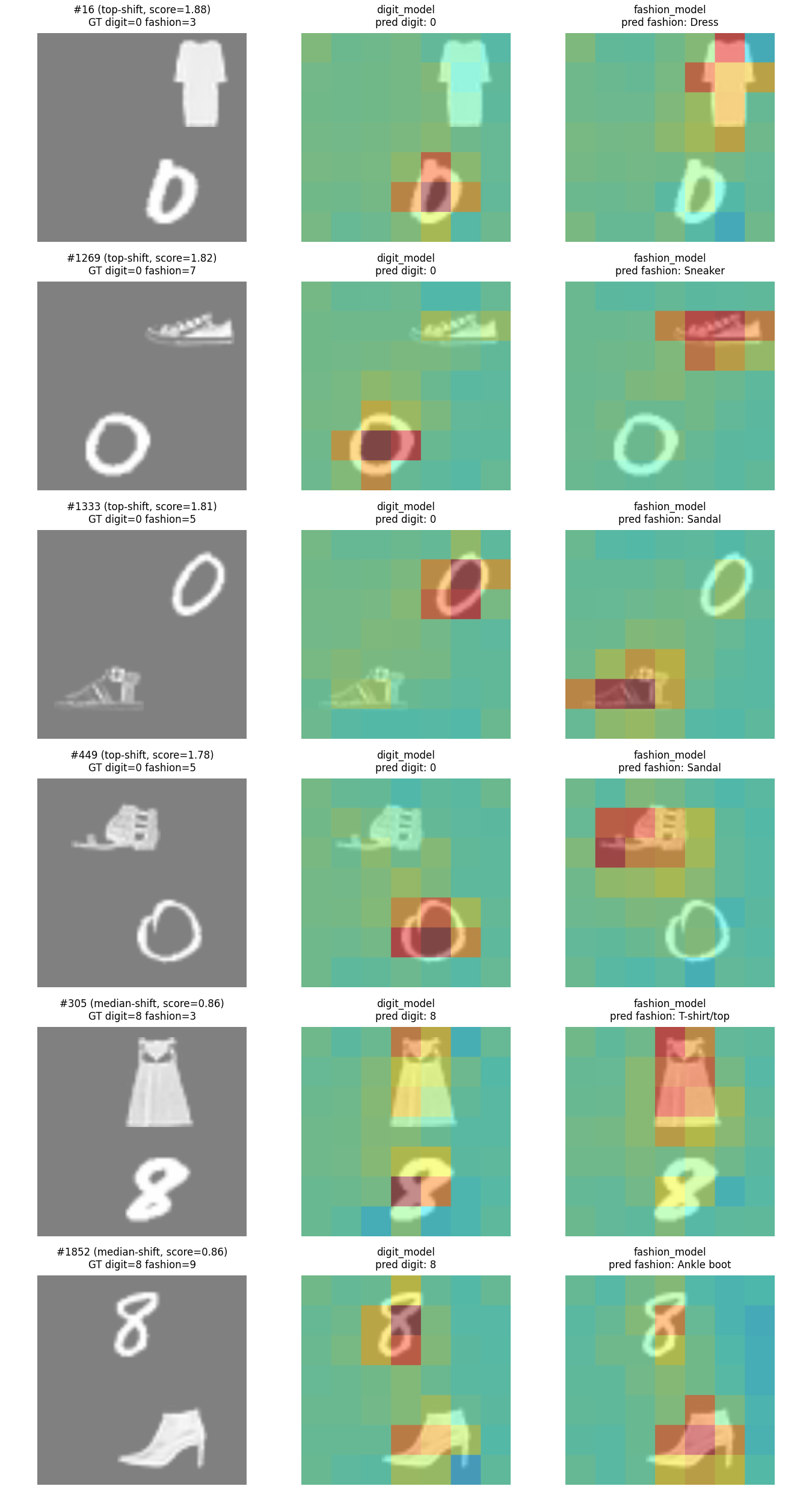}
\caption{Self-explanations of the digit model (middle) and the clothing model (right) on the same composited test images (left), both obtained by adding the same $\btau^{\rm S}_{\star}$ to a classification-only model. Heatmaps show the attribution for the predicted class ($7 \times 7$ patches, shared color scale per row). The first four rows are the test images with the largest attribution shift between the two models, and the last two are images at the median shift. All six are classified correctly by both models.}
\label{fig:appendix:stress}
\end{figure}

\paragraph{Results}
Table~\ref{tab:appendix:stress} shows that the same vector steers the attributions of each model to the region of its own task.
The digit model places $56.8\%$ of its top-$k$ patches on the digit ($3.7\times$ chance), and the clothing model places $91.6\%$ on the clothing item ($4.1\times$ chance).
Adding the vector lowers accuracy by at most $0.3$ points.
The per-image paired difference between the two models in the digit-region rate has a bootstrap $95\%$ interval of $[+0.475, +0.496]$, and that in the clothing-region rate has $[-0.501, -0.481]$ ($10^4$ resamples).
A fixed top-$5$ variant and a continuous attribution-mass variant give the same conclusion.
The effect is attributable to the vector.
Without $\btau^{\rm S}_{\star}$, the digit model places only $1.0\%$ of its top-$5$ patches on the digit, which is below chance.
The clothing model then places only $4.6\%$ on the clothing item and $65.8\%$ on the digit instead.
Classification-only finetuning with a frozen head does not directly supervise patch-level attributions.
Figure~\ref{fig:appendix:stress} shows examples in which the heatmap moves between the digit and the clothing item within the same pixels as the task changes.
The separation is not complete, since $42.5\%$ of the digit model's top patches fall on the larger clothing region.
Composited images are also simpler than natural scenes, and the validity on natural images rests on the 10-dataset experiments with the same vector.

\paragraph{Interpretation}
This experiment supports the following reading of the transfer rule~\bref{eq:proposed:theta_init}, which Appendix~\ref{sec:appendix:controls} supports independently.
The target-side terms, namely the target task vector $\btau^{\rm T}_{\rm ft}$ and the class-text embeddings in $g_{\bW}$, determine which evidence the model attends to.
The vector $\btau^{\rm S}_{\star}$ adds the ability to read that task-specific evidence out as patch-level attributions.
The result is consistent with this reading, since the same vector serves both tasks without any information about them.

\begin{table}[t]
\centering
\caption{Explanations on two-task images ($2{,}000$ shared test images). Region rates are the fraction of the top-$k$ self-explanation patches ($k = \min(5, |\text{region}|)$) that fall in each object region. Chance is the region's share of the $49$ patches. The last row uses a fixed top-$5$ (near-identical to the adaptive $k$ in this setting).}
\label{tab:appendix:stress}
\small
\begin{tabular}{@{}lcc@{}}
\toprule
 & Digit model & Clothing model \\
\midrule
Accuracy, with $\btau^{\rm S}_{\star}$ / without & $0.9915$ / $0.9925$ & $0.9165$ / $0.9195$ \\
Top-$k$ patches in the digit region & $\mathbf{56.8\%}$ ($3.7\times$ chance) & $8.3\%$ \\
Top-$k$ patches in the clothing region & $42.5\%$ & $\mathbf{91.6\%}$ ($4.1\times$ chance) \\
Chance rate (digit / clothing region) & \multicolumn{2}{c}{$15.2\%$ / $22.2\%$} \\
\midrule
Own-region rate without $\btau^{\rm S}_{\star}$ (top-$5$) & $1.0\%$ & $4.6\%$ \\
\bottomrule
\end{tabular}
\end{table}

\space%

\section{Limitations}\label{sec:limitations}

\paragraph{Architecture}
The current method is specifically designed for Vision Transformer (ViT) models pretrained with vision-language objectives such as CLIP.
It relies on the patch token structure of ViTs and the shared linear head $g_{\bW}$ constructed from text embeddings.
It also assumes that the source and target models are finetuned from the same pretrained checkpoint with this head kept frozen.
Prediction-only classifiers whose head is replaced or trained for the target task, as is common in transfer learning, are not directly covered, because their parameters do not lie in a shared space across label sets.
Extending the approach to other architectures, such as CNNs or ViTs trained with different strategies (e.g., supervised pretraining~\citep{dosovitskiy2021an}, MAE~\citep{he2022masked}, DINO~\citep{caron2021emerging}), would require architecture-specific design considerations and is an important direction for future work.

\paragraph{Explanation type}
This work focuses on patch-level attributions as the form of visual explanation.
Other types of explanations, such as concept-based explanations~\citep{kim2018interpretability,koh2020concept} or prototype-based explanations~\citep{chen2019looks}, are not covered by the current framework.
Generalizing the idea of transferring explainability via task arithmetic to these alternative explanation forms is a promising future direction.

\paragraph{Evaluation scope}
Our primary metrics measure agreement with a user-specified explanation target.
Faithfulness is evaluated within the transferred model (Section~\ref{sec:experiment:imagenet:faithfulness} and Appendix~\ref{sec:appendix:faithfulness}), and we do not rank explanation quality across methods whose explained classifiers differ.
The control experiments (Appendix~\ref{sec:appendix:controls}) show that part of the transferred effect is a content-free attribution prior.
A vector trained on shuffled explanation supervision recovers most of the coarse ranking metrics but only 57.4\% of the IoU@1 improvement, and it is below the vector trained on real supervision on IoU@1 for 9 of the 10 targets.
The two-task experiment (Appendix~\ref{sec:appendix:stress}) uses composited images with exact region masks, and its validity on natural images rests on the 10-dataset experiments.

\paragraph{When the target must not be modified}
Our method adds a vector to the parameters of the target classifier, and its accuracy may change slightly.
When the deployed model must remain unmodified, e.g., for certification or compliance reasons, external explainers are the appropriate choice.
Transferable explainers such as TVE~\cite{wang2024tve} require an explainer trained for the target's backbone, and in our evaluation with its public checkpoint TVE explains a separate linear probe (Appendix~\ref{sec:appendix:alternatives}). Post-hoc methods such as Kernel SHAP come at the cost of hundreds of forward passes per image (Section~\ref{sec:experiment:imagenet:faithfulness}).

\paragraph{Scope of the predictability finding}
The predictability examined in Section~\ref{sec:experiment:predict} holds for a fixed source with held-out targets, and on most explanation-quality metrics but not all.
On ViT-B/32, IoU@1 prediction is weak (Ridge $r = +0.33$, the better of the two regressors, Appendix~\ref{sec:appendix:predict}), so the six features do not capture which single patch is most important.
On ViT-L/14, the raw $\Delta_q$ target is not predictable due to per-target variance.
Under the oracle-normalized target $R_q$ (Appendix~\ref{sec:appendix:predict}), three of five metrics become significant at uncorrected $p < 0.05$ (IoU@1, E-MSE, and Rank corr.), while IoU@5 and IoU@10 remain unpredictable.
$R_q$ requires per-target oracle scores only at predictor-fitting time.
Predicting the outcome on a new target itself uses only the six model-internal features and no target-side explanation supervision.
Closing the remaining metric gap, e.g., by enlarging $N$ beyond 10 source--target pairs or by introducing scale-aware features, remains future work.
The held-out 56-pair evaluation in Appendix~\ref{sec:appendix:pair:predictor} also relies on source fixed effects to absorb per-source intercepts.
Under leave-one-source-out evaluation, where such intercepts are unavailable, none of 40 predictor configurations (including ones given oracle features of the target) meets our criteria.
The feasibility result is therefore confined to source datasets observed at predictor-fitting time, and prediction for genuinely unseen sources remains open.
\space%

\section{Impact Statement}\label{sec:appendix:impact_statement}
This paper presents a method for transferring visual explainability from self-explaining to prediction-only VLM-based image classifiers.
Our approach requires explanation supervision only once on the source domain and reuses it across target domains, so users need no target-side explanation supervision or training, and inference stays a single forward pass.
The one-time source-side cost (about $100$ GPU-hours in our setup, Appendix~\ref{sec:appendix:complexity}) is borne by the provider of the explainability vector.
This can lower a barrier to adopting explainable AI in resource-constrained scenarios.
As a result, this work may help users inspect the decisions of AI systems and improve their transparency.

However, there are potential risks associated with our method. Since the transferred explanation capability is not necessarily optimized for the target domain of the prediction-only model, it may produce lower-quality or misleading explanations.
Users relying on such explanations could make incorrect decisions based on misinterpreted model behavior.
Plausible-looking explanations can also raise trust beyond what is warranted, especially since part of the transferred effect is a content-free attribution prior (Appendix~\ref{sec:appendix:controls}).
We encourage practitioners to carefully evaluate the quality of transferred explanations using faithfulness checks such as those of Appendix~\ref{sec:appendix:faithfulness}, or a small set of explanation labels when available, and, when necessary, to finetune the model with a small amount of labeled explanation data to mitigate these risks.
\space%

\section{Additional Visualization Examples}\label{sec:appendix:additional_examples}

We provide additional randomly selected visualization examples for each target dataset to complement the examples shown in Figure~\ref{fig:experiment:imagenet:example}.
The examples use CLIP ViT-B/32 and the same models as Table~\ref{tab:experiment:imagenet}.
Each figure shows 10 randomly selected samples from the corresponding dataset, highlighting the top-10 patches of each attribution, as in Figure~\ref{fig:experiment:imagenet:example}.

\begin{figure}[p]
\centering
\includegraphics[width=\textwidth]{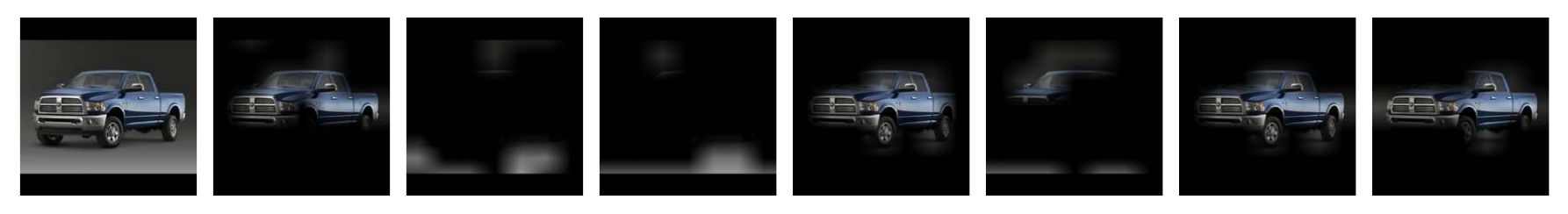}\\
\vspace{-0.2cm}
\includegraphics[width=\textwidth]{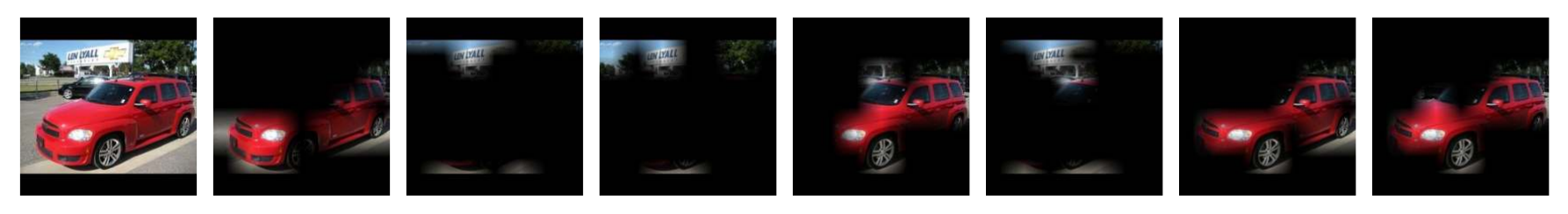}\\
\vspace{-0.2cm}
\includegraphics[width=\textwidth]{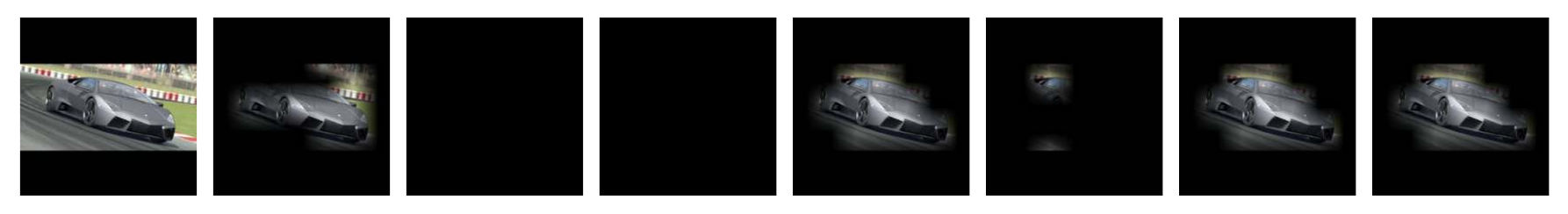}\\
\vspace{-0.2cm}
\includegraphics[width=\textwidth]{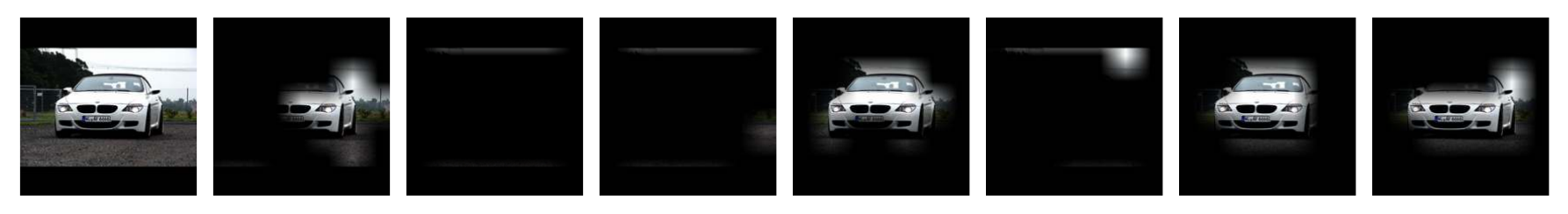}\\
\vspace{-0.2cm}
\includegraphics[width=\textwidth]{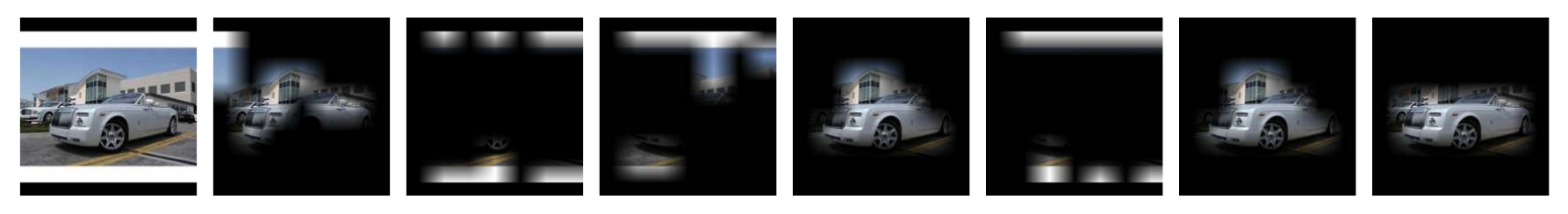}\\
\vspace{-0.2cm}
\includegraphics[width=\textwidth]{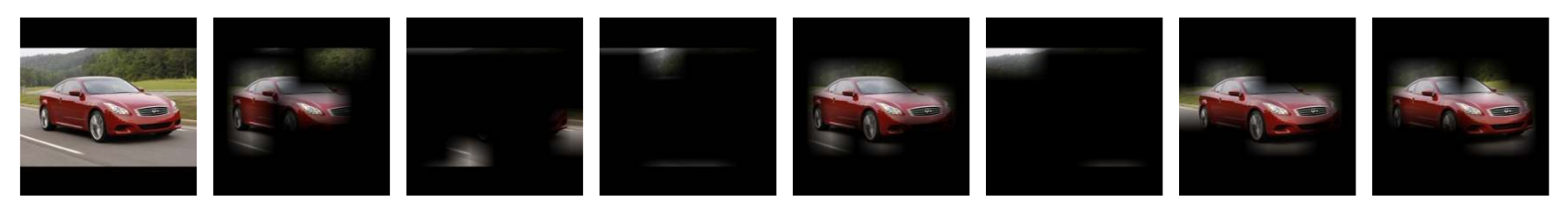}\\
\vspace{-0.2cm}
\includegraphics[width=\textwidth]{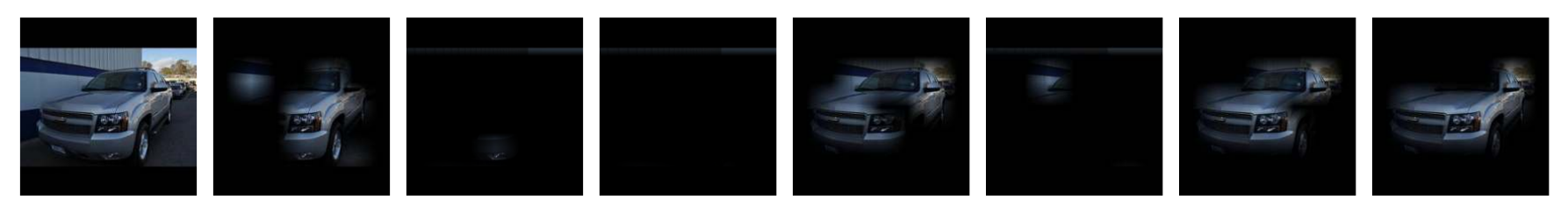}\\
\vspace{-0.2cm}
\includegraphics[width=\textwidth]{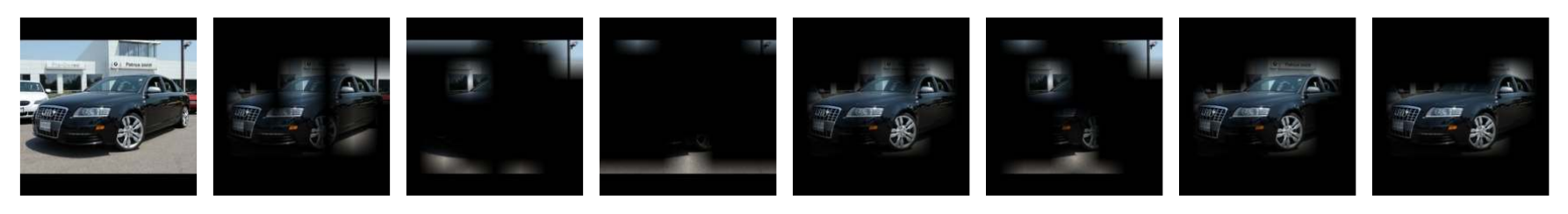}\\
\vspace{-0.2cm}
\includegraphics[width=\textwidth]{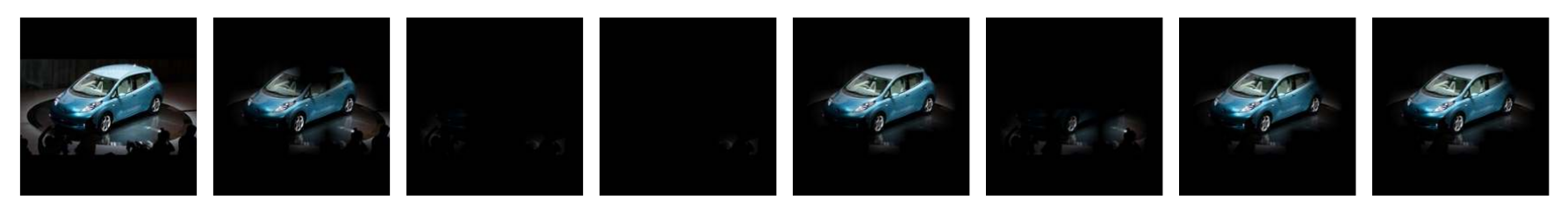}\\
\vspace{-0.2cm}
\includegraphics[width=\textwidth]{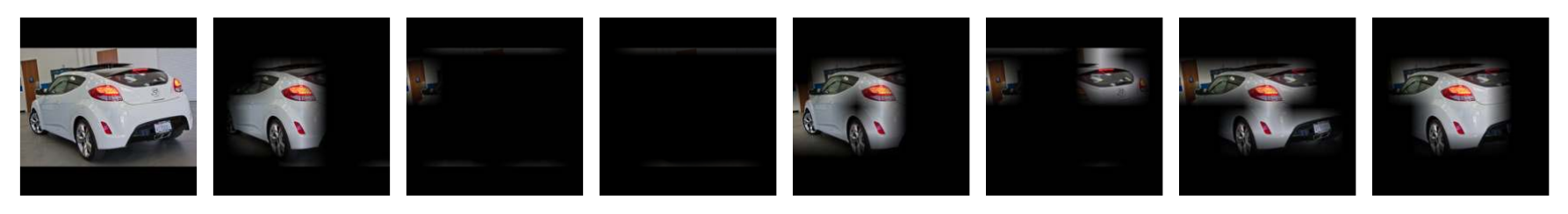}\\
\vspace{-2mm}
\noindent\scriptsize
\begin{minipage}[t]{0.125\textwidth}\centering (a) Input\end{minipage}%
\begin{minipage}[t]{0.125\textwidth}\centering (b) Ground-truth\end{minipage}%
\begin{minipage}[t]{0.125\textwidth}\centering (c) Zero-shot\end{minipage}%
\begin{minipage}[t]{0.125\textwidth}\centering (d) Finetuned on ImageNet\end{minipage}%
\begin{minipage}[t]{0.125\textwidth}\centering (e) Finetuned on ImageNet+X\end{minipage}%
\begin{minipage}[t]{0.125\textwidth}\centering (f) Finetuned on target\end{minipage}%
\begin{minipage}[t]{0.125\textwidth}\centering (g) Proposed\end{minipage}%
\begin{minipage}[t]{0.125\textwidth}\centering (h) Finetuned on target+X\end{minipage}%
\normalsize
\caption{Randomly selected visualization examples for Cars.}
\label{fig:appendix:examples:cars}
\end{figure}

\begin{figure}[p]
\centering
\includegraphics[width=\textwidth]{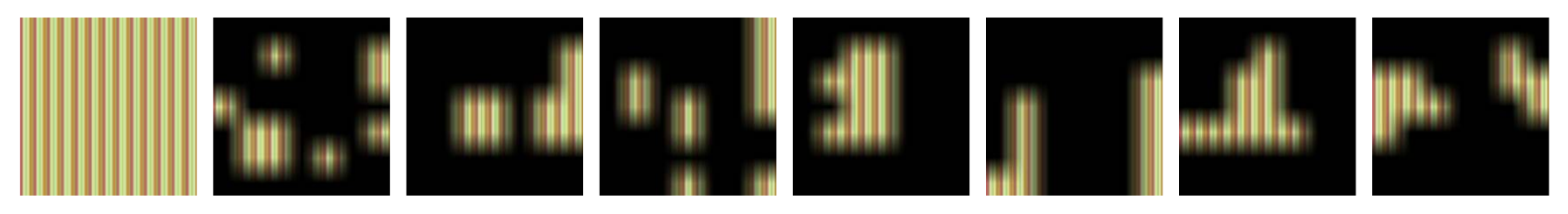}\\
\vspace{-0.2cm}
\includegraphics[width=\textwidth]{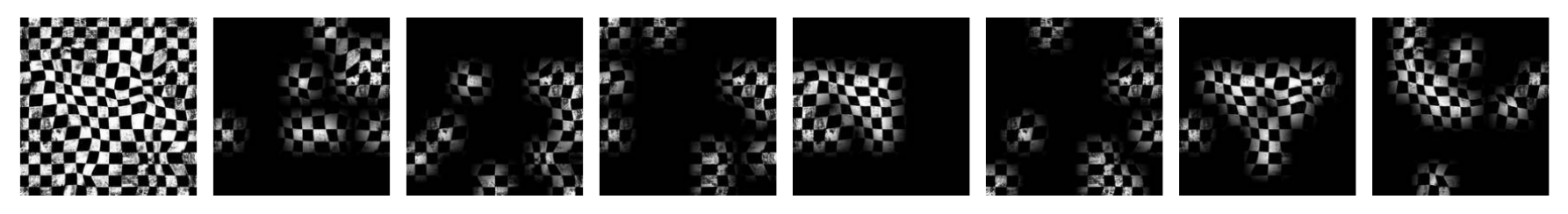}\\
\vspace{-0.2cm}
\includegraphics[width=\textwidth]{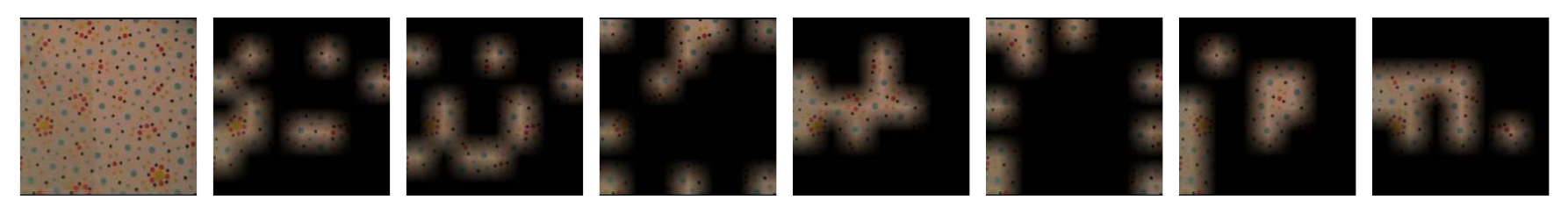}\\
\vspace{-0.2cm}
\includegraphics[width=\textwidth]{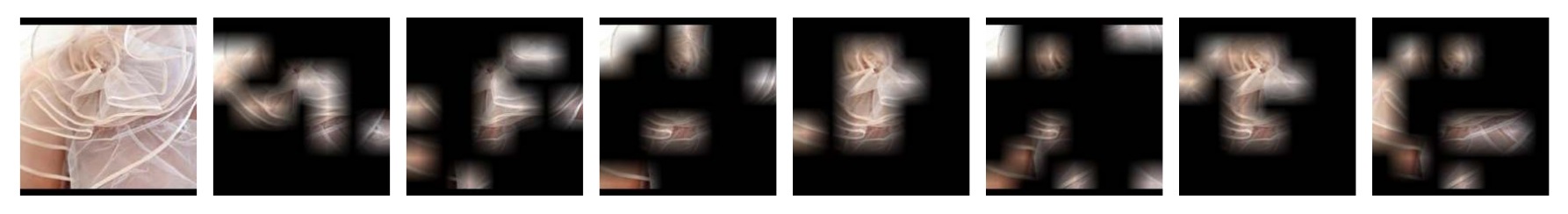}\\
\vspace{-0.2cm}
\includegraphics[width=\textwidth]{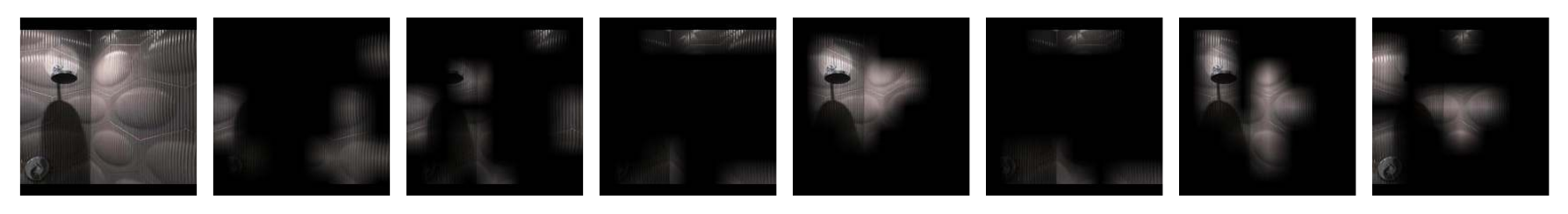}\\
\vspace{-0.2cm}
\includegraphics[width=\textwidth]{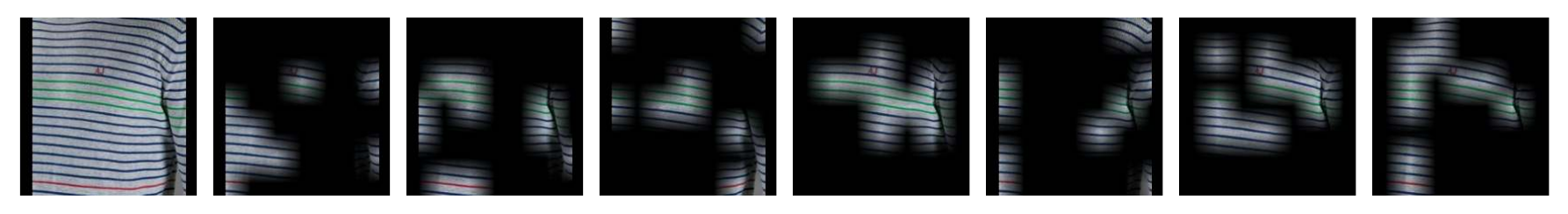}\\
\vspace{-0.2cm}
\includegraphics[width=\textwidth]{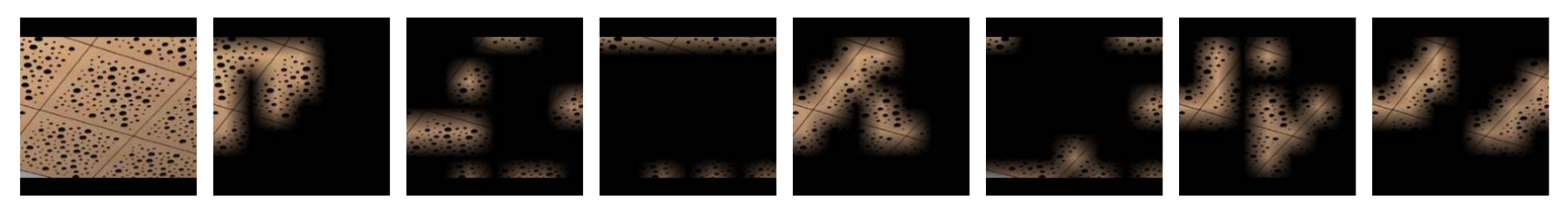}\\
\vspace{-0.2cm}
\includegraphics[width=\textwidth]{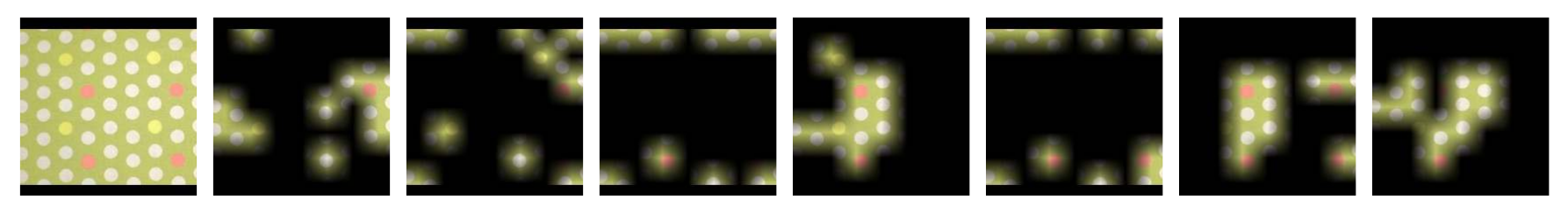}\\
\vspace{-0.2cm}
\includegraphics[width=\textwidth]{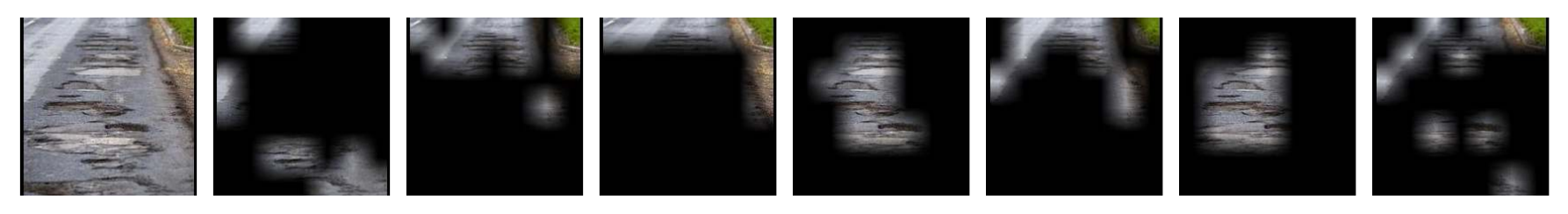}\\
\vspace{-0.2cm}
\includegraphics[width=\textwidth]{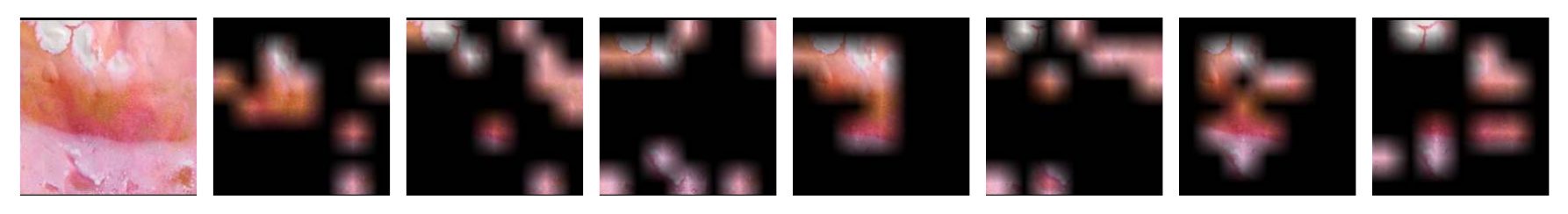}\\
\vspace{-2mm}
\noindent\scriptsize
\begin{minipage}[t]{0.125\textwidth}\centering (a) Input\end{minipage}%
\begin{minipage}[t]{0.125\textwidth}\centering (b) Ground-truth\end{minipage}%
\begin{minipage}[t]{0.125\textwidth}\centering (c) Zero-shot\end{minipage}%
\begin{minipage}[t]{0.125\textwidth}\centering (d) Finetuned on ImageNet\end{minipage}%
\begin{minipage}[t]{0.125\textwidth}\centering (e) Finetuned on ImageNet+X\end{minipage}%
\begin{minipage}[t]{0.125\textwidth}\centering (f) Finetuned on target\end{minipage}%
\begin{minipage}[t]{0.125\textwidth}\centering (g) Proposed\end{minipage}%
\begin{minipage}[t]{0.125\textwidth}\centering (h) Finetuned on target+X\end{minipage}%
\normalsize
\caption{Randomly selected visualization examples for DTD.}
\label{fig:appendix:examples:dtd}
\end{figure}

\begin{figure}[p]
\centering
\includegraphics[width=\textwidth]{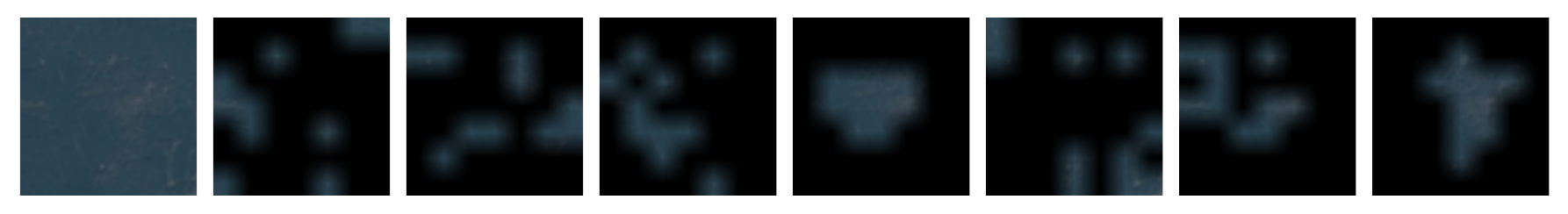}\\
\vspace{-0.2cm}
\includegraphics[width=\textwidth]{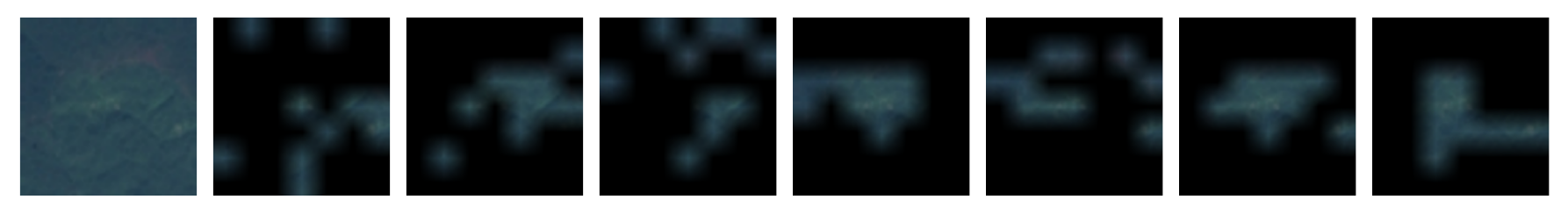}\\
\vspace{-0.2cm}
\includegraphics[width=\textwidth]{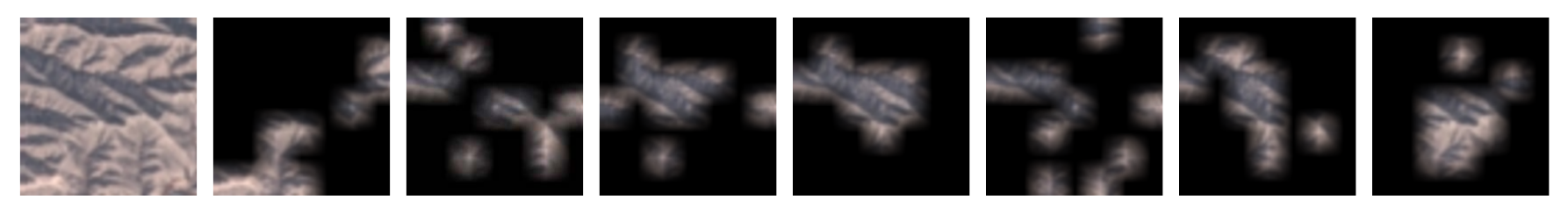}\\
\vspace{-0.2cm}
\includegraphics[width=\textwidth]{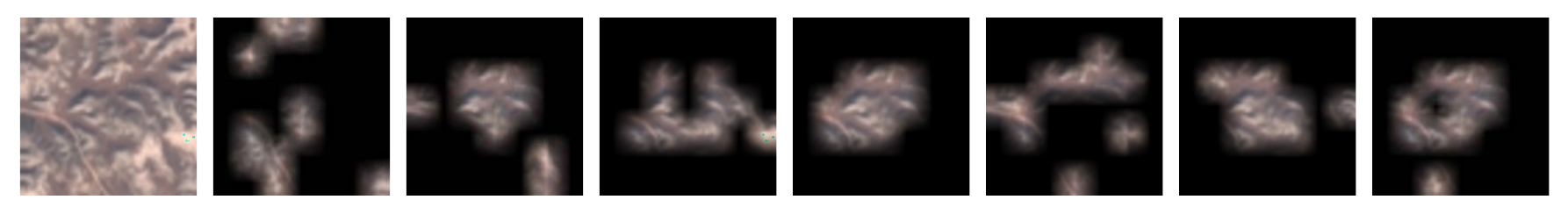}\\
\vspace{-0.2cm}
\includegraphics[width=\textwidth]{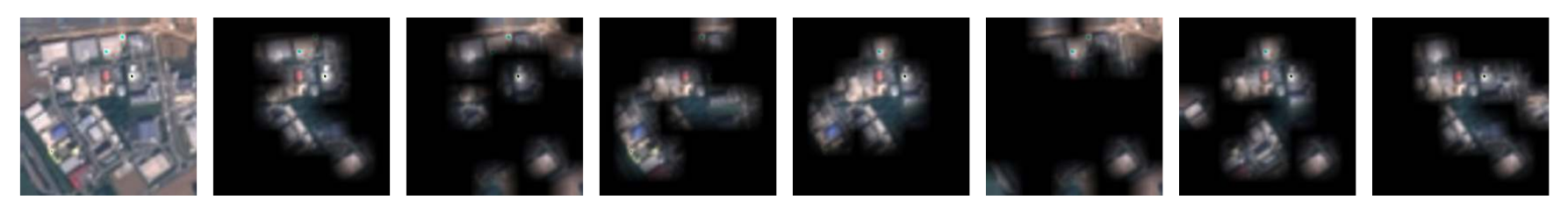}\\
\vspace{-0.2cm}
\includegraphics[width=\textwidth]{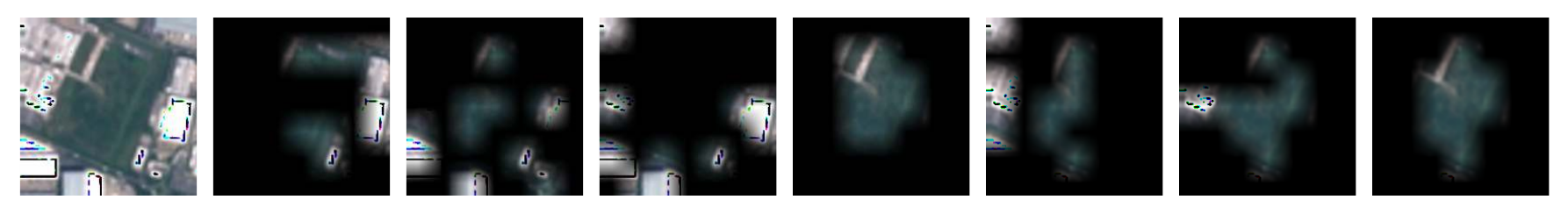}\\
\vspace{-0.2cm}
\includegraphics[width=\textwidth]{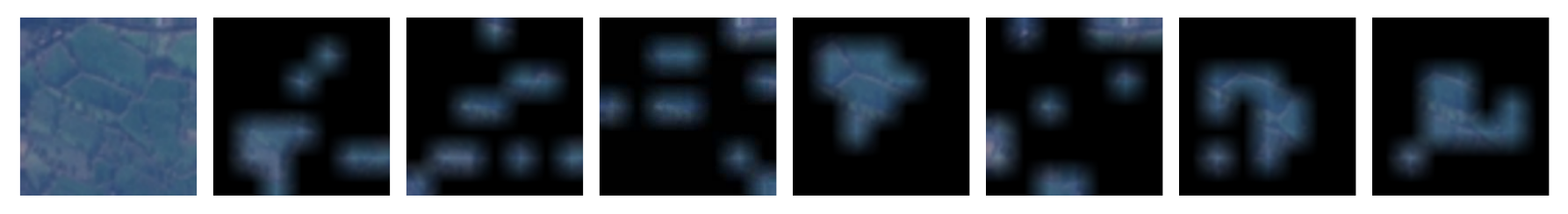}\\
\vspace{-0.2cm}
\includegraphics[width=\textwidth]{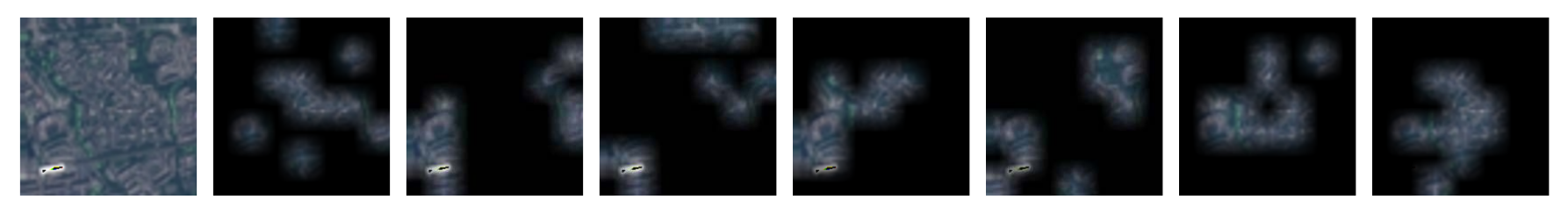}\\
\vspace{-0.2cm}
\includegraphics[width=\textwidth]{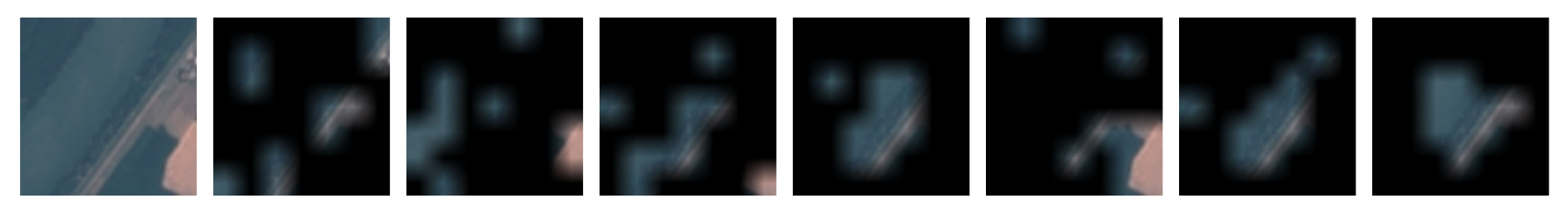}\\
\vspace{-0.2cm}
\includegraphics[width=\textwidth]{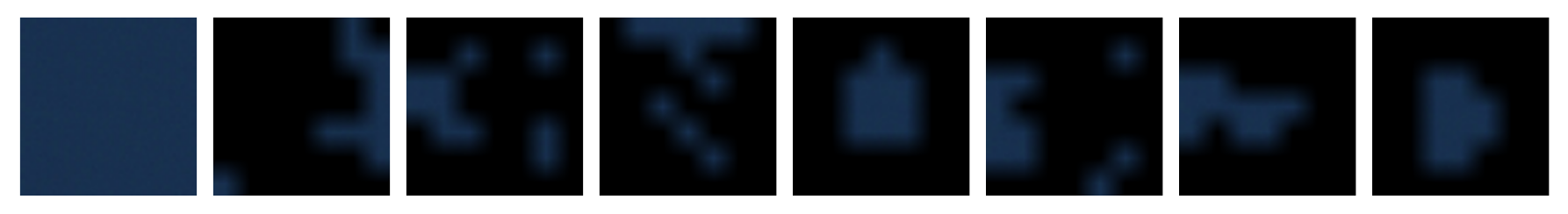}\\
\vspace{-2mm}
\noindent\scriptsize
\begin{minipage}[t]{0.125\textwidth}\centering (a) Input\end{minipage}%
\begin{minipage}[t]{0.125\textwidth}\centering (b) Ground-truth\end{minipage}%
\begin{minipage}[t]{0.125\textwidth}\centering (c) Zero-shot\end{minipage}%
\begin{minipage}[t]{0.125\textwidth}\centering (d) Finetuned on ImageNet\end{minipage}%
\begin{minipage}[t]{0.125\textwidth}\centering (e) Finetuned on ImageNet+X\end{minipage}%
\begin{minipage}[t]{0.125\textwidth}\centering (f) Finetuned on target\end{minipage}%
\begin{minipage}[t]{0.125\textwidth}\centering (g) Proposed\end{minipage}%
\begin{minipage}[t]{0.125\textwidth}\centering (h) Finetuned on target+X\end{minipage}%
\normalsize
\caption{Randomly selected visualization examples for EuroSAT.}
\label{fig:appendix:examples:eurosat}
\end{figure}

\begin{figure}[p]
\centering
\includegraphics[width=\textwidth]{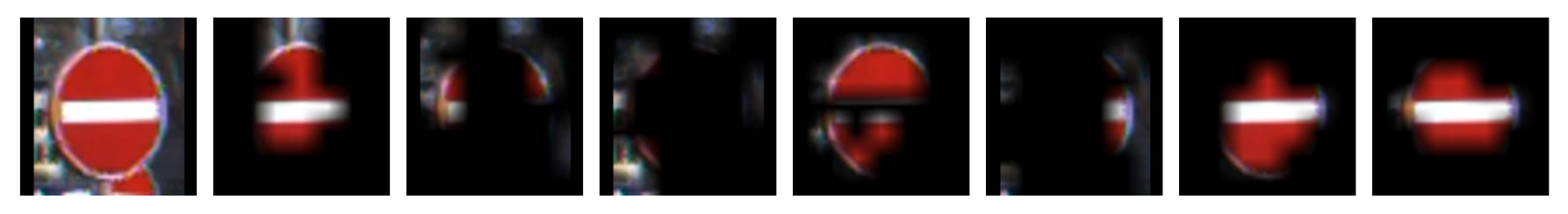}\\
\vspace{-0.2cm}
\includegraphics[width=\textwidth]{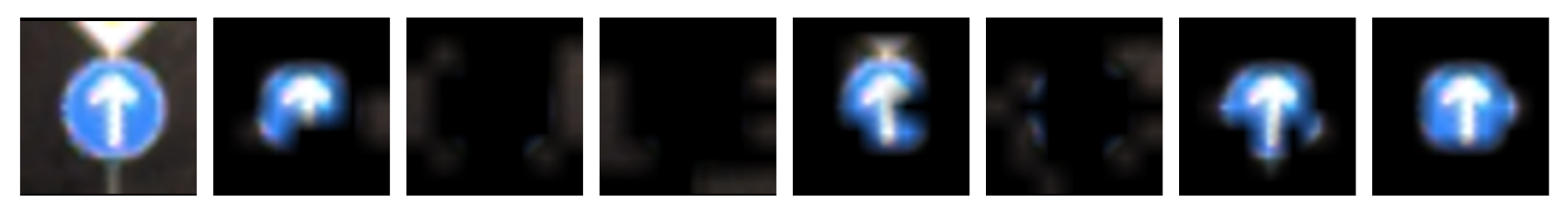}\\
\vspace{-0.2cm}
\includegraphics[width=\textwidth]{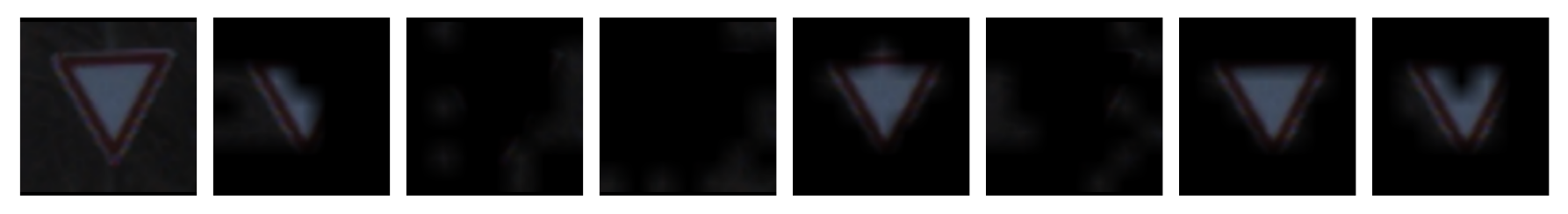}\\
\vspace{-0.2cm}
\includegraphics[width=\textwidth]{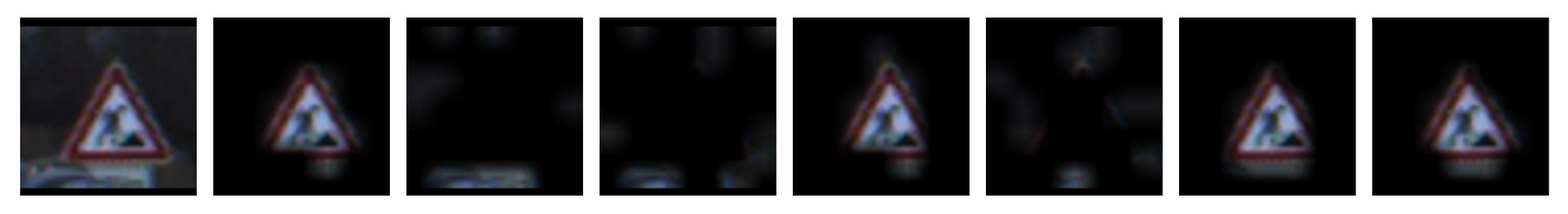}\\
\vspace{-0.2cm}
\includegraphics[width=\textwidth]{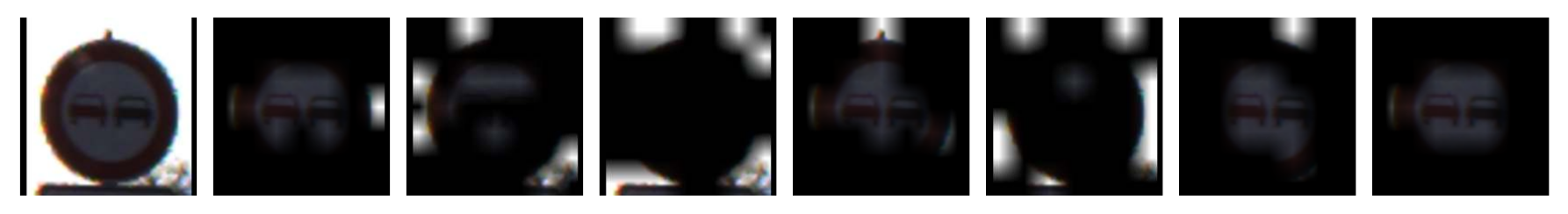}\\
\vspace{-0.2cm}
\includegraphics[width=\textwidth]{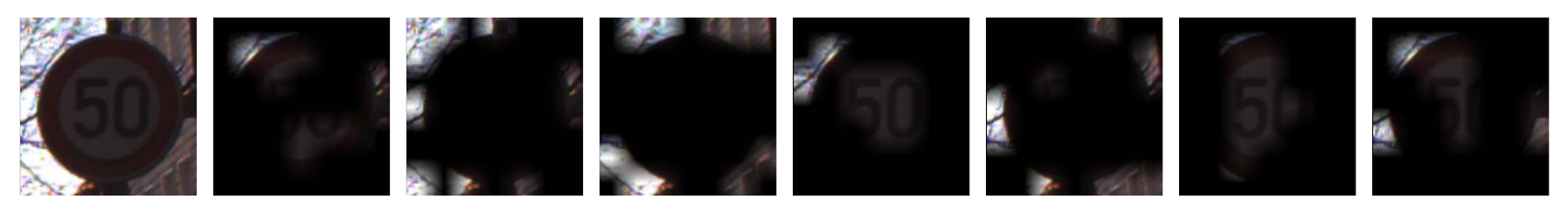}\\
\vspace{-0.2cm}
\includegraphics[width=\textwidth]{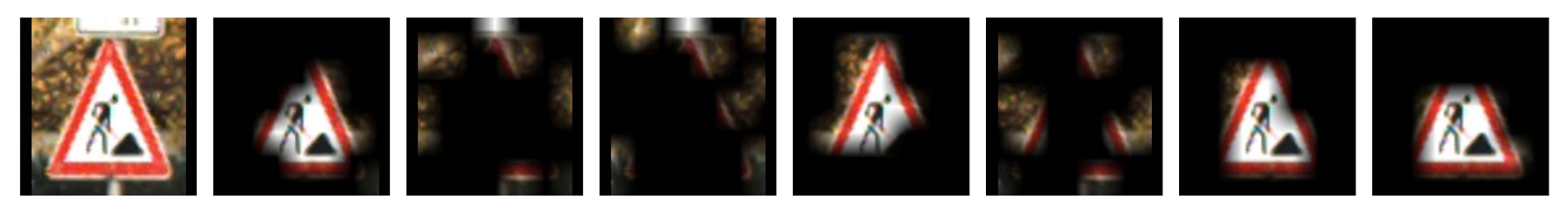}\\
\vspace{-0.2cm}
\includegraphics[width=\textwidth]{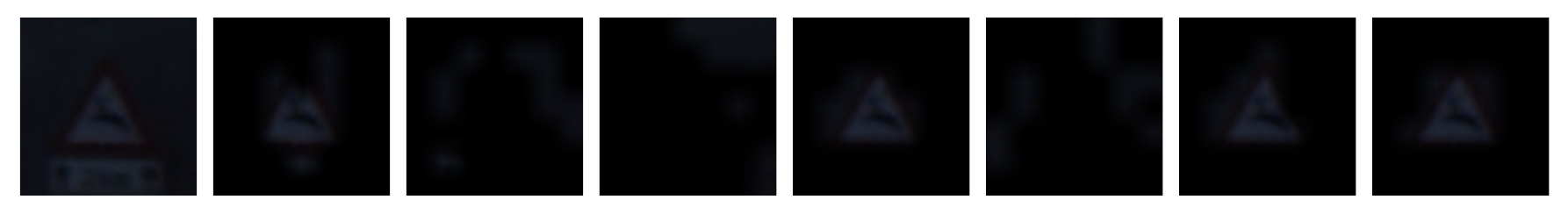}\\
\vspace{-0.2cm}
\includegraphics[width=\textwidth]{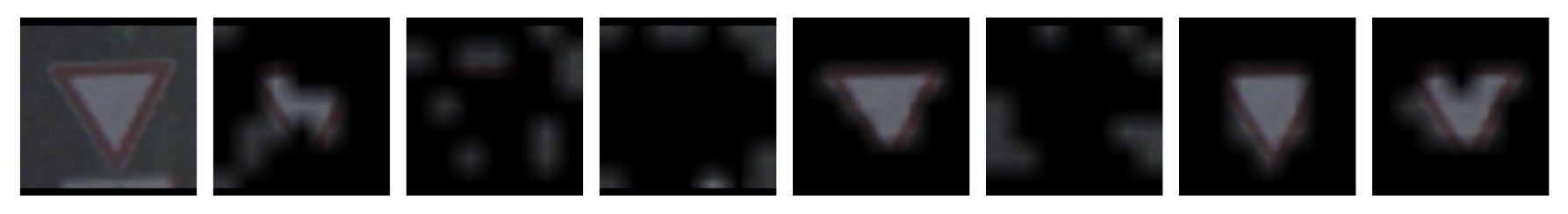}\\
\vspace{-0.2cm}
\includegraphics[width=\textwidth]{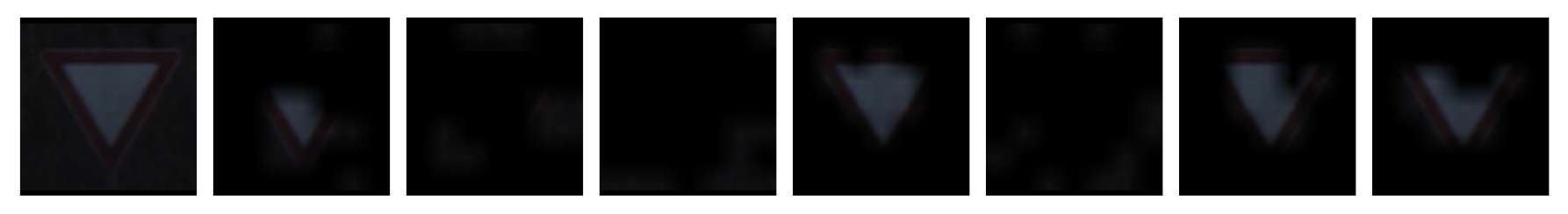}\\
\vspace{-2mm}
\noindent\scriptsize
\begin{minipage}[t]{0.125\textwidth}\centering (a) Input\end{minipage}%
\begin{minipage}[t]{0.125\textwidth}\centering (b) Ground-truth\end{minipage}%
\begin{minipage}[t]{0.125\textwidth}\centering (c) Zero-shot\end{minipage}%
\begin{minipage}[t]{0.125\textwidth}\centering (d) Finetuned on ImageNet\end{minipage}%
\begin{minipage}[t]{0.125\textwidth}\centering (e) Finetuned on ImageNet+X\end{minipage}%
\begin{minipage}[t]{0.125\textwidth}\centering (f) Finetuned on target\end{minipage}%
\begin{minipage}[t]{0.125\textwidth}\centering (g) Proposed\end{minipage}%
\begin{minipage}[t]{0.125\textwidth}\centering (h) Finetuned on target+X\end{minipage}%
\normalsize
\caption{Randomly selected visualization examples for GTSRB.}
\label{fig:appendix:examples:gtsrb}
\end{figure}

\begin{figure}[p]
\centering
\includegraphics[width=\textwidth]{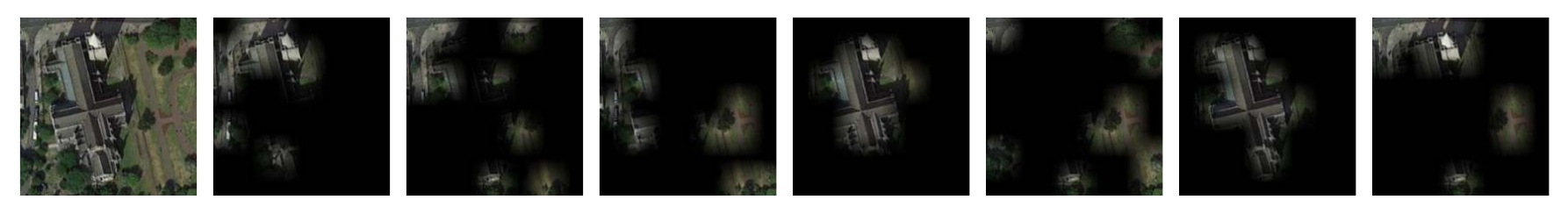}\\
\vspace{-0.2cm}
\includegraphics[width=\textwidth]{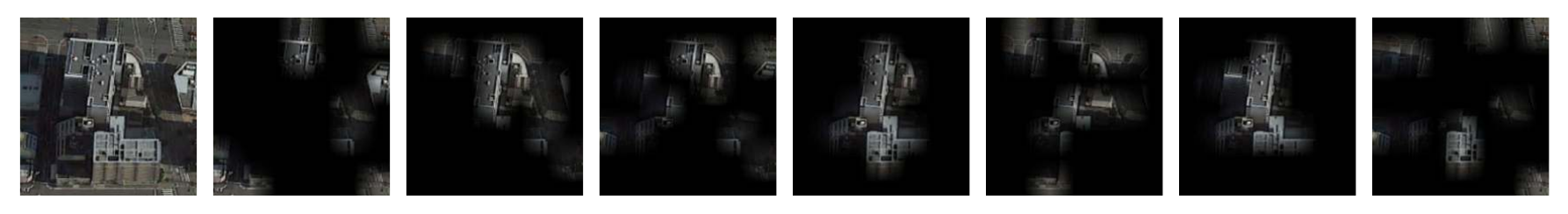}\\
\vspace{-0.2cm}
\includegraphics[width=\textwidth]{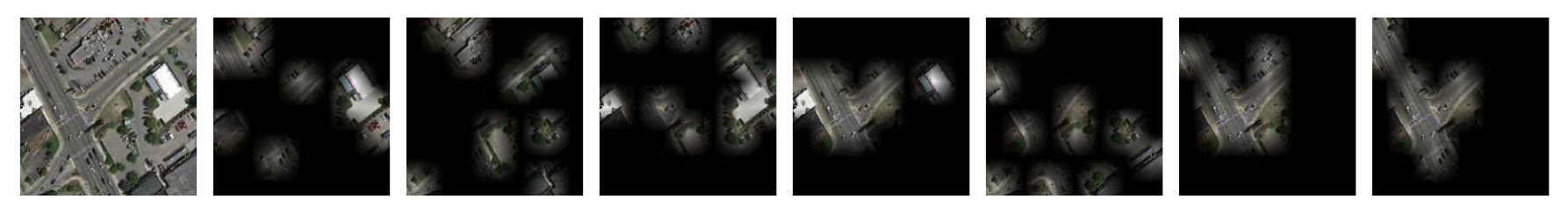}\\
\vspace{-0.2cm}
\includegraphics[width=\textwidth]{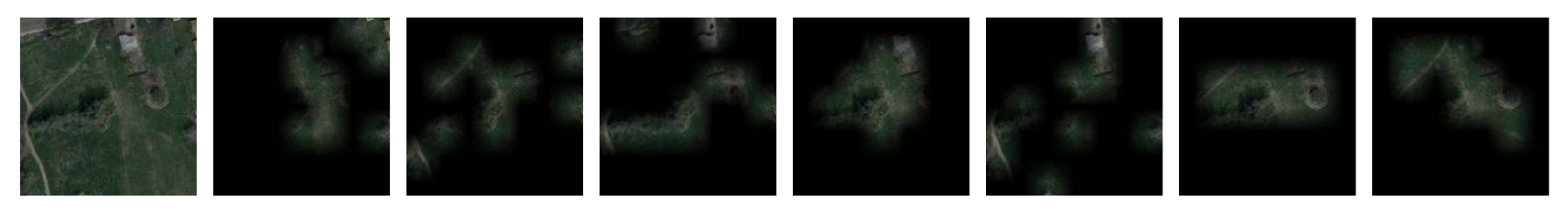}\\
\vspace{-0.2cm}
\includegraphics[width=\textwidth]{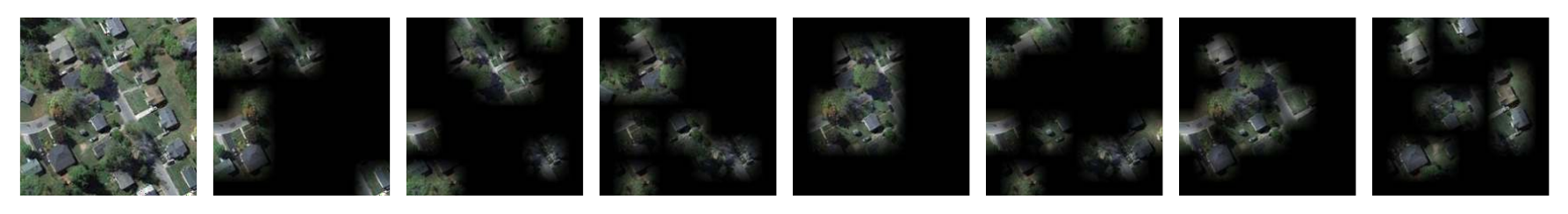}\\
\vspace{-0.2cm}
\includegraphics[width=\textwidth]{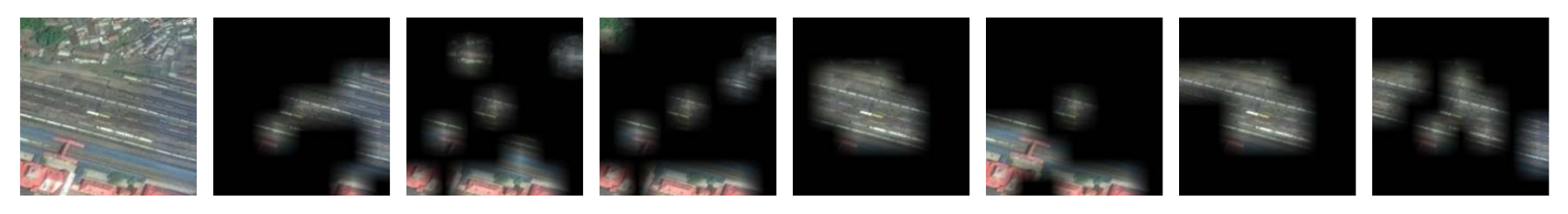}\\
\vspace{-0.2cm}
\includegraphics[width=\textwidth]{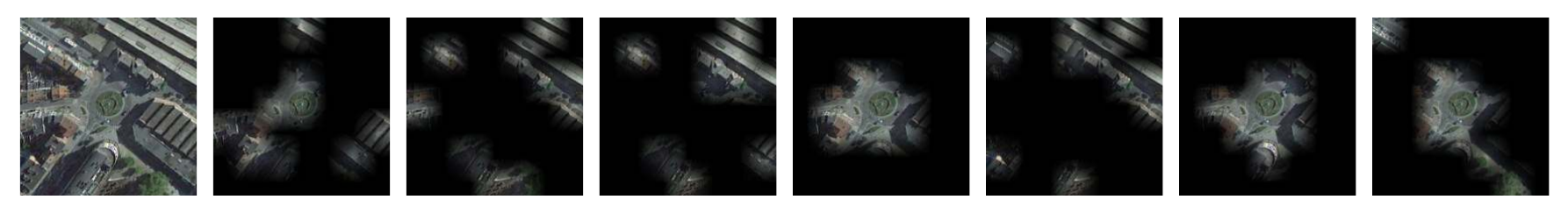}\\
\vspace{-0.2cm}
\includegraphics[width=\textwidth]{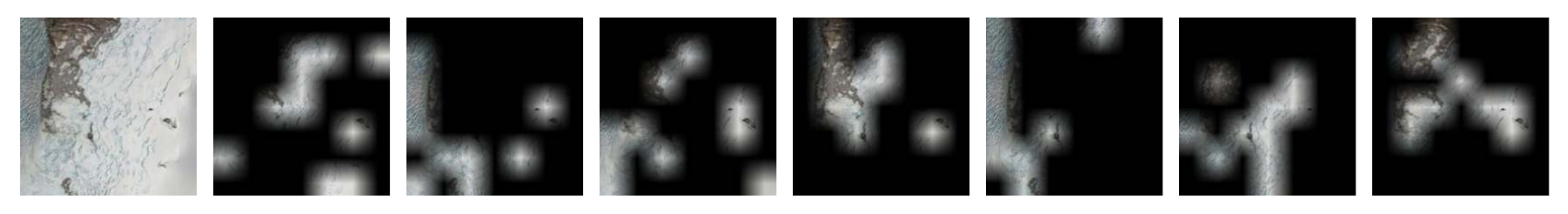}\\
\vspace{-0.2cm}
\includegraphics[width=\textwidth]{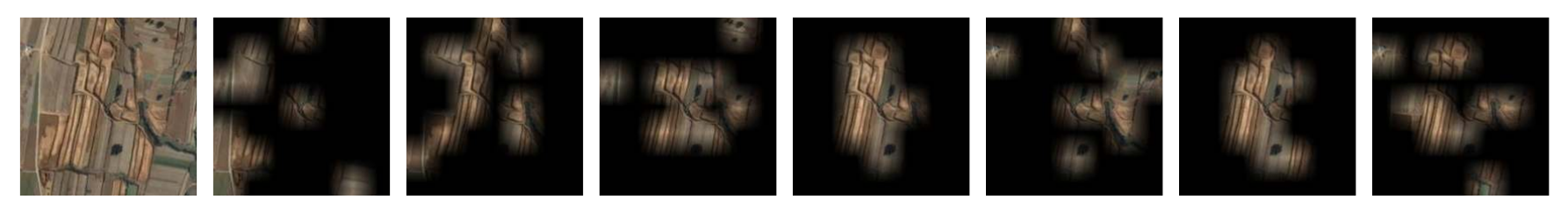}\\
\vspace{-0.2cm}
\includegraphics[width=\textwidth]{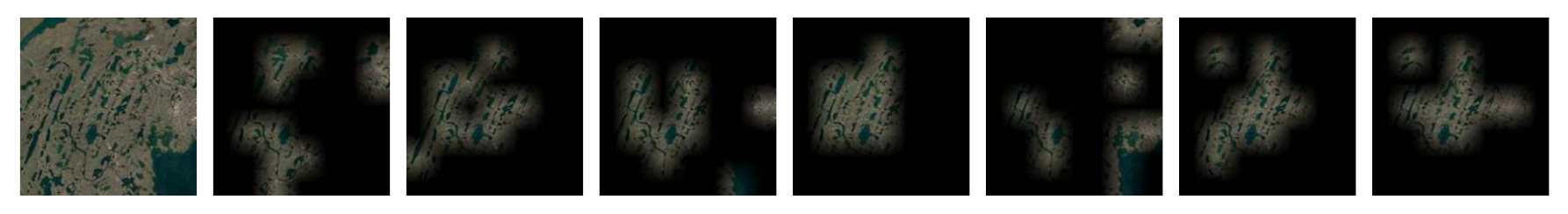}\\
\vspace{-2mm}
\noindent\scriptsize
\begin{minipage}[t]{0.125\textwidth}\centering (a) Input\end{minipage}%
\begin{minipage}[t]{0.125\textwidth}\centering (b) Ground-truth\end{minipage}%
\begin{minipage}[t]{0.125\textwidth}\centering (c) Zero-shot\end{minipage}%
\begin{minipage}[t]{0.125\textwidth}\centering (d) Finetuned on ImageNet\end{minipage}%
\begin{minipage}[t]{0.125\textwidth}\centering (e) Finetuned on ImageNet+X\end{minipage}%
\begin{minipage}[t]{0.125\textwidth}\centering (f) Finetuned on target\end{minipage}%
\begin{minipage}[t]{0.125\textwidth}\centering (g) Proposed\end{minipage}%
\begin{minipage}[t]{0.125\textwidth}\centering (h) Finetuned on target+X\end{minipage}%
\normalsize
\caption{Randomly selected visualization examples for RESISC45.}
\label{fig:appendix:examples:resisc45}
\end{figure}

\begin{figure}[p]
\centering
\includegraphics[width=\textwidth]{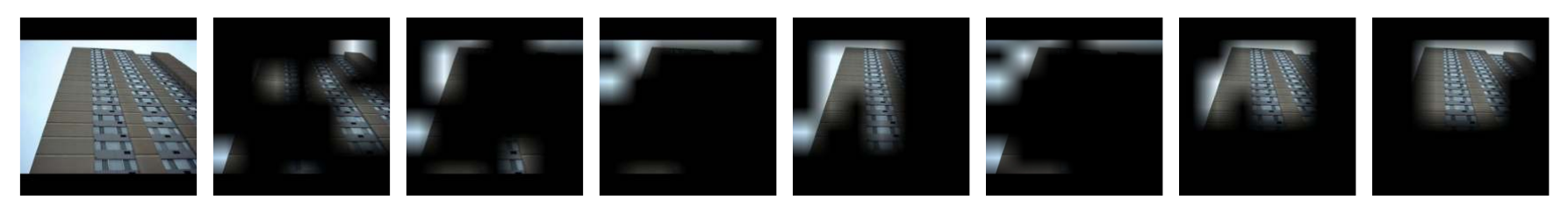}\\
\vspace{-0.2cm}
\includegraphics[width=\textwidth]{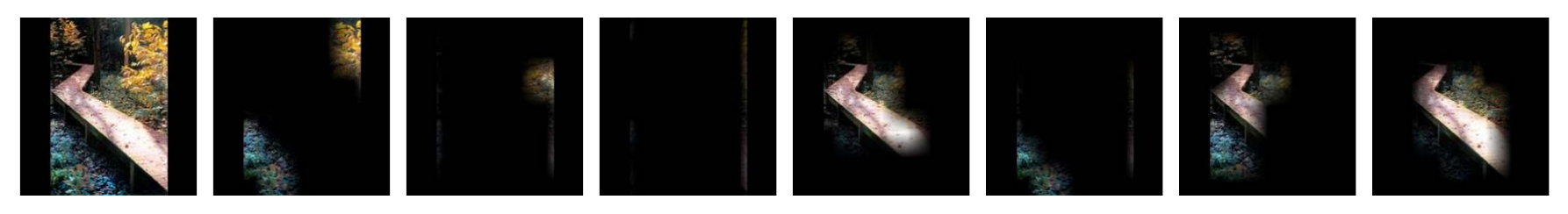}\\
\vspace{-0.2cm}
\includegraphics[width=\textwidth]{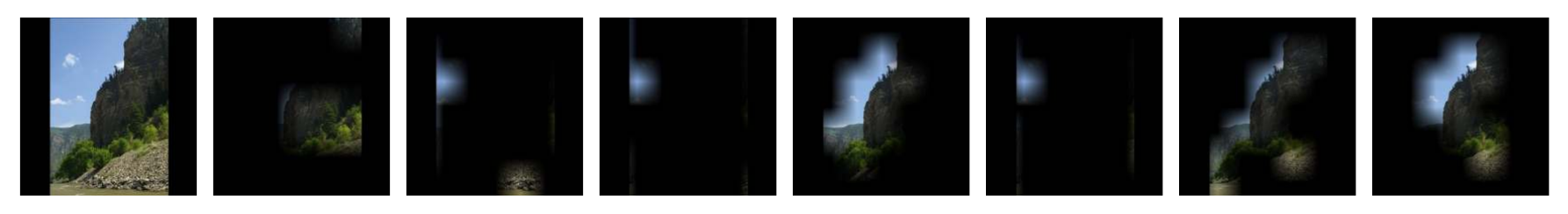}\\
\vspace{-0.2cm}
\includegraphics[width=\textwidth]{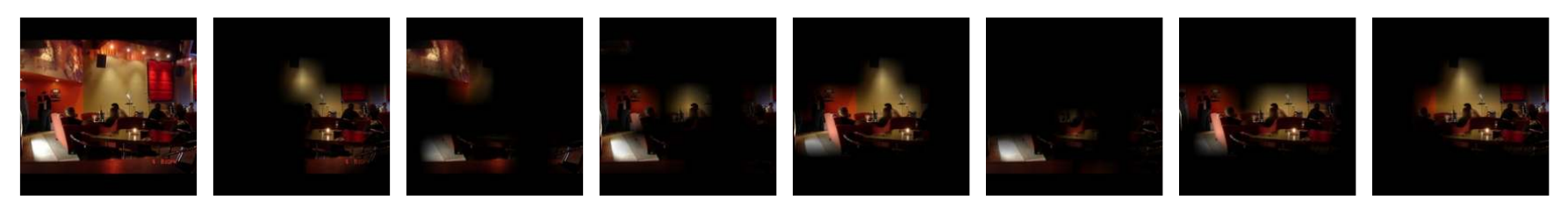}\\
\vspace{-0.2cm}
\includegraphics[width=\textwidth]{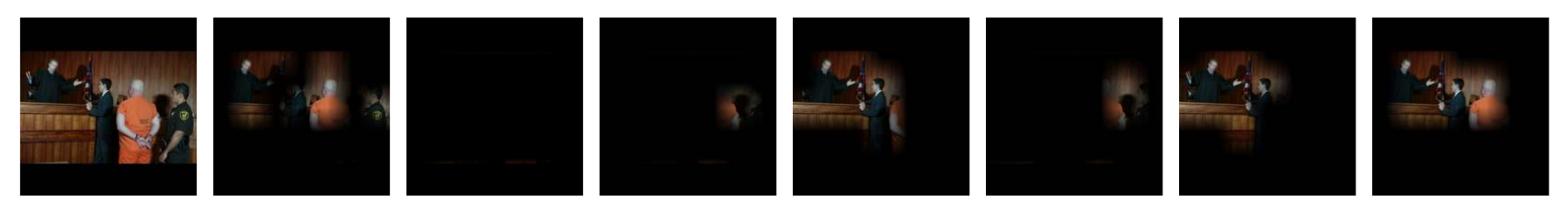}\\
\vspace{-0.2cm}
\includegraphics[width=\textwidth]{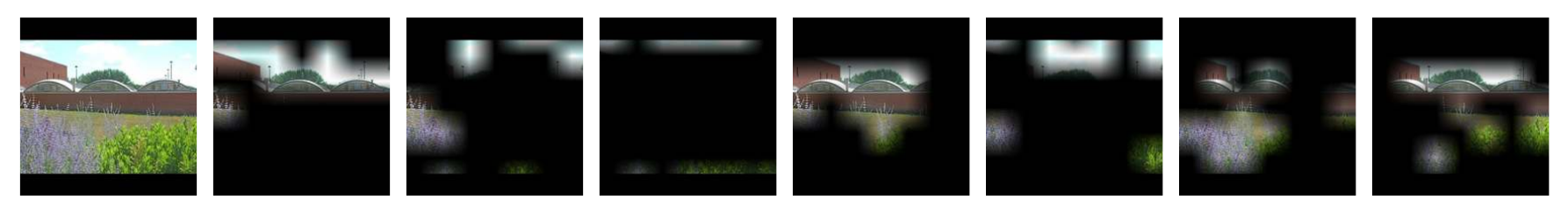}\\
\vspace{-0.2cm}
\includegraphics[width=\textwidth]{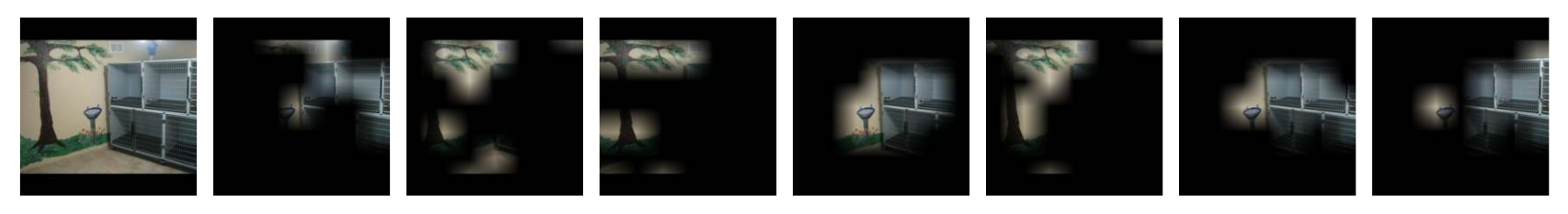}\\
\vspace{-0.2cm}
\includegraphics[width=\textwidth]{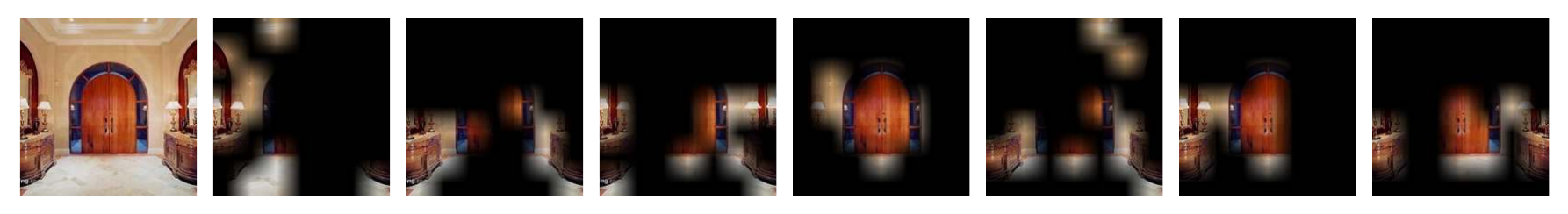}\\
\vspace{-0.2cm}
\includegraphics[width=\textwidth]{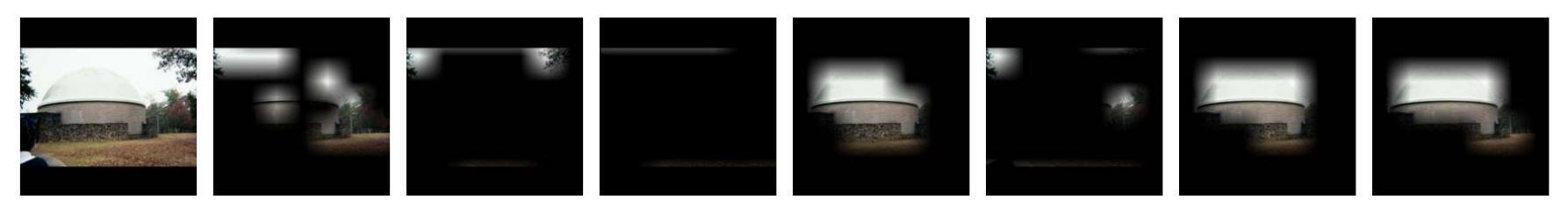}\\
\vspace{-0.2cm}
\includegraphics[width=\textwidth]{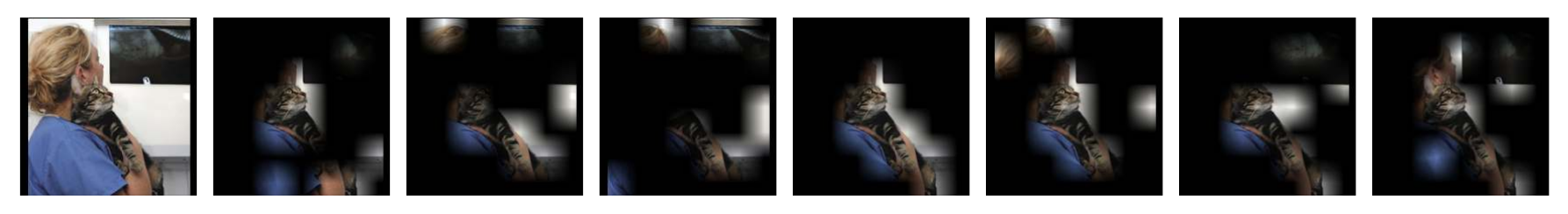}\\
\vspace{-2mm}
\noindent\scriptsize
\begin{minipage}[t]{0.125\textwidth}\centering (a) Input\end{minipage}%
\begin{minipage}[t]{0.125\textwidth}\centering (b) Ground-truth\end{minipage}%
\begin{minipage}[t]{0.125\textwidth}\centering (c) Zero-shot\end{minipage}%
\begin{minipage}[t]{0.125\textwidth}\centering (d) Finetuned on ImageNet\end{minipage}%
\begin{minipage}[t]{0.125\textwidth}\centering (e) Finetuned on ImageNet+X\end{minipage}%
\begin{minipage}[t]{0.125\textwidth}\centering (f) Finetuned on target\end{minipage}%
\begin{minipage}[t]{0.125\textwidth}\centering (g) Proposed\end{minipage}%
\begin{minipage}[t]{0.125\textwidth}\centering (h) Finetuned on target+X\end{minipage}%
\normalsize
\caption{Randomly selected visualization examples for SUN397.}
\label{fig:appendix:examples:sun397}
\end{figure}

\begin{figure}[p]
\centering
\includegraphics[width=\textwidth]{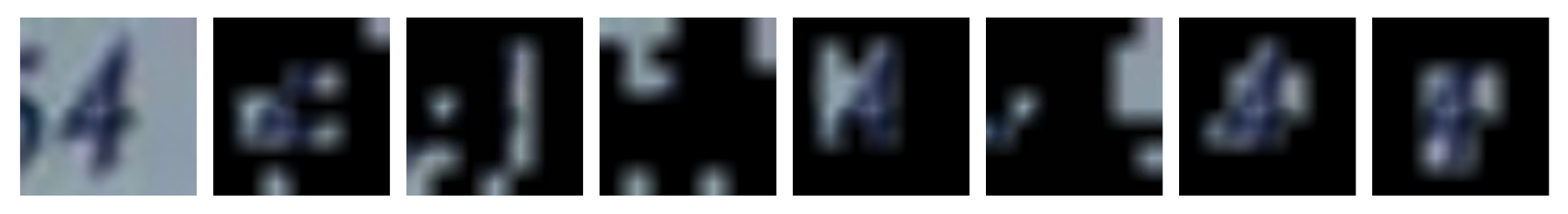}\\
\vspace{-0.2cm}
\includegraphics[width=\textwidth]{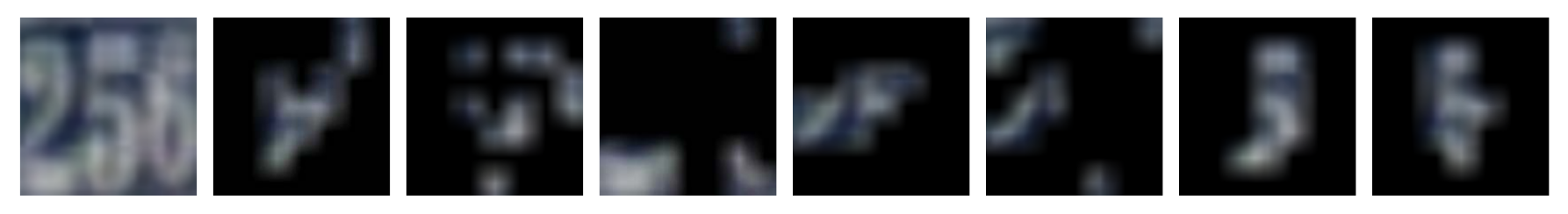}\\
\vspace{-0.2cm}
\includegraphics[width=\textwidth]{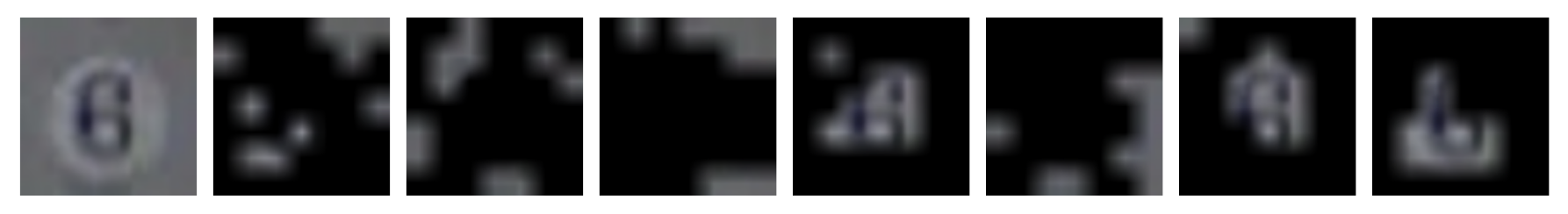}\\
\vspace{-0.2cm}
\includegraphics[width=\textwidth]{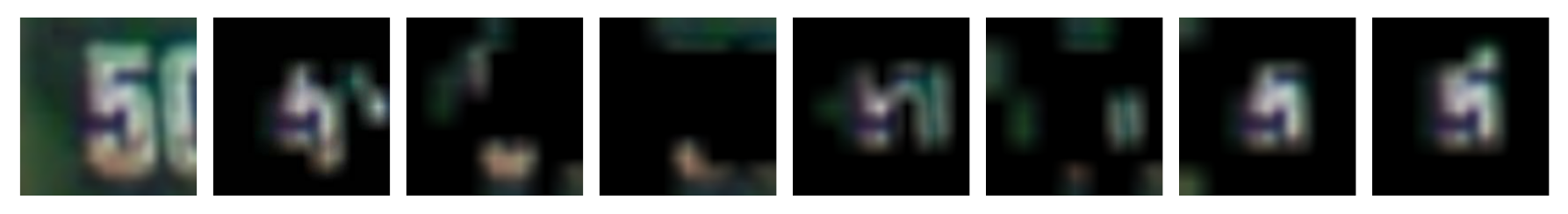}\\
\vspace{-0.2cm}
\includegraphics[width=\textwidth]{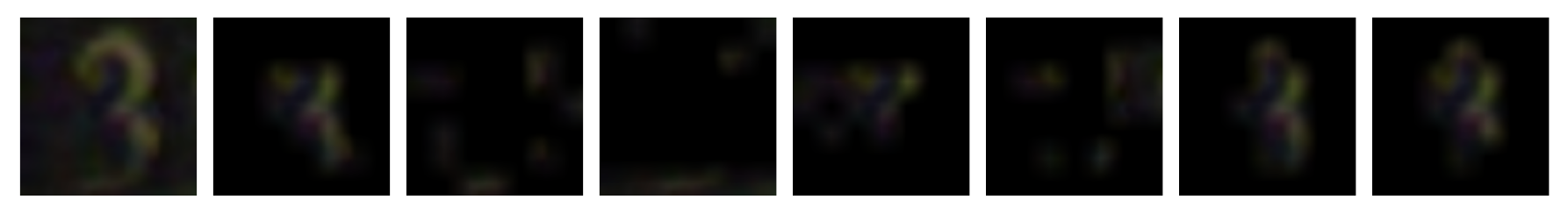}\\
\vspace{-0.2cm}
\includegraphics[width=\textwidth]{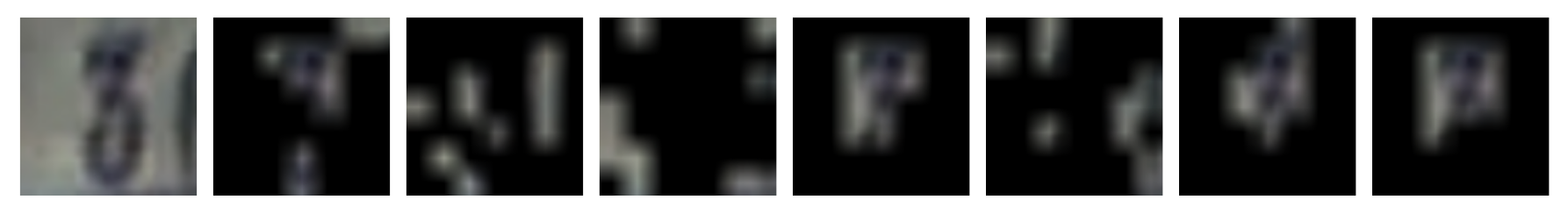}\\
\vspace{-0.2cm}
\includegraphics[width=\textwidth]{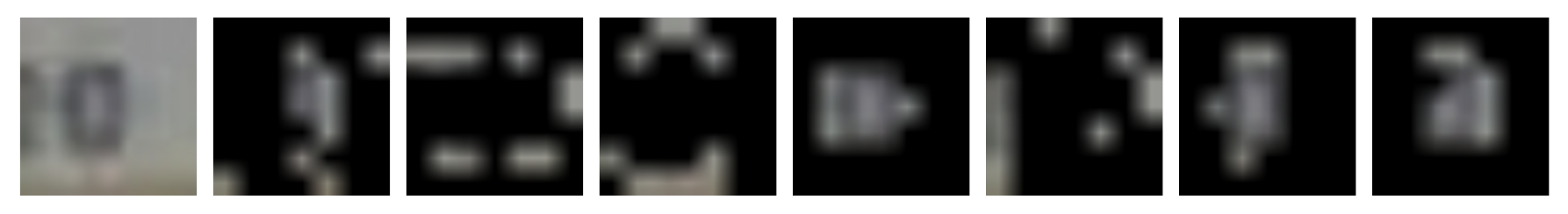}\\
\vspace{-0.2cm}
\includegraphics[width=\textwidth]{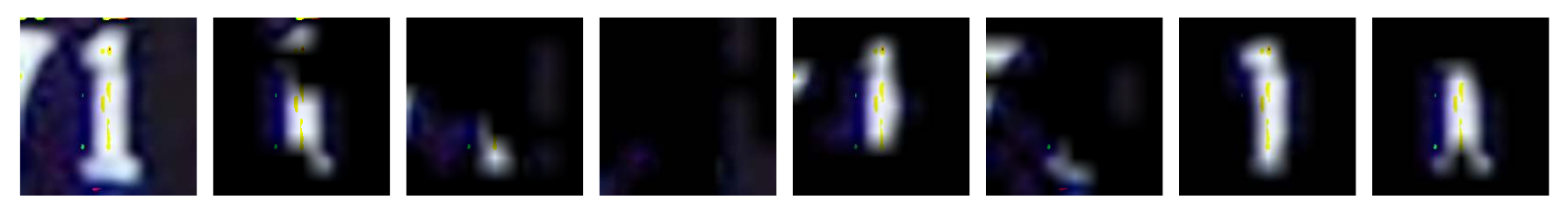}\\
\vspace{-0.2cm}
\includegraphics[width=\textwidth]{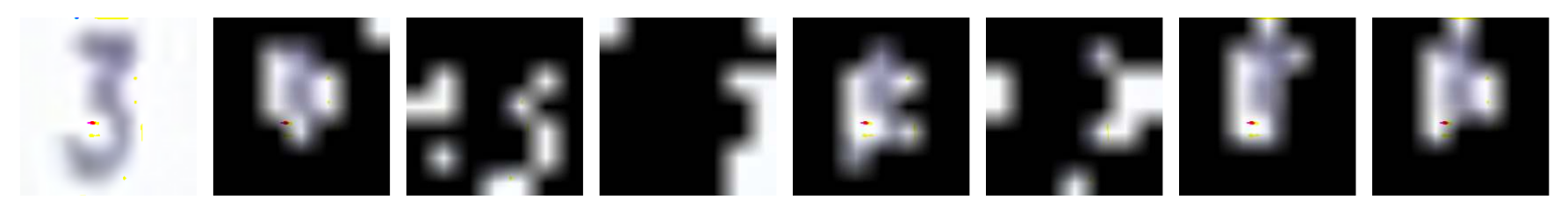}\\
\vspace{-0.2cm}
\includegraphics[width=\textwidth]{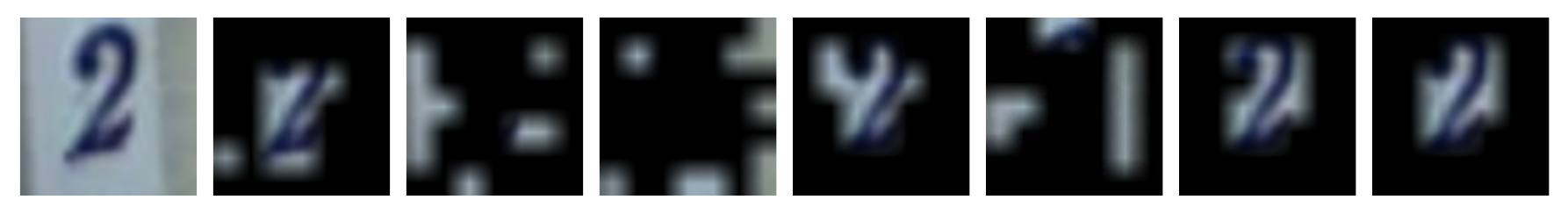}\\
\vspace{-2mm}
\noindent\scriptsize
\begin{minipage}[t]{0.125\textwidth}\centering (a) Input\end{minipage}%
\begin{minipage}[t]{0.125\textwidth}\centering (b) Ground-truth\end{minipage}%
\begin{minipage}[t]{0.125\textwidth}\centering (c) Zero-shot\end{minipage}%
\begin{minipage}[t]{0.125\textwidth}\centering (d) Finetuned on ImageNet\end{minipage}%
\begin{minipage}[t]{0.125\textwidth}\centering (e) Finetuned on ImageNet+X\end{minipage}%
\begin{minipage}[t]{0.125\textwidth}\centering (f) Finetuned on target\end{minipage}%
\begin{minipage}[t]{0.125\textwidth}\centering (g) Proposed\end{minipage}%
\begin{minipage}[t]{0.125\textwidth}\centering (h) Finetuned on target+X\end{minipage}%
\normalsize
\caption{Randomly selected visualization examples for SVHN. Visual artifacts in the input images are due to the denormalization and upscaling process.}
\label{fig:appendix:examples:svhn}
\end{figure}

\begin{figure}[p]
\centering
\includegraphics[width=\textwidth]{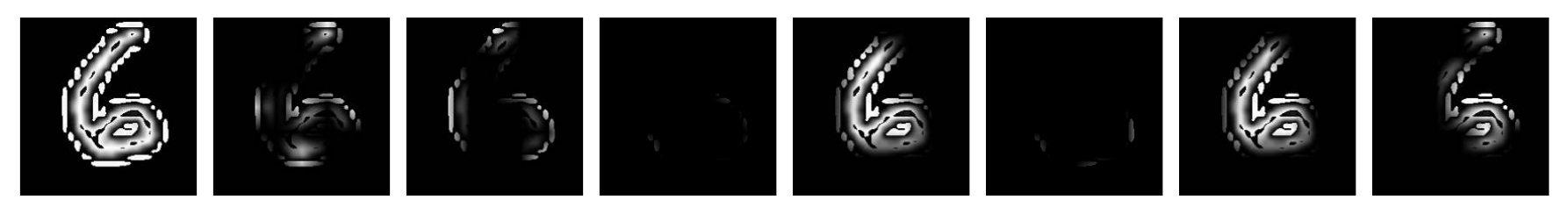}\\
\vspace{-0.2cm}
\includegraphics[width=\textwidth]{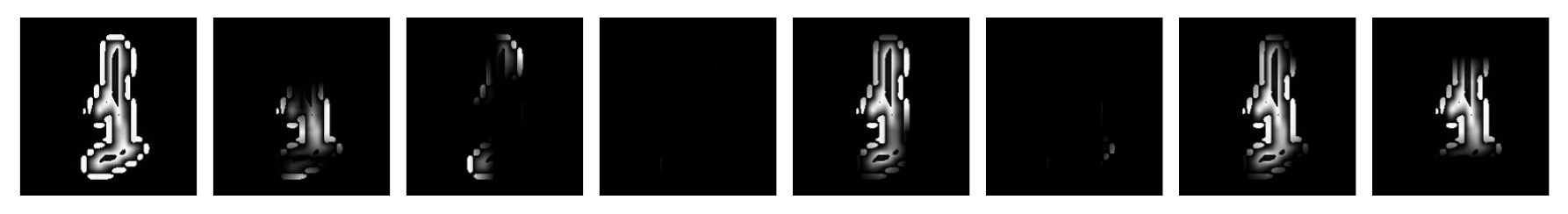}\\
\vspace{-0.2cm}
\includegraphics[width=\textwidth]{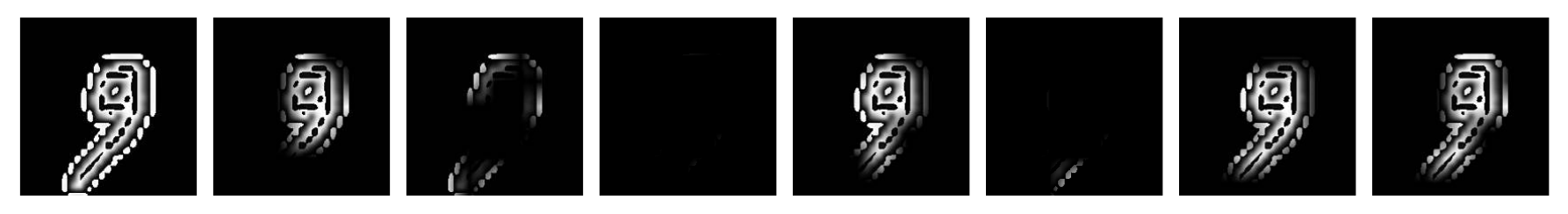}\\
\vspace{-0.2cm}
\includegraphics[width=\textwidth]{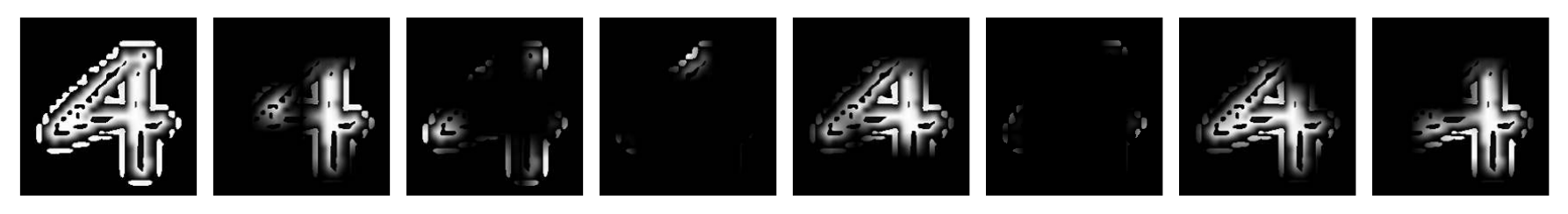}\\
\vspace{-0.2cm}
\includegraphics[width=\textwidth]{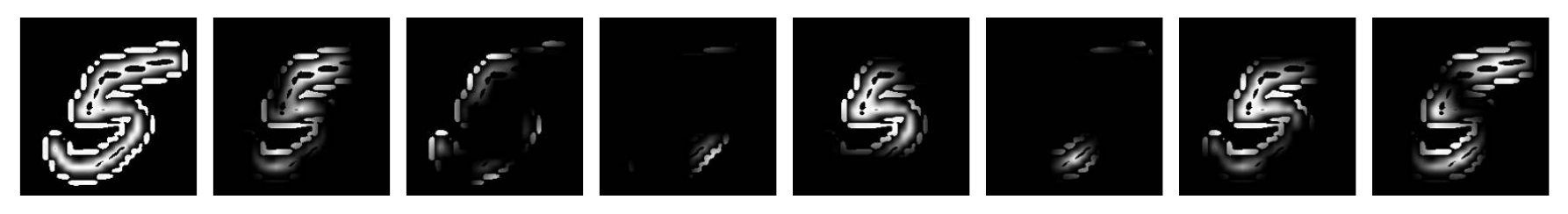}\\
\vspace{-0.2cm}
\includegraphics[width=\textwidth]{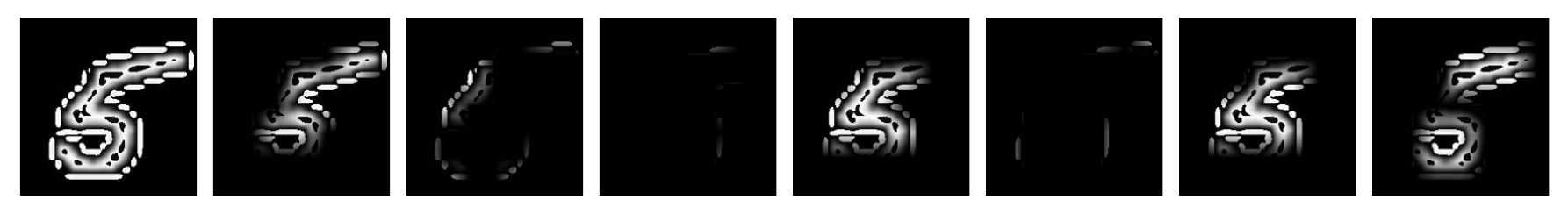}\\
\vspace{-0.2cm}
\includegraphics[width=\textwidth]{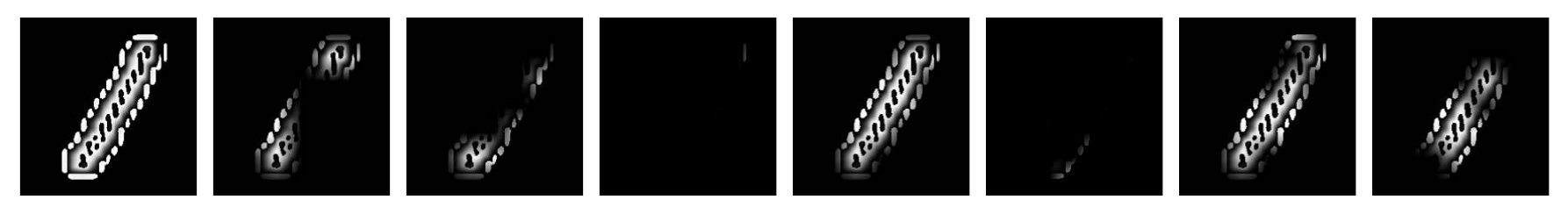}\\
\vspace{-0.2cm}
\includegraphics[width=\textwidth]{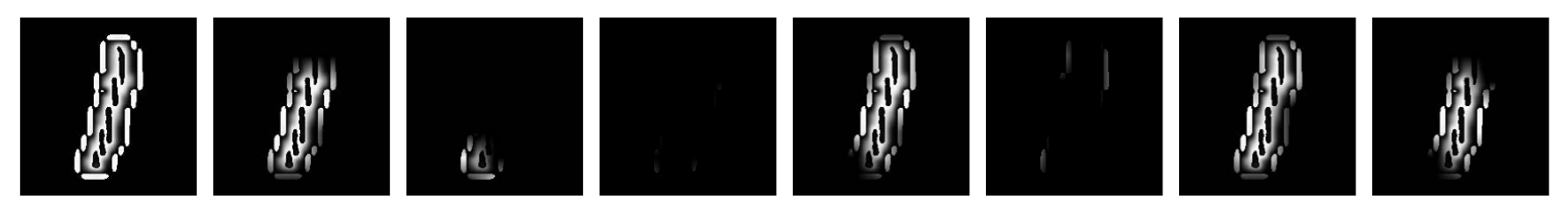}\\
\vspace{-0.2cm}
\includegraphics[width=\textwidth]{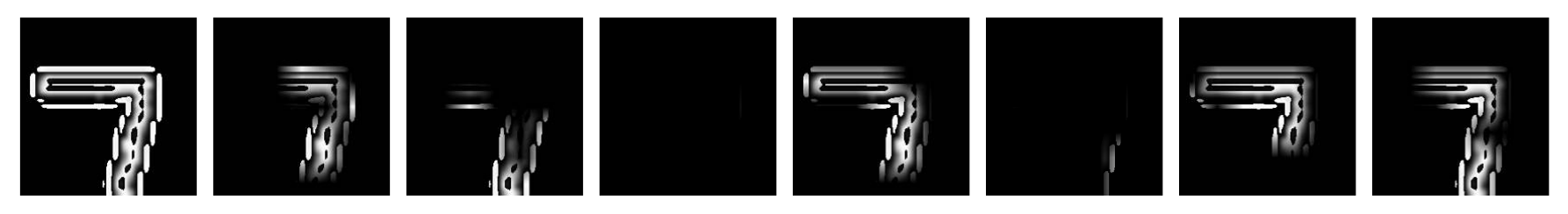}\\
\vspace{-0.2cm}
\includegraphics[width=\textwidth]{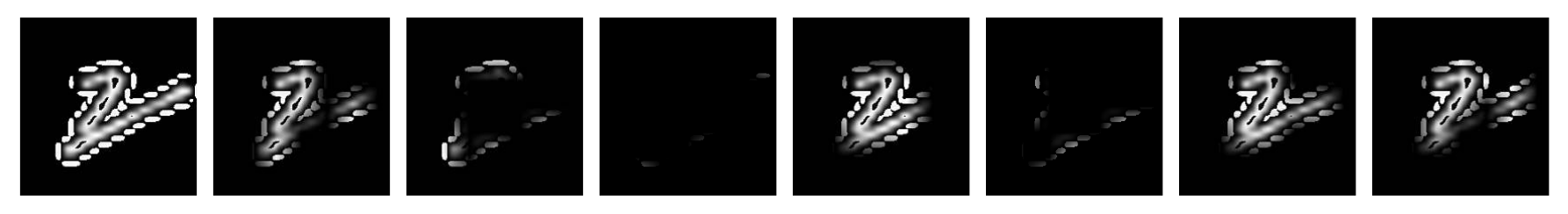}\\
\vspace{-2mm}
\noindent\scriptsize
\begin{minipage}[t]{0.125\textwidth}\centering (a) Input\end{minipage}%
\begin{minipage}[t]{0.125\textwidth}\centering (b) Ground-truth\end{minipage}%
\begin{minipage}[t]{0.125\textwidth}\centering (c) Zero-shot\end{minipage}%
\begin{minipage}[t]{0.125\textwidth}\centering (d) Finetuned on ImageNet\end{minipage}%
\begin{minipage}[t]{0.125\textwidth}\centering (e) Finetuned on ImageNet+X\end{minipage}%
\begin{minipage}[t]{0.125\textwidth}\centering (f) Finetuned on target\end{minipage}%
\begin{minipage}[t]{0.125\textwidth}\centering (g) Proposed\end{minipage}%
\begin{minipage}[t]{0.125\textwidth}\centering (h) Finetuned on target+X\end{minipage}%
\normalsize
\caption{Randomly selected visualization examples for MNIST. Visual artifacts in the input images are due to the denormalization and upscaling process.}
\label{fig:appendix:examples:mnist}
\end{figure}

\begin{figure}[p]
\centering
\includegraphics[width=\textwidth]{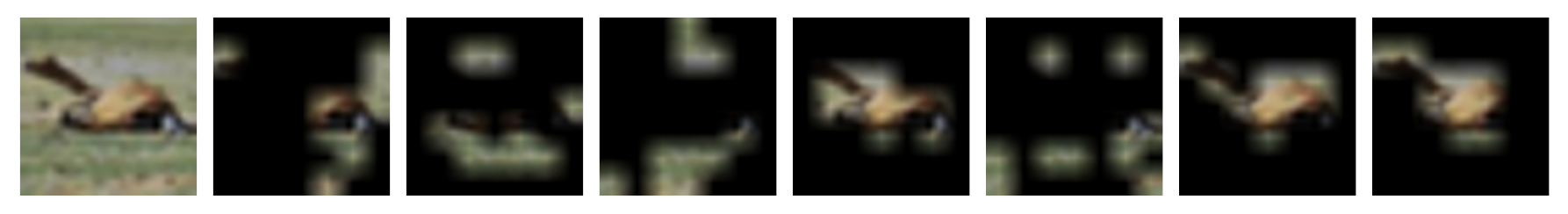}\\
\vspace{-0.2cm}
\includegraphics[width=\textwidth]{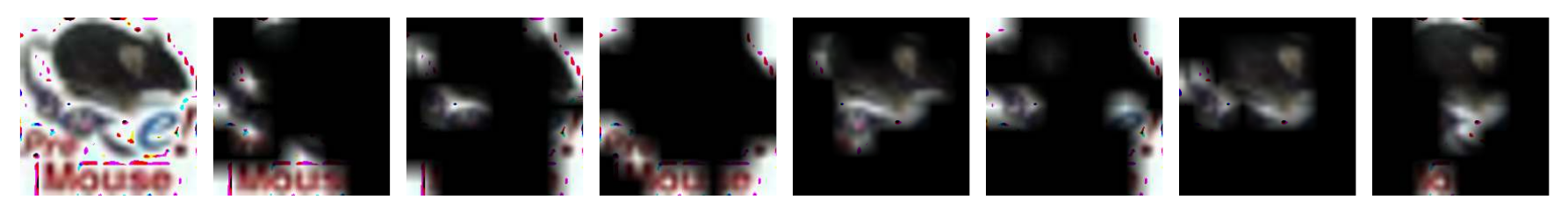}\\
\vspace{-0.2cm}
\includegraphics[width=\textwidth]{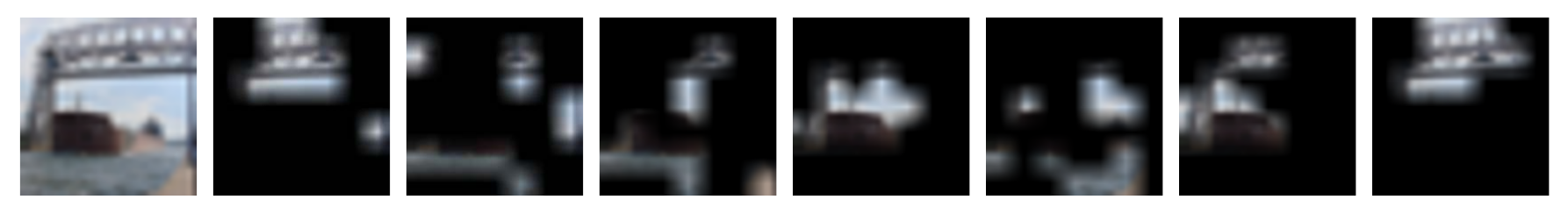}\\
\vspace{-0.2cm}
\includegraphics[width=\textwidth]{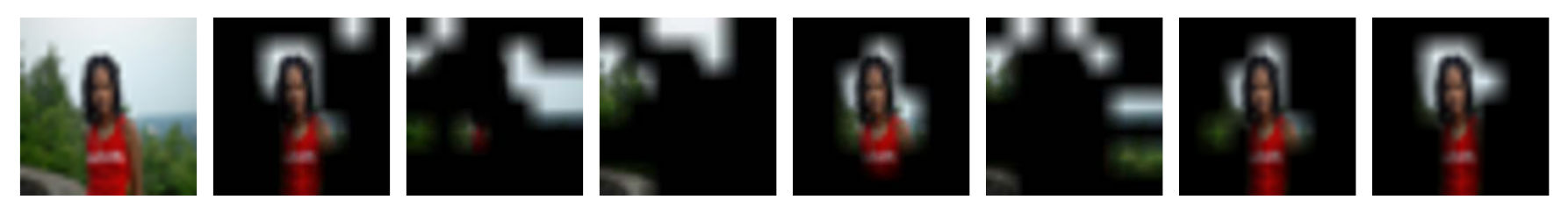}\\
\vspace{-0.2cm}
\includegraphics[width=\textwidth]{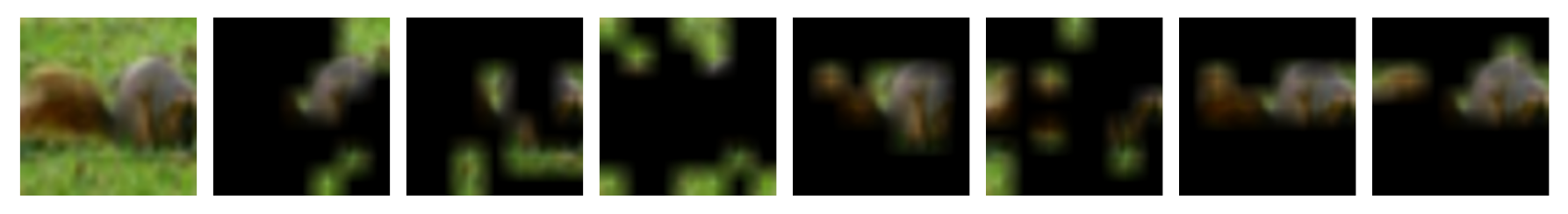}\\
\vspace{-0.2cm}
\includegraphics[width=\textwidth]{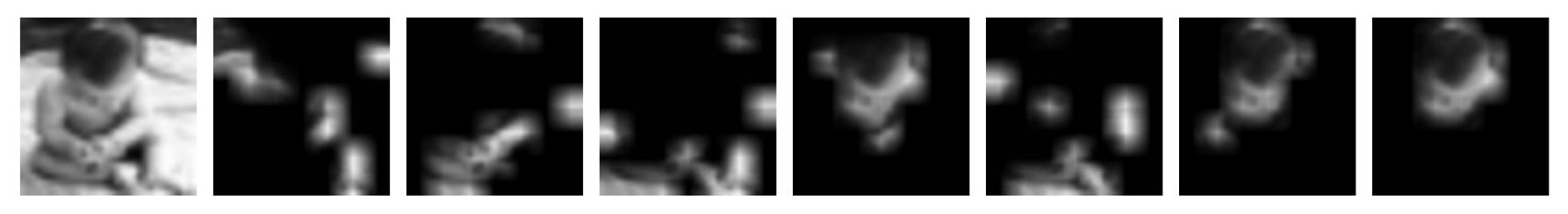}\\
\vspace{-0.2cm}
\includegraphics[width=\textwidth]{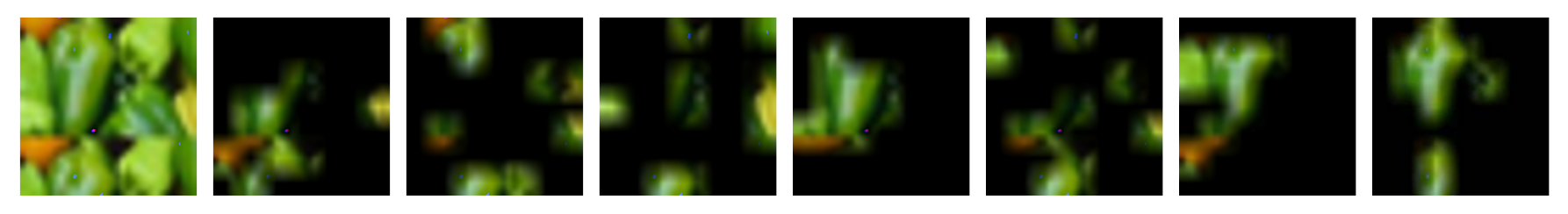}\\
\vspace{-0.2cm}
\includegraphics[width=\textwidth]{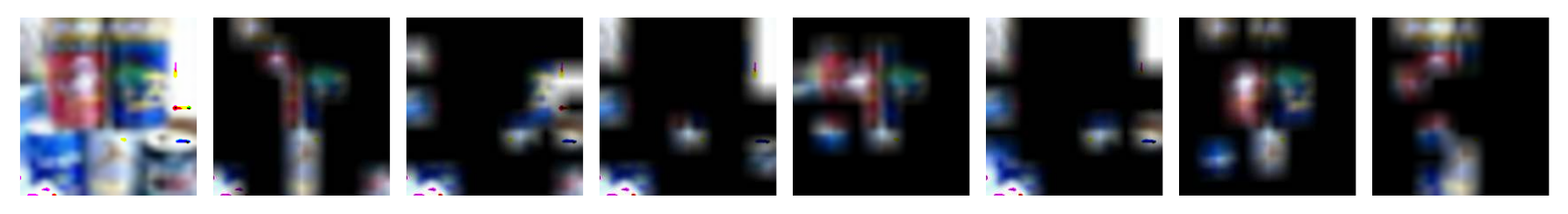}\\
\vspace{-0.2cm}
\includegraphics[width=\textwidth]{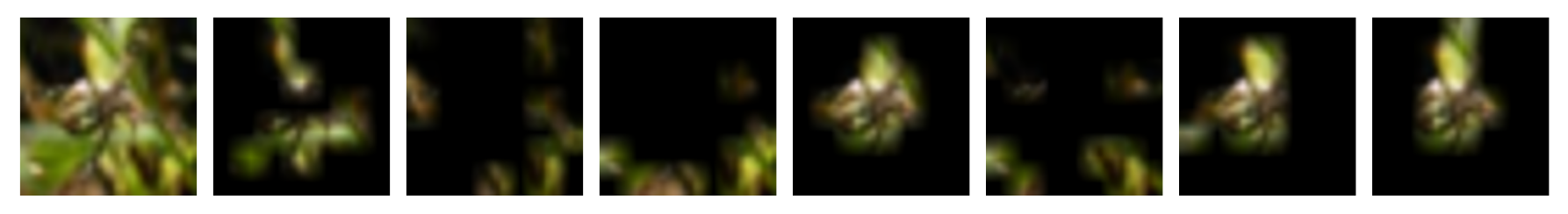}\\
\vspace{-0.2cm}
\includegraphics[width=\textwidth]{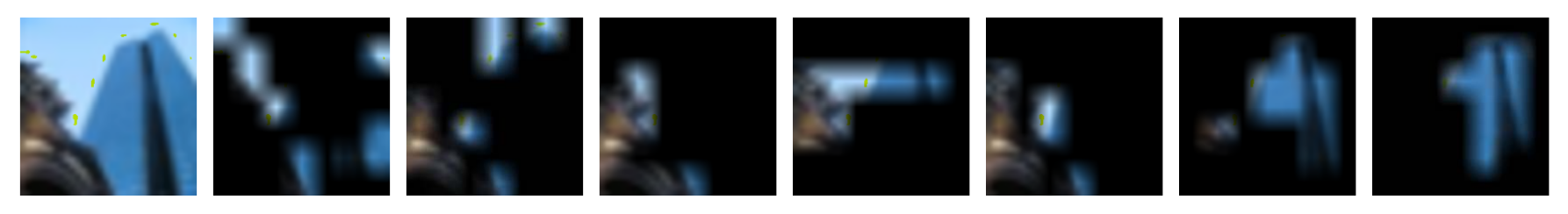}\\
\vspace{-2mm}
\noindent\scriptsize
\begin{minipage}[t]{0.125\textwidth}\centering (a) Input\end{minipage}%
\begin{minipage}[t]{0.125\textwidth}\centering (b) Ground-truth\end{minipage}%
\begin{minipage}[t]{0.125\textwidth}\centering (c) Zero-shot\end{minipage}%
\begin{minipage}[t]{0.125\textwidth}\centering (d) Finetuned on ImageNet\end{minipage}%
\begin{minipage}[t]{0.125\textwidth}\centering (e) Finetuned on ImageNet+X\end{minipage}%
\begin{minipage}[t]{0.125\textwidth}\centering (f) Finetuned on target\end{minipage}%
\begin{minipage}[t]{0.125\textwidth}\centering (g) Proposed\end{minipage}%
\begin{minipage}[t]{0.125\textwidth}\centering (h) Finetuned on target+X\end{minipage}%
\normalsize
\caption{Randomly selected visualization examples for CIFAR100.}
\label{fig:appendix:examples:cifar100}
\end{figure}

\begin{figure}[p]
\centering
\includegraphics[width=\textwidth]{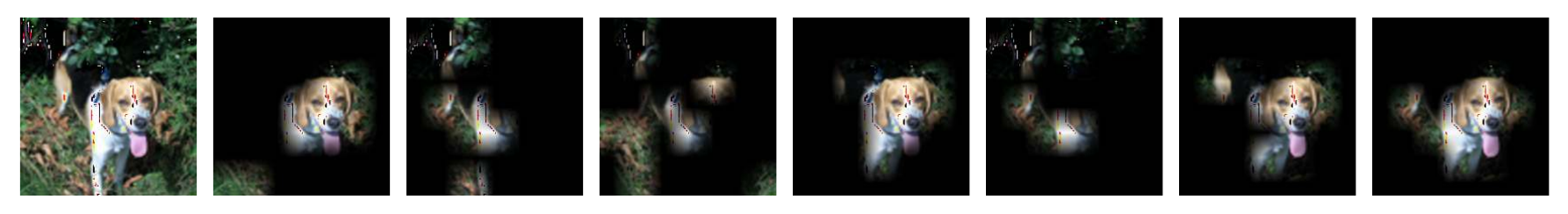}\\
\vspace{-0.2cm}
\includegraphics[width=\textwidth]{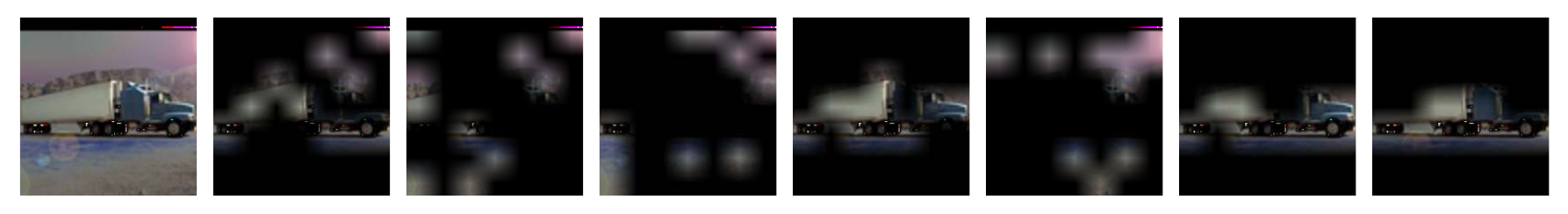}\\
\vspace{-0.2cm}
\includegraphics[width=\textwidth]{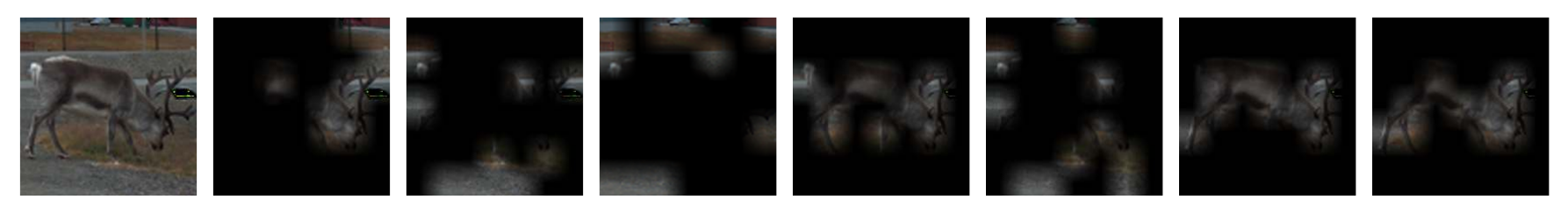}\\
\vspace{-0.2cm}
\includegraphics[width=\textwidth]{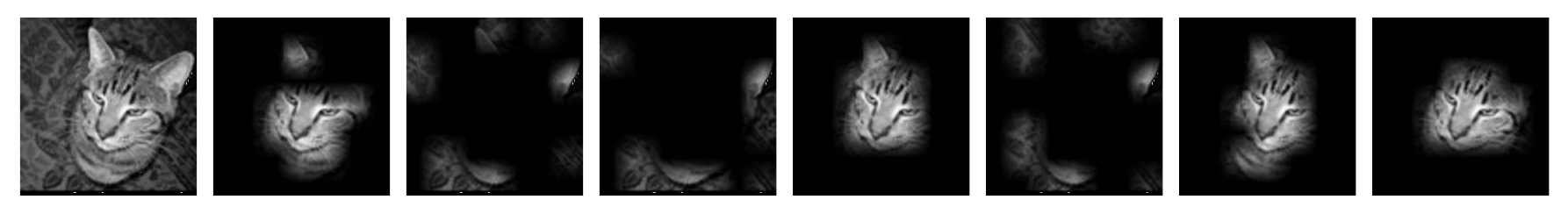}\\
\vspace{-0.2cm}
\includegraphics[width=\textwidth]{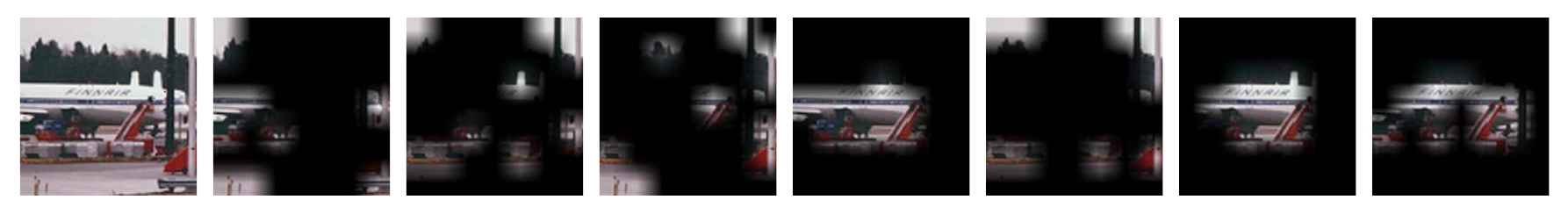}\\
\vspace{-0.2cm}
\includegraphics[width=\textwidth]{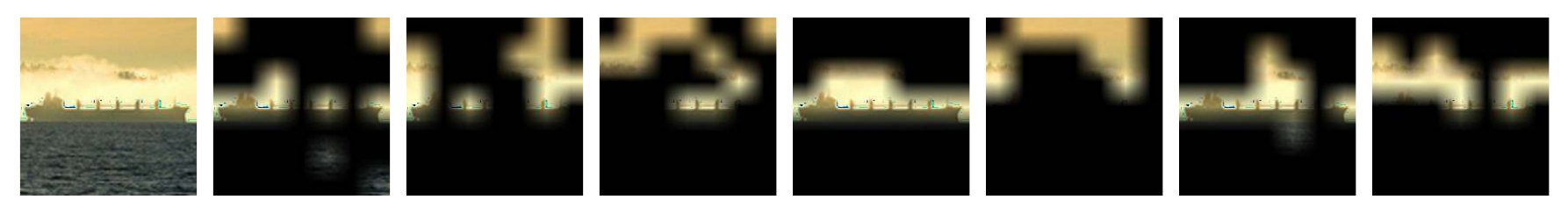}\\
\vspace{-0.2cm}
\includegraphics[width=\textwidth]{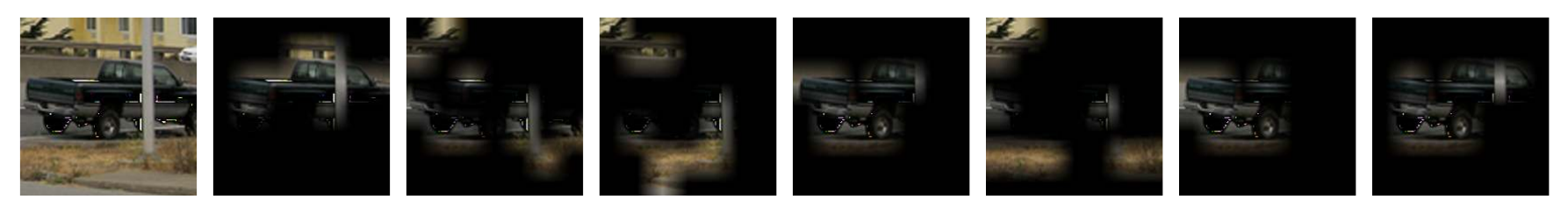}\\
\vspace{-0.2cm}
\includegraphics[width=\textwidth]{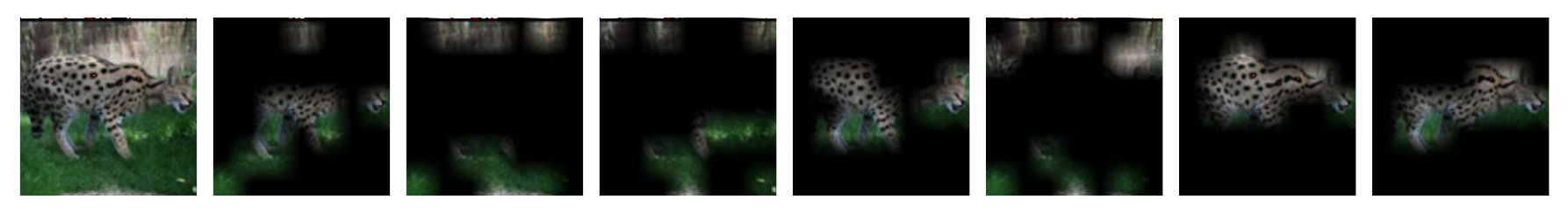}\\
\vspace{-0.2cm}
\includegraphics[width=\textwidth]{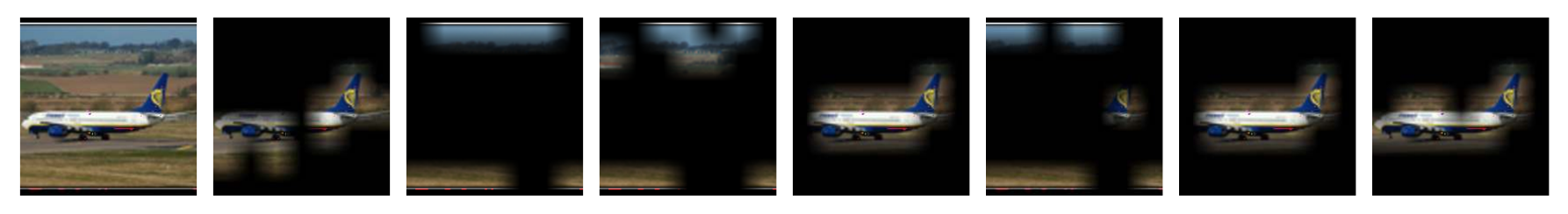}\\
\vspace{-0.2cm}
\includegraphics[width=\textwidth]{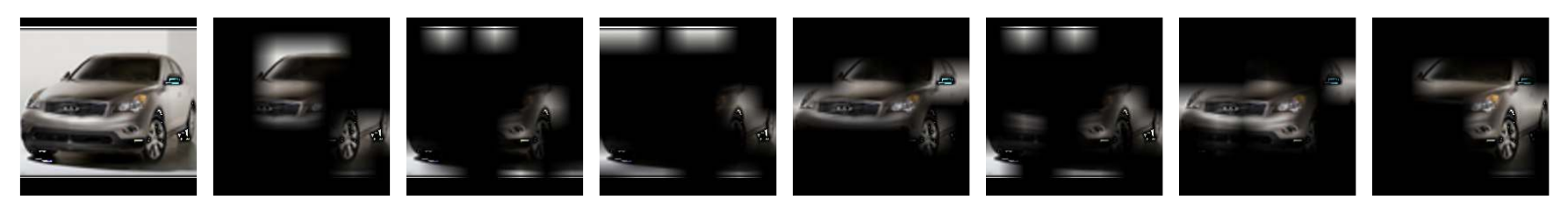}\\
\vspace{-2mm}
\noindent\scriptsize
\begin{minipage}[t]{0.125\textwidth}\centering (a) Input\end{minipage}%
\begin{minipage}[t]{0.125\textwidth}\centering (b) Ground-truth\end{minipage}%
\begin{minipage}[t]{0.125\textwidth}\centering (c) Zero-shot\end{minipage}%
\begin{minipage}[t]{0.125\textwidth}\centering (d) Finetuned on ImageNet\end{minipage}%
\begin{minipage}[t]{0.125\textwidth}\centering (e) Finetuned on ImageNet+X\end{minipage}%
\begin{minipage}[t]{0.125\textwidth}\centering (f) Finetuned on target\end{minipage}%
\begin{minipage}[t]{0.125\textwidth}\centering (g) Proposed\end{minipage}%
\begin{minipage}[t]{0.125\textwidth}\centering (h) Finetuned on target+X\end{minipage}%
\normalsize
\caption{Randomly selected visualization examples for STL10.}
\label{fig:appendix:examples:stl10}
\end{figure}
\space%

\clearpage
\section*{NeurIPS Paper Checklist}

\begin{enumerate}

\item {\bf Claims}
    \item[] Question: Do the main claims made in the abstract and introduction accurately reflect the paper's contributions and scope?
    \item[] Answer: \answerYes{} %
    \item[] Justification: Section~\ref{sec:intro} states three contributions. (i) The task-arithmetic-based transfer method without target-side explanation supervision is introduced in Section~\ref{sec:proposed:transfer}. (ii) The consistent transfer of a single explainability vector learned on ImageNet-1k to 10 diverse target datasets, together with its faithfulness and specificity, is shown in Section~\ref{sec:experiment:imagenet}. (iii) The predictability of transfer success is examined in Sections~\ref{sec:proposed:predict} and~\ref{sec:experiment:predict}. The scope is stated explicitly. The method targets prediction-only, zero-shot-capable VLM-based ViTs, the faithfulness and control-vector evidence is reported in Section~\ref{sec:experiment:imagenet:faithfulness}, and the predictability claim is restricted to a fixed source with held-out targets (Section~\ref{sec:experiment:predict} and Appendix~\ref{sec:appendix:pair:predictor}).

\item {\bf Limitations}
    \item[] Question: Does the paper discuss the limitations of the work performed by the authors?
    \item[] Answer: \answerYes{} %
    \item[] Justification: Appendix~\ref{sec:limitations} discusses five limitations. (i) The method is tailored to ViT-based vision-language models such as CLIP and would require architecture-specific redesign for CNNs or ViTs trained with other strategies. (ii) The framework targets patch-level attributions and does not directly cover concept- or prototype-based explanations. (iii) The evaluation scope is limited, with GT agreement as the primary metric, faithfulness evaluated within the transferred model only, part of the transferred effect being a content-free attribution prior, and the two-task experiment using composited images. (iv) The method modifies the target model and is not appropriate when the deployed model must remain unmodified. (v) The transfer-success predictor is unreliable on a subset of metrics (IoU@1 on ViT-B/32, and IoU@5 and IoU@10 on ViT-L/14 under the oracle-normalized target) and does not generalize to a held-out source under leave-one-source-out evaluation.

\item {\bf Theory assumptions and proofs}
    \item[] Question: For each theoretical result, does the paper provide the full set of assumptions and a complete (and correct) proof?
    \item[] Answer: \answerYes{} %
    \item[] Justification: Appendix~\ref{sec:appendix:theory} derives two theoretical results step by step. (i) The explainability vector $\btau^{\rm S}_{\star}$ is the optimal perturbation under a local quadratic approximation of the training objective in~\bref{eq:proposed:objective}. (ii) Under a local quadratic approximation around $\btheta^{\rm T}_{\rm ft}$, adding $\lambda_2 \btau^{\rm S}_{\star}$ increases the target classification loss by $\frac{\lambda_2^2}{2} R(\btau^{\rm S}_{\star}; \bH^{\rm T})$ with $R(\btau^{\rm S}_{\star}; \bH^{\rm T}) = (\btau^{\rm S}_{\star})^{\top} \bH^{\rm T} \btau^{\rm S}_{\star}$. The underlying assumptions (near-stationary points such as $\nabla \set{L}_{\rm cls}(\btheta^{\rm S}_{\rm ft}) \approx \bzero$ and small-perturbation Taylor expansions) are stated explicitly there.

\item {\bf Experimental result reproducibility}
    \item[] Question: Does the paper fully disclose all the information needed to reproduce the main experimental results of the paper to the extent that it affects the main claims and/or conclusions of the paper (regardless of whether the code and data are provided or not)?
    \item[] Answer: \answerYes{} %
    \item[] Justification: The common experimental protocol (CLIP ViT-B/32 backbone, 10 publicly available target datasets, evaluation metrics, and ground-truth attribution generation via Kernel SHAP with $500$ perturbations) is specified in Sections~\ref{sec:experiment} and~\ref{sec:experiment:imagenet}. The method-specific hyperparameters ($\alpha$, $\lambda_1$, $\lambda_2$) and the reported choices are stated in Sections~\ref{sec:proposed:transfer} and~\ref{sec:experiment}. The predictor pool construction ($N=10$ ImageNet-source pairs with leave-one-target-out evaluation, six model-internal features, and the Lasso/Ridge training protocol) is documented in Section~\ref{sec:experiment:predict} together with Appendix~\ref{sec:appendix:predict}. The complete computing environment (hardware, libraries, and versions) is reported in Appendix~\ref{sec:appendix:environment}, and the ViT-L/14 setup is given in Appendix~\ref{sec:appendix:vitl:setup}. The protocols of the additional experiments (within-model faithfulness metrics and evaluation subsets, control-vector construction, the amortized and TVE baselines including the probe protocol, the leave-one-dataset-out selection of $\lambda_2$, the leave-one-source-out evaluation with bootstrap intervals, the two-task experiment, and the cost measurements) are specified in Appendices~\ref{sec:appendix:faithfulness}--\ref{sec:appendix:stress}, \ref{sec:appendix:pair:predictor}, and~\ref{sec:appendix:complexity}, including seeds and subset-construction rules.

\item {\bf Open access to data and code}
    \item[] Question: Does the paper provide open access to the data and code, with sufficient instructions to faithfully reproduce the main experimental results, as described in supplemental material?
    \item[] Answer: \answerYes{} %
    \item[] Justification: The code is publicly available at the repository URL given in the abstract, and its README gives instructions for reproducing the main experimental results and the additional experiments of the appendix. All datasets are publicly available benchmarks, and the ground-truth attributions are generated by the released code.

\item {\bf Experimental setting/details}
    \item[] Question: Does the paper specify all the training and test details (e.g., data splits, hyperparameters, how they were chosen, type of optimizer) necessary to understand the results?
    \item[] Answer: \answerYes{} %
    \item[] Justification: Section~\ref{sec:experiment} fixes the backbone (CLIP ViT-B/32 with the DataComp-1B pretrain), the input resolution ($224 \times 224$), the patch size ($32 \times 32$), and the number of patches ($M = 49$). It also states $\lambda_1 = 1.0$, the chosen values $\alpha = 0.8$ and $\lambda_2 = 0.7$, and the sweep ranges $\alpha \in \{0.1, \ldots, 0.9\}$ and $\lambda_2 \in \{0.4, \ldots, 1.2\}$. Section~\ref{sec:experiment:imagenet} describes the ground-truth attribution generation (Kernel SHAP with $500$ perturbations on the classifier finetuned on each dataset). The ViT-L/14 counterparts ($224 \times 224$ input, $16 \times 16$ patches of size $14 \times 14$, $M = 256$, and Grad-CAM ground truth) are given in Appendix~\ref{sec:appendix:vitl:setup}. The remaining training-loop details (optimizer, learning rate, batch size, and number of epochs) are provided as configuration files in the public code repository. The settings of the additional experiments (evaluation subsets and metrics, control vectors, baselines, and the construction of the two-task images) are given in Appendices~\ref{sec:appendix:faithfulness}--\ref{sec:appendix:stress}.

\item {\bf Experiment statistical significance}
    \item[] Question: Does the paper report error bars suitably and correctly defined or other appropriate information about the statistical significance of the experiments?
    \item[] Answer: \answerYes{} %
    \item[] Justification: Table~\ref{tab:experiment:imagenet} reports the mean and standard deviation across the 10 target datasets and marks, per metric, the methods with no statistically significant difference from the best (paired $t$-test, $p < 0.05$), as stated in the table caption. The same convention is used for the ViT-L/14 results in Appendix~\ref{sec:appendix:vitl:quantitative} and for the shared- vs.\ separated-head comparison in Appendix~\ref{sec:appendix:shared_vs_separated}. Figure~\ref{fig:experiment:imagenet:scaling_coef} reports 95\% bootstrap intervals of the mean over the nine $\alpha$ values, and the Kernel SHAP comparison figure in Appendix~\ref{sec:appendix:vs_shap} reports 95\% confidence intervals. The transfer-success predictor results in Section~\ref{sec:experiment:predict} and Appendix~\ref{sec:appendix:predict} report permutation-test $p$-values for the held-out Pearson correlations, and Appendix~\ref{sec:appendix:pair:predictor} adds bootstrap 95\% intervals (pair-level, and source-level cluster intervals for the leave-one-source-out evaluation). The two-task experiment in Appendix~\ref{sec:appendix:stress} reports paired bootstrap 95\% intervals. The faithfulness and control-vector results in Appendices~\ref{sec:appendix:faithfulness} and~\ref{sec:appendix:controls} are reported per target as win counts and per-target tables rather than as aggregate significance tests. The circuit-overlap analysis in Appendix~\ref{sec:appendix:circuit_overlap:rq2} reports paired Wilcoxon $p$-values.

\item {\bf Experiments compute resources}
    \item[] Question: For each experiment, does the paper provide sufficient information on the computer resources (type of compute workers, memory, time of execution) needed to reproduce the experiments?
    \item[] Answer: \answerYes{} %
    \item[] Justification: Appendix~\ref{sec:appendix:environment} reports the compute environment. The main experiments use Rocky Linux machines with two Intel Xeon Platinum 8558 2.1GHz (48-core) CPUs, 2{,}048~GB of RAM, and eight NVIDIA H200 SXM GPUs with 141~GB of memory, with the code implemented in Python 3.9.21 and PyTorch 2.6. The additional experiments of the appendix were run on NVIDIA RTX 6000 Ada GPUs. Appendix~\ref{sec:appendix:complexity} reports the one-time source-side cost ($100.4$ GPU-hours for the Kernel SHAP supervision and source training) and the per-image inference cost of each compared method.
    
\item {\bf Code of ethics}
    \item[] Question: Does the research conducted in the paper conform, in every respect, with the NeurIPS Code of Ethics \url{https://neurips.cc/public/EthicsGuidelines}?
    \item[] Answer: \answerYes{} %
    \item[] Justification: The work uses only publicly available image classification benchmarks (ImageNet-1k, the 10 target datasets listed in Section~\ref{sec:experiment:imagenet}, and FashionMNIST for the two-task images), does not involve human subjects, and does not collect or release any new personal data. The authors have reviewed the NeurIPS Code of Ethics, and the research conforms to it.

\item {\bf Broader impacts}
    \item[] Question: Does the paper discuss both potential positive societal impacts and negative societal impacts of the work performed?
    \item[] Answer: \answerYes{} %
    \item[] Justification: Appendix~\ref{sec:appendix:impact_statement} discusses both sides. On the positive side, the explanation supervision is needed only once on the source, which can lower the barrier to adopting explainable AI in resource-constrained scenarios and may help users inspect the decisions of AI systems. On the negative side, the transferred explanation capability is not necessarily optimized for the target domain and may produce lower-quality or misleading explanations on which users could base incorrect decisions. Plausible-looking explanations can also raise trust beyond what is warranted, since part of the transferred effect is a content-free attribution prior. As mitigations, the section recommends checking the transferred explanations with faithfulness checks or a small set of explanation labels and, when necessary, finetuning with a small amount of labeled explanation data.
    
\item {\bf Safeguards}
    \item[] Question: Does the paper describe safeguards that have been put in place for responsible release of data or models that have a high risk for misuse (e.g., pretrained language models, image generators, or scraped datasets)?
    \item[] Answer: \answerNA{} %
    \item[] Justification: The paper does not release any high-risk artifact. No generative model, no language model, and no scraped dataset is introduced. The only released artifact is a code repository that finetunes a publicly available CLIP backbone on standard image classification benchmarks and computes patch-level attributions, which does not pose a misuse risk that would require additional safeguards.

\item {\bf Licenses for existing assets}
    \item[] Question: Are the creators or original owners of assets (e.g., code, data, models), used in the paper, properly credited and are the license and terms of use explicitly mentioned and properly respected?
    \item[] Answer: \answerYes{} %
    \item[] Justification: All existing assets used in the paper are credited via citations to their original publications: the CLIP backbone with its DataComp-1B pretrain~\cite{radford2021learning,gadre2023datacomp,ilharco_gabriel_2021_5143773}, ImageNet-1k~\cite{deng2009imagenet} as the source dataset, the 10 target datasets~\cite{krause20133d,cimpoi2014describing,helber2019eurosat,stallkamp2011german,cheng2017remote,xiao2016sun,netzer2011reading,lecun1998mnist,Krizhevsky2009LearningML,coates2011analysis}, the \texttt{pytorch-grad-cam}~\cite{jacobgilpytorchcam} library used to generate the ViT-L/14 ground-truth attributions, FashionMNIST~\cite{xiao2017fashionmnist} used in the two-task experiment of Appendix~\ref{sec:appendix:stress}, and the public TVE checkpoint~\cite{wang2024tve} used as a baseline in Appendix~\ref{sec:appendix:alternatives}. All of these are publicly available research assets used in accordance with their stated terms of use.

\item {\bf New assets}
    \item[] Question: Are new assets introduced in the paper well documented and is the documentation provided alongside the assets?
    \item[] Answer: \answerYes{} %
    \item[] Justification: The only new asset introduced by the paper is the public code repository. It covers source and target finetuning, explainability-vector transfer via task arithmetic, evaluation on the 10 target datasets, the additional experiments of the appendix, and the script that composites the two-task images from MNIST and FashionMNIST. The repository is accompanied by a README describing dependencies, license, and entry-point commands.

\item {\bf Crowdsourcing and research with human subjects}
    \item[] Question: For crowdsourcing experiments and research with human subjects, does the paper include the full text of instructions given to participants and screenshots, if applicable, as well as details about compensation (if any)?
    \item[] Answer: \answerNA{} %
    \item[] Justification: The paper does not involve crowdsourcing or research with human subjects. All experiments are conducted on publicly available image classification benchmarks.

\item {\bf Institutional review board (IRB) approvals or equivalent for research with human subjects}
    \item[] Question: Does the paper describe potential risks incurred by study participants, whether such risks were disclosed to the subjects, and whether Institutional Review Board (IRB) approvals (or an equivalent approval/review based on the requirements of your country or institution) were obtained?
    \item[] Answer: \answerNA{} %
    \item[] Justification: The paper does not involve research with human subjects. All experiments are conducted on publicly available image classification benchmarks, so no IRB approval is required.

\item {\bf Declaration of LLM usage}
    \item[] Question: Does the paper describe the usage of LLMs if it is an important, original, or non-standard component of the core methods in this research? Note that if the LLM is used only for writing, editing, or formatting purposes and does \emph{not} impact the core methodology, scientific rigor, or originality of the research, declaration is not required.
    \item[] Answer: \answerNA{} %
    \item[] Justification: The core methodology is the explainability transfer between CLIP-based ViT image classifiers via task arithmetic. LLMs are not used as an important, original, or non-standard component of this method. LLM-based tools were used only to assist with writing, editing, and figure scripting, which does not require declaration under the policy.

\end{enumerate}
\space%

\end{document}